\DocumentMetadata{}
\documentclass[backref=page]{article}

\def\inclapp{1}
\def\viewchanges{0}
\def\viewauthors{1}
\def\preprint{0}
\def\viewkeywords{0}
\def\usehyperlinks{1}
\def\addackn{0}

\usepackage[dvipsnames]{xcolor}

\if\preprint1
    \usepackage[preprint]{tmlr}
\else
    \if\viewauthors1
        \usepackage[accepted]{tmlr}
    \else
        \usepackage{tmlr}
    \fi
\fi

\usepackage{natbib}

\usepackage[utf8]{inputenc} 
\usepackage[T1]{fontenc}    
\usepackage{url}            
\usepackage{booktabs}       
\usepackage{amsfonts}       
\usepackage{nicefrac}       
\usepackage{microtype}      
\usepackage{amsthm}
\usepackage{amsmath}
\usepackage{amssymb}
\usepackage{enumitem}
\usepackage{bbm}
\allowdisplaybreaks        
\definecolor{darkmagenta}{rgb}{0.55, 0.0, 0.55}
\if\usehyperlinks1
	\usepackage[colorlinks, citecolor=blue]{hyperref}       
    \colorlet{linkcolor}{darkmagenta}
    \newcommand{\mathlinkcolored}[1]{\mathcolor{linkcolor}{#1}}
\else
    \usepackage{nohyperref}
    \newcommand{\mathlinkcolored}[1]{#1}
\fi
\hypersetup{linkcolor=linkcolor} 
\usepackage{cleveref}
\usepackage{csquotes}

\usepackage{graphicx,wrapfig,lipsum}
\usepackage{subfigure}
\usepackage{verbatim}
\usepackage{bm}
\usepackage{adjustbox}
\usepackage{footnote}
\let\myAND\AND
\let\AND\undefined
\usepackage{algorithm}
\usepackage{algorithmic}
\usepackage[acronym,toc]{glossaries} 
\usepackage{mathtools}
\usepackage[mathscr]{eucal}

\if\inclapp0
	\usepackage{xr}
\fi

\newtheorem{theorem}{Theorem}[section]
\newtheorem{lemma}[theorem]{Lemma}
\newtheorem{definition}[theorem]{Definition}
\newtheorem{lem}[theorem]{Lemma}
\newtheorem{prop}[theorem]{Proposition}
\newtheorem{rem}[theorem]{Remark}
\newtheorem{cor}[theorem]{Corollary}

\newtheorem{example}[theorem]{Example}
\newtheorem{assumption}[theorem]{Assumption}
\crefname{assumption}{assumption}{assumptions}

\let\P\undefined%
\newcommand{\1}{\mathbbm{1}}
\newcommand{\PBlack}{\mathbb{P}}
\newcommand{\F}{\mathcal{F}}
\newcommand{\E}{\mathbb{E}} 
\newcommand{\N}{\mathbb{N}} 
\newcommand{\R}{\mathbb{R}} 
\newcommand{\argmin}{\operatorname{argmin}} 
\newcommand{\id}{\operatorname{id}} 
 
\newcommand{\sigmab}{\boldsymbol{\sigma}}

\def\fastCompileJakob{1}
\usepackage{transparent}
\DeclareMathOperator{\supp}{supp}

\glsaddkey
{alttextLong}           
{todo}              
{\glsentryLong}  
{\GlsentryLong}  
{\glsLong}       
{\GlsLong}       
{\GLSLong}       

\glsaddkey
{alttextt}           
{todo}              
{\glsentryt}  
{\Glsentryt}  
{\glst}       
{\Glst}       
{\GLSt}       

\glsaddkey
{alttextT}           
{todo}              
{\glsentryT}  
{\GlsentryT}  
{\glsT}       
{\GlsT}       
{\GLST}       

\glsaddkey
{alttexttk}           
{todo}              
{\glsentrytk}  
{\Glsentrytk}  
{\glstk}       
{\Glstk}       
{\GLStk}       

\glsaddkey
{alttexts}           
{todo}              
{\glsentrys}  
{\Glsentrys}  
{\glss}       
{\Glss}       
{\GLSs}       

\glsaddkey
{alttextOm}           
{todo}              
{\glsentryOm}  
{\GlsentryOm}  
{\glsOm}       
{\GlsOm}       
{\GLSOm}       

\glsaddkey
{alttextF}           
{todo}              
{\glsentryF}  
{\GlsentryF}  
{\glsF}       
{\GlsF}       
{\GLSF}       

\glsaddkey
{alttextFF}           
{todo}              
{\glsentryFF}  
{\GlsentryFF}  
{\glsFF}       
{\GlsFF}       
{\GLSFF}       

\glsaddkey
{alttextFt}           
{todo}              
{\glsentryFt}  
{\GlsentryFt}  
{\glsFt}       
{\GlsFt}       
{\GLSFt}       

\glsaddkey
{alttextP}           
{todo}              
{\glsentryP}  
{\GlsentryP}  
{\glsP}       
{\GlsP}       
{\GLSP}       

\newcommand{\optionalSub}[1]{\ifblank{#1}{}{\ensuremath{_{#1}}}}
\newcommand{\optionalSup}[1]{\ifblank{#1}{}{\ensuremath{^{#1}}}}

\NewExpandableDocumentCommand{\mandatorySub}{O{command}m}{\ifblank{#2}{\PackageWarning {NotationAppendixGlossaryJakob}{The argument of #1 is not supposed to be empty}}{\ensuremath{_{#2}}}}
\newcommand{\mandatorySup}[2][command]{\ifblank{#2}{\PackageWarning {NotationAppendixGlossaryJakob}{The argument of #1 is not supposed to be empty}}{\ensuremath{^{#2}}}}

\newcommand{\nblack}[1]{n}
\newcommand{\njNotNumberOfObservations}[1]{n\mandatorySup[\njNotNumberOfObservations]{#1}}

\newcommand{\mathglslink}[2]{\mathrlap{\glslink{#1}{\phantom{\ensuremath{#2}}}}\mathlinkcolored{#2}} 

\if\fastCompileJakob1
\newcommand{\myglslink}[2]{\rlap{\glslink{#1}{\phantom{\ensuremath{#2}}}}\mathlinkcolored{#2}} 
\else
\newcommand{\myglslink}[2]{\mathrlap{\glslink{#1}{\phantom{\ensuremath{#2}}}}\mathlinkcolored{#2}} 
\fi

\newcommand{\makeCommandsForGLSEntryWithoutArguments}[3]{
\newcommand{#1}{\ensuremath{\myglslink{#2}{#3}}%
}}

\if\fastCompileJakob1
\newcommand{\makeCommandsForGLSEntryColoredSubSuper}[3]{
\newcommand{#1}[2]{\ensuremath{\glslink{#2}{#3\ifblank{##1}{\PackageWarning {NotationAppendixGlossaryJakob}{The argument of #1 is not supposed to be empty}}{_{##1}}\ifblank{##2}{}{^{\mathcolor{black}{##2}}}}}%
}}
\else
\newcommand{\makeCommandsForGLSEntryColoredSubSuper}[3]{
\newcommand{#1}[2]{\ensuremath{\myglslink{#2}{#3\ifblank{##1}{\PackageWarning {NotationAppendixGlossaryJakob}{The argument of #1 is not supposed to be empty}}{_{##1}}}\ifblank{##2}{}{^{##2}}}%
}}
\fi

\newcommand{\makeCommandsForGLSEntryColoredMandatorySub}[3]{
\NewDocumentCommand#1{m}{\ensuremath{\myglslink{#2}{#3\mandatorySub[#1]{##1}}}%
}}

\newcommand{\makeCommandsForGLSEntryOptionalSub}[3]{
\NewDocumentCommand#1{m}{\ensuremath{\myglslink{#2}{#3}\optionalSub{##1}}%
}}

\newcommand{\makeCommandsForGLSEntrySubIcludedSuper}[4]{
\NewDocumentCommand#1{mm}{\colorlet{saved}{.}\ensuremath{\glslink{#2}{#3\ifblank{##1}{}{_{\mathcolor{saved}{##1}}}\ifblank{##2}{^{#4}}{^{#4,\mathcolor{saved}{##2}}}}}%
}}

\if\fastCompileJakob1
\newcommand{\makeCommandsForGLSEntrySuper}[3]{
\NewDocumentCommand#1{m}{\colorlet{saved}{.}\ensuremath{\glslink{#2}{#3\ifblank{##1}{}{^{\mathcolor{saved}{##1}}}}}%
}}
\else
\newcommand{\makeCommandsForGLSEntrySuper}[3]{
\NewDocumentCommand#1{m}{\ensuremath{\myglslink{#2}{#3}\optionalSup{##1}}}%
}
\fi

\newcommand{\makeCommandsForGLSEntryPost}[3]{
\NewDocumentCommand#1{m}{\colorlet{saved}{.}\ensuremath{\glslink{#2}{\mathcolor{saved}{\mathlinkcolored{#3}##1}}}%
}}

\newcommand{\inSectionTildeButInOtherSection}[3][1]{In \if#11\Cref{sec:Recall: the PD-NJ-ODE,sec:PD-NJ-ODE with Noisy Observations}\else\Cref{sec:PD-NJ-ODE with Noisy Observations}\fi, #2, but in \Cref{sec:PD-NJ-ODE with Dependence betweenXand Observation Framework} (and all sections thereafter including the appendix), #3}

\newcommand{\twoDifferentGeneralSpacesText}[4][is]{\inSectionTildeButInOtherSection{$#2$ #1 defined on #3}{ $#2$ #1 defined on #4}}

\newcommand{\twoDifferentSpacesText}[2][is]{\twoDifferentGeneralSpacesText[#1]{#2}{\tOmFP{}}{\OmFP{}}}

\newcommand{\twoDifferentFilteredSpacesText}[2][is]{\twoDifferentGeneralSpacesText[#1]{#2}{\tOmFFP{}}{\OmFFP{}}}

\newcommand{\inSectionNoiselessButInOtherSection}[2]{In \Cref{sec:Recall: the PD-NJ-ODE,sec:PD-NJ-ODE with Dependence betweenXand Observation Framework,sec:Dependent Observation Framework -- a Black--Scholes Example}, #1, but in \Cref{sec:PD-NJ-ODE with Noisy Observations} and in the appendix, #2. In \Cref{sec:Noisy Observations -- Brownian Motion with Gaussian Observation Noise,sec:Physionet with Observation Noise} we compare both losses against each other. Note that in \Cref{sec:More General Noise Structure Conditional Moments} we use another modification of the loss~\eqref{equ:loss noisy obs generalised} for the case of noise with known non-zero mean}

\makeglossaries
\newglossaryentry{X}
{
  name={\ensuremath{X}},
  alttextt={\ensuremath{X_{\normalcolor t}}},
  alttexts={\ensuremath{X_{\normalcolor s}}},
  description={The stochastic process~$X :={(X_t)}_{t \in [0,T]}$ we want to study (which is an adapted c\`adl\`ag  stochastic process $X:\gls{OmFFP}\to {\R^{\gls{dX}}}^{[0,\gls{T}]}$). In our notation $\gX{t,k}^{(j)}$ refers to the $k$-th coordinate of the $j$-th path at time $t$ for $1\le k \le \gls{dX}$, $1\le j \le N$ and $t\in [0,\gls{T}]$ (see \Cref{sec:Recall: the PD-NJ-ODE}). E.g., for a medical data-set, the index $j$ could correspond to a patient and the first coordinate could correspond to the body-temperature and the second coordinate could correspond to the blood-pressure, then $\gX{t,1}^{(7)}\omb=36.9$ would mean that the 7\textsuperscript{th} patient had a body temperature of $36.9$\textdegree C at time (or age) $t$}
}
\makeCommandsForGLSEntryOptionalSub{\gX}{X}{X}
\newcommand{\mathbfX}{\mathbf{X}}
\newcommand{\Xblack}{X}

\newglossaryentry{dX}
{
  name={\ensuremath{d_X}},
  description={The dimension~$d_X \in \N$ of $\gls{X}$ (i.e., $\gX{}_t\omb\in\R^{d_X}$) (see \Cref{sec:Recall: the PD-NJ-ODE})}
}
\newglossaryentry{T}
{
  name={\ensuremath{T}},
  description={The largest time~$T \in\Rp$ we consider (see \Cref{sec:Recall: the PD-NJ-ODE})}
}
\newglossaryentry{OmFFP}
{
  name={\ensuremath{\left(\Omega, \F, \mathbb{F}, \PBlack \right)}},
  alttextLong={\ensuremath{\left(\Omega, \F, \mathbb{F} := (\F_t)_{0 \leq t \leq \gls{T}}, \PBlack\right)}},
  alttextOm={\ensuremath{\Omega}},
  alttextF={\ensuremath{\F}},
  alttextFF={\mathbb{F}},
  alttextFt={\ensuremath{\F_t}},
  alttextP={\ensuremath{\PBlack}},
  description={The filtered probability space  $\left(\Omega, \F, \mathbb{F} := (\F_t)_{0 \leq t \leq \gls{T}}, \PBlack\right)$ on which \gls{X} is defined (see \Cref{sec:Recall: the PD-NJ-ODE}). In \Cref{sec:PD-NJ-ODE with Dependence betweenXand Observation Framework} (and all sections thereafter including the appendix) all stochastic objects $\gls{X}, \gn{}, \gls{K}, \tk{i}, \tau, \gM{}, \At{t}$ are defined on this probability space. However, in \Cref{sec:Recall: the PD-NJ-ODE,sec:PD-NJ-ODE with Noisy Observations} only $\gls{X}$ is defined on this space, while the observation framework $\gn{}, \gls{K}, \tk{i}, \tau, \gM{}$ is defined on a different filtered probability space \gls{tOmFFP}}
}
\makeCommandsForGLSEntryColoredSubSuper{\gF}{OmFFP}{\glsentryF{OmFFP}}
\newcommand{\Om}[1]{\glsOm{OmFFP}#1}
\makeCommandsForGLSEntryPost{\OmFFP}{OmFFP}{\glsentrytext{OmFFP}}
\makeCommandsForGLSEntryPost{\OmFP}{OmFFP}{\left(\Omega, \F, \PBlack\right)}
\makeCommandsForGLSEntryPost{\gP}{OmFFP}{\PBlack}
\makeCommandsForGLSEntryWithoutArguments{\P}{OmFFP}{\PBlack}
\newcommand{\Omblack}{\Omega}
\newcommand{\Fblack}{\F}

\newglossaryentry{Xs}
{
  name={\ensuremath{X^\star}},
  alttextt={\ensuremath{X^\star_{\normalcolor t}}},
  alttextT={\ensuremath{X^\star_{\normalcolor \gls{T}}}},
  description={The running maximum process of \gls{X}, i.e., $X^\star_t\omb :=\sup_{s \in [0,t]}\onenorm{\gX{s}\omb} \ \forall t \in [0,\gls{T}] \ \forall \omega\in\glsOm{OmFFP}$ with $X^\star:\gls{OmFFP}\to {\Rpz}^{[0,\gls{T}]}$ (see \Cref{sec:Recall: the PD-NJ-ODE})}
}
\makeCommandsForGLSEntrySubIcludedSuper{\gXs}{Xs}{X}{\star}

\newglossaryentry{J}
{
  name={\ensuremath{\mathcal{J}}},
  description={The random set of the jump times $\mathcal{J}:\OmFP{}\to \Powerset{([0,\gls{T}])}$ of \gls{X} (see \Cref{sec:Recall: the PD-NJ-ODE})}
}

\newglossaryentry{powerset}
{
  name={\ensuremath{\mathscr{P}}},
  description={The powerset $\mathscr{P}(S)$ of a set $S$ is the set of all subsets of $S$}
}
\newcommand{\Powerset}[1]{\gls{powerset}\left(#1\right)}

\newglossaryentry{tOmFFP}
{
  name={\ensuremath{\left(\tilde{\Omega}, \tilde{\F}, \tilde{\mathbb{F}}, \tilde{\PBlack} \right)}},
  alttextLong={\ensuremath{\left(\tilde\Omega, \tilde{\F},  \tilde{\mathbb{F}} := ( \tilde{\F}_t )_{0 \leq t \leq \gls{T}}, \tilde{\PBlack}\right)}},
  alttextOm={\ensuremath{\tilde{\Omega}}},
  alttextF={\ensuremath{\tilde{\F}}},
  alttextFF={\ensuremath{\tilde{\mathbb{F}}}},
  alttextFt={\ensuremath{\tilde{F_t}}},
  alttextP={\ensuremath{\tilde{\PBlack}}},
  description={The filtered probability space  $\left(\tilde\Omega, \tilde{\F},  \tilde{\mathbb{F}} := ( \tilde{\F}_t )_{0 \leq t \leq \gls{T}}, \tilde{\PBlack}\right)$ on which $\gls{n}, \gls{K}, \tk{i}, \tau, \gls{M}$ are defined describing the observation framework, but $\gls{X}$ is not defined on this space (see \Cref{sec:Recall: the PD-NJ-ODE}). In \Cref{sec:PD-NJ-ODE with Dependence betweenXand Observation Framework} (and all sections thereafter including the appendix) this space is not  needed since all stochastic objects $\gls{X}, \gls{n}, \gls{K}, \tk{i}, \tau, \gls{M}, \At{t}$ are defined on \gls{OmFFP}. However, in \Cref{sec:Recall: the PD-NJ-ODE,sec:PD-NJ-ODE with Noisy Observations} this space is used for $\gls{n}, \gls{K}, \tk{i}, \tau, \gls{M}$ related to the events of observations}
}
\makeCommandsForGLSEntryPost{\tOmFFP}{tOmFFP}{\glsentrytext{tOmFFP}}
\makeCommandsForGLSEntryPost{\tOmFP}{tOmFFP}{\left(\tilde\Omega, \tilde{\F}, \tilde{\PBlack}\right)}
\makeCommandsForGLSEntryPost{\tP}{tOmFFP}{\tilde{\PBlack}}
\newcommand{\tFt}[1]{\glslink{tOmFFP}{\tilde{\mathcal{F}}_{#1}}}

\newglossaryentry{n}
{
  name={\ensuremath{n}},
  description={The random number of observations $n: \tOmFP{}  \to \N_{\geq 0}$ up to time $T$ 
  (see \Cref{sec:Recall: the PD-NJ-ODE}). Every observation time $\tk{i}\in[0,\gls{T}]$ counts as 1 observation for this count (also for incomplete observations). \twoDifferentSpacesText{n}. In our medical example, $n^{(j)}$ denotes the number of observation times for the $j$-th patient
  }
}
\makeCommandsForGLSEntrySuper{\gn}{n}{n} 

\newglossaryentry{K}
{
  name={\ensuremath{K}},
  description={The \enquote{maximal} value of $\gn{}$, i.e., the essential supremum $K := \sup \left\{k \in \N \, | \, \glsP{tOmFFP}(\gn{} \geq k) > 0 \right\} \in \N \cup\{\infty\}$ of $\gn{}$ (see \Cref{sec:Recall: the PD-NJ-ODE}). 
  \inSectionTildeButInOtherSection{$K$ is defined with respect to \tP{}}{$K$ is defined with respect to \gP{}}}
}

\newglossaryentry{tk}
{
  name={\ensuremath{t_i}},
  description={The random observation times $t_i: \tOmFFP{}  \to [0,\gls{T}] \cup \{ \infty \}$ for $0 \leq i \leq \gls{K}$ are sorted stopping times, with $t_i(\tilde{\omega}) := \infty$ if $\gn{}(\tilde{\omega}) < i$ (see \Cref{sec:Recall: the PD-NJ-ODE}). In our practical implementation we replace \enquote{$\infty$} by \enquote{$\gls{T}$} (see \Cref{algo:1}), since we are not interested in times after $\gls{T}$ anyway. \twoDifferentFilteredSpacesText[are]{t_i}}
}
\makeCommandsForGLSEntryColoredSubSuper{\tksup}{tk}{t}
\makeCommandsForGLSEntryColoredMandatorySub{\tk}{tk}{t}

\let\oldtau\tau
\newglossaryentry{tau}
{
  name={\ensuremath{\oldtau}},
  description={The last observation time $\oldtau(t)$ before a certain time $t$, i.e., $\oldtau : [0,\gls{T}] \times \glsOm{tOmFFP}  \to [0,\gls{T}], \; (t, \tilde\omega) \mapsto \tau(t, \tilde\omega) := \max\{ \tk{i}(\tilde\omega) | 0 \leq i \leq \gn{}(\tilde\omega), \tk{i}(\tilde\omega) \leq t \}$ (see \Cref{sec:Recall: the PD-NJ-ODE}). \twoDifferentSpacesText{\oldtau}, i.e., $\oldtau : [0,\gls{T}] \times \glsOm{OmFFP}  \to [0,\gls{T}]$}
}
\renewcommand{\tau}{\ensuremath{\myglslink{tau}{\oldtau}}}

\newglossaryentry{M}
{
  name={\ensuremath{M}},
  description={The observation mask~$M = (M_k)_{0 \leq k \leq \gls{K}}$, which is a sequence of random variables on  $(\glsOm{tOmFFP} , \glsF{tOmFFP}, \glsP{tOmFFP} )$ taking values in $\{ 0,1 \}^{\gls{dX}}$ such that $M_k$ is $\tFt{\tk{k}}$-measurable.
The $j$-th coordinate of the $k$-th element of the sequence $M$, i.e., $M_{k,j}$, signals whether $\gX{}_{\tk{k}, j}$, denoting the $j$-th coordinate of the stochastic process at observation time $\tk{k}$, is observed. By abuse of notation we also write $M_{\tk{k}} := M_{k}$ (see \Cref{sec:Recall: the PD-NJ-ODE}). \twoDifferentSpacesText{M}}
}
\makeCommandsForGLSEntryOptionalSub{\gM}{M}{M}

\newglossaryentry{At}
{
  name={\ensuremath{\mathbb{A} := (\mathcal{A}_t)_{t \in [0,\gls{T}]}}},
  description={The filtration of the currently available information $\mathbb{A} := (\mathcal{A}_t)_{t \in [0,\gls{T}]}$ defined by 
\begin{equation*}
\mathcal{A}_t := \boldsymbol{\sigma}\left(\gX{}_{\tk{i}, j}, \tk{i}, \gM{\tk{i}} \,\middle|\, \tk{i} \leq t,\, j \in \{1 \leq l \leq \gls{dX} | \gM{\tk{i}, l} = 1  \} \right),
\end{equation*}%
in all sections, where we have noiseless observations $\gX{\tk{i}}$ and analogously defined as
\begin{equation*}
\mathcal{A}_t := \boldsymbol{\sigma}\left(\gO{\tk{i}, j}, \tk{i}, \gM{\tk{i}} \,\middle|\, \tk{i} \leq t,\, j \in \{1 \leq l \leq \gls{dX} | \gM{\tk{i}, l} = 1  \} \right),
\end{equation*}
in all other sections, where we have noisy observations $\gO{\tk{i}}$.
In both cases this corresponds to the information obtained from seeing the observations until time $t$ where $\boldsymbol\sigma(\cdot)$ denotes the generated $\sigma$-algebra (see \Cref{sec:Recall: the PD-NJ-ODE,sec:Setting with Noisy Observations}).
We use the notion of stopped $\sigmab$-algebras
\begin{equation*}
\mathcal{A}_{\tk{k}} := \boldsymbol{\sigma}\left(\gX{}_{\tk{i}, j}, \tk{i}, \gM{\tk{i}} \,\middle|\, i\leq k,\, j \in \{1 \leq l \leq \gls{dX} | \gM{\tk{i}, l} = 1  \} \right),
\end{equation*}%
and pre-stopped $\sigmab$-algebras
\begin{equation*}
\mathcal{A}_{\tk{k}-} := \boldsymbol{\sigma}\left(\gX{}_{\tk{i}, j}, \tk{i}, \gM{\tk{i}}, \tk{k} \,\middle|\, i < k,\, j \in \{1 \leq l \leq \gls{dX} | \gM{\tk{i}, l} = 1  \} \right)
\end{equation*}
from \citet[Definitions~2.37 and~8.1]{KarandikarRao2018}. In the sections with noise, we replace $\gls{X}$ by $\gls{O}$%
}%
}
\newcommand{\bA}{\ensuremath{\myglslink{At}{\ensuremath{\mathbb{A}}}}}
\makeCommandsForGLSEntryColoredMandatorySub{\At}{At}{\mathcal{A}}

\newglossaryentry{hX}
{
  name={\ensuremath{\hat{X} = (\hat X_t)_{0 \leq t \leq \gls{T}}}},
  description={The conditional expectation process of $\gls{X}$, which is its $L^2$-optimal prediction \citep[Proposition 2.5]{krach2022optimal} given the currently available information, is defined as $\hat{X} = (\hat X_t)_{0 \leq t \leq \gls{T}}$, with $\hat{X}_t := \E_{\glsP{OmFFP}\times\glsP{tOmFFP}}[\gX{}_t | \At{t}]$ (see \Cref{sec:Recall: the PD-NJ-ODE}). \inSectionTildeButInOtherSection{the expectation is taken with respect to $\glsP{OmFFP}\times\glsP{tOmFFP}$}{the expectation is taken with respect to $\glsP{OmFFP}$}}
}
\makeCommandsForGLSEntryOptionalSub{\hX}{hX}{\hat{X}}
\newcommand{\hXblack}{\hat{X}}

\newglossaryentry{tildeX}
{
  name={\ensuremath{\tilde{X}^{\leq t}}},
  description={The interpolated observation process $\tilde X^{\leq t}$ continuously interpolates\footnote{Interpolation also includes extrapolation within this paper. $\tilde X^{\leq t}$ interpolates the observations before time $t$ without leaking any information from observations after time $t$. Furthermore, its time-consistency allows for efficient online updates of its signature instead of recomputing its signature for the whole path at every new observation time $\tk{k}$.} the observations of $\gls{X}$ that where observed before time $t$. To be precise the first \gls{dX} coordinates of $\tilde X^{\leq t}$ interpolate the observations of $\gls{X}$, while the next \gls{dX} coordinates of $\tilde X^{\leq t}$ captures explicit information on when which coordinate was observed and the last coordinate is just the time.
  At time $s \in [0,\gls{T}]$, the $j$-th coordinate $\tilde X^{\leq t}_{s,j}\omb$ of $\tilde X^{\leq t}_{s}\omb\in \R^{2\gls{dX}+1}$ is defined in \Cref{sec:Recall: the PD-NJ-ODE}. 
  \inSectionTildeButInOtherSection{$\tilde X^{\leq t}$ is an adapted stochastic process on $(\glsOm{OmFFP} \times \glsOm{tOmFFP}  , \glsF{OmFFP} \otimes \glsF{tOmFFP}, \glsFF{OmFFP} \otimes \glsFF{tOmFFP}, \glsP{OmFFP} \times \glsP{tOmFFP})$}{$\tilde X^{\leq t}$ is an adapted stochastic process on $\gls{OmFFP}$}%
  }
}
\newcommand{\tildeXle}[1]{\ensuremath{\myglslink{tildeX}{\tilde{X}^{\leq #1}}}}
\newcommand{\tildeXleSup}[2]{%
\rlap{\glslink{tildeX}{\phantom{\ensuremath{\tilde{X}^{\leq #1}}}}}\mathlinkcolored{\tilde{X}}^{\mathlinkcolored{\leq #1,} #2}%
}

\newglossaryentry{dk}
{
  name={\ensuremath{d_k}},
  description={A pseudo metric between two c\`adl\`ag $\bA{}$-adapted processes defined in \Cref{def:indistinguishability} in \Cref{sec:Recall: the PD-NJ-ODE} to measure the distance between processes}
}
\makeCommandsForGLSEntryColoredMandatorySub{\dk}{dk}{d}

\newglossaryentry{tu}
{
  name={\ensuremath{\tilde{u}}},
  description={The jump process $\tilde{u}_{t,j} := \sum_{k=0}^{\gls{K}} \gM{k, j} \1_{\tk{k} \leq t}$ counts the coordinate-wise observations (see \Cref{sec:Recall: the PD-NJ-ODE})}
}
\makeCommandsForGLSEntryOptionalSub{\tu}{tu}{\tilde{u}}

\newglossaryentry{u}
{
  name={\ensuremath{u}},
  description={The jump process $u_t := \sum_{k=1}^{\gls{K}} \1_{\tk{k} \leq t}$ counts the observations without considering which coordinates where observed (see \Cref{sec:Recall: the PD-NJ-ODE})}
}
\makeCommandsForGLSEntryOptionalSub{\gu}{u}{u}
\newcommand{\uBlack}{u}

\newglossaryentry{pim}
{
  name={\ensuremath{\pi_m}},
  description={The truncated signature~$\pi_m(\mathbfX{})$ of order $m \in \N$ of a continuous path with finite variation $\mathbfX{}$ is defined in \Cref{def:signature}. In simple words, $\pi_m(\mathbfX{})$ is a finite dimensional feature-vector representing a continuous path~$\mathbfX{}$. For every finite truncation level $m$, $\pi_m(\mathbfX{})$ does not capture all the information about the infinite dimensional object $\mathbfX{}$, but in our proof we use that there always exists a $m\in\N$ such that $\pi_m(\mathbfX{})$ describes $\mathbfX{}$ sufficiently well}
}
\makeCommandsForGLSEntryColoredMandatorySub{\pim}{pim}{\ensuremath{\pi}}

\let\oldPsi\Psi
\newglossaryentry{Psi}
{
  name={\ensuremath{\oldPsi}},
  description={The objective function~$\oldPsi:\bD\to\R$ (cf.\ \emph{equivalent objective function} from Remark~4.7 \& Appendix A.1.4 of \cite{krach2022optimal}) on the path-space $\bD$ in expectation (i.e., \enquote{for an infinite amount of training paths}). 
  \inSectionNoiselessButInOtherSection{the old objective function~\eqref{equ:Psi} that only works for noiseless observations is used}{the new objective function~\eqref{equ:Psi noisy obs} that also works for noisy observations is used}%
  }
}
\renewcommand{\Psi}{\ensuremath{\myglslink{Psi}{\oldPsi}}}

\newglossaryentry{bD}
{
  name={\ensuremath{\mathbb{D}}},
  description={The set of all c\`adl\`ag $\R^{\gls{dX}}$-valued $\bA{}$-adapted processes (see \Cref{sec:Recall: the PD-NJ-ODE}).
  \inSectionTildeButInOtherSection{the stochastic processes in $\mathbb{D}$ live on $(\glsOm{OmFFP} \times \glsOm{tOmFFP}  , \glsF{OmFFP} \otimes \glsF{tOmFFP}, \glsFF{OmFFP} \otimes \glsFF{tOmFFP}, \glsP{OmFFP} \times \glsP{tOmFFP})$}{they live on $\gls{OmFFP}$}}
}
\newcommand{\bD}{\ensuremath{\myglslink{bD}{\mathbb{D}}}}

\newglossaryentry{Y}
{
  name={\ensuremath{Y}},
  description={The output $Y^{\theta}(\tildeXle{\tau(\cdot)})$ of our PD-NJ-ODE~\eqref{equ:PD-NJ-ODE} which should approximate $\hX{}$ for properly trained parameters $\theta$ (see \Cref{def:Sig-NJ-ODE}). We use the short notation $Y^{\theta, j} := Y^{\theta }(\tildeXleSup{\tau(\cdot)}{(j)})$}
}
\makeCommandsForGLSEntryOptionalSub{\gY}{Y}{Y}
\newcommand{\Yblack}{Y}
\newcommand{\Zfilter}{Z}

\let\oldtheta\theta
\newglossaryentry{theta}
{
  name={\ensuremath{\oldtheta}},
  description={The trainable parameters $\oldtheta = (\oldtheta_1, \oldtheta_2, \tildetheta{}_3) \in \Theta$ contain all trainable parameters. I.e., $\oldtheta$ contains $\oldtheta_i = (\tildetheta{}_i, \gamma_i)$ for $i \in \{1,2 \}$ which parameterise the bounded output feedforward neural networks $f_{\oldtheta_1}, \rho_{\oldtheta_2} \in \cN{}$ and $\tildetheta{}_3$ parameterize the feedforward neural network $\tilde g_{\tildetheta{}_3}  \in \tcN{}$. Thus, $\oldtheta = (\oldtheta_1, \oldtheta_2, \tildetheta{}_3) \in \Theta$ parameterize all 3 parameterized functions $f_{\oldtheta_1}, \rho_{\oldtheta_2},  g_{\tildetheta{}_3}$ in our PD-NJ-ODE~\eqref{equ:PD-NJ-ODE} (see \Cref{def:Sig-NJ-ODE})}
}
\renewcommand{\theta}{\ensuremath{\mathglslink{theta}{\oldtheta}}}

\newglossaryentry{tildetheta}
{
  name={\ensuremath{\tilde{\oldtheta}}},
  description={The trainable parameters $\tilde\oldtheta = (\tilde\oldtheta_1, \tilde\oldtheta_2, \tilde \oldtheta_3)$ contain all trainable weights and biases of $\theta = (\theta_1, \theta_2, \tilde \oldtheta_3) \in \Theta$ but not the bounds $\gamma_1$ and $\gamma_2$. I.e., the classical feedforward neural network $\tilde g_{\tilde \oldtheta_3}  \in \tcN{}$ is fully parametrized by $\tilde \oldtheta_3$, while the bounded output feedforward neural networks $f_{\theta_1}, \rho_{\theta_2} \in \cN{}$ are paremetrized by $\theta_i = (\tilde \oldtheta_i, \gamma_i)$ for $i \in \{1,2 \}$ which also includes $\gamma_1$ and $\gamma_2$ additionally to $\tilde\oldtheta_1$ and $\tilde\oldtheta_2$ (see \Cref{def:Sig-NJ-ODE})}
}
\makeCommandsForGLSEntryOptionalSub{\tildetheta}{tildetheta}{\tilde{\oldtheta}}

\let\oldgamma\gamma
\newglossaryentry{gamma}
{
  name={\ensuremath{\oldgamma}},
  description={The parameters $\oldgamma$ are contained in $\theta=\left((\tildetheta{1},\oldgamma_1),(\tildetheta{2},\oldgamma_2),\tildetheta{3}\right)$ and bound the outputs of bounded output neural networks.
  I.e., for every bounded output neural network $f_{\theta_1}, \rho_{\theta_2} \in \cN{}$ it holds that $\left|f_{\theta_1}(x)\right|_2 \leq \oldgamma_1$ and $\left|\rho_{\theta_2}(x)\right|_2 \leq \gamma_2$, because of the bounded output activation function $\Gamma_{\oldgamma_i} : \R^d \to \R^d, x \mapsto \Gamma_{\oldgamma_i}(x)=x \cdot \min\left(1, \frac{\oldgamma_i}{|x|_2}\right)$ (see \Cref{sec:Recall: the PD-NJ-ODE}).
  I.e., $f_{\theta_1}(x)=f_{(\tildetheta{1},\oldgamma_{1})}(x)=\Gamma_{\oldgamma_1}\left(\tilde{f}_{\tildetheta{1}} (x)\right)$ and $\rho_{\theta_2}(x)=\rho_{(\tildetheta{2},\oldgamma_{2})}(x)=\Gamma_{\oldgamma_2}\left(\tilde{\rho}_{\tildetheta{2}} (x)\right)$%
  }
}
\renewcommand{\gamma}{\ensuremath{\mathglslink{gamma}{\oldgamma}}}

\let\oldTheta\Theta
\newglossaryentry{Theta}
{
  name={\ensuremath{\oldTheta}},
  description={The set of all possible trainable parameters $\theta = (\theta_1, \theta_2, \tildetheta{}_3) \in \Theta$ for our our PD-NJ-ODE~\eqref{equ:PD-NJ-ODE} (see \Cref{def:Sig-NJ-ODE})}
}
\renewcommand{\Theta}{\ensuremath{\myglslink{Theta}{\oldTheta}}}

\newglossaryentry{Thetam}
{
  name={\ensuremath{\oldTheta_m}},
  text={\ensuremath{\oldTheta_m}},
  description={The  compact subset $\oldTheta_m\subset\Theta$ consists of all $\theta = \left(\theta_1, \theta_2, \tildetheta{}_3\right) = \left((\tildetheta{}_1,\gamma_1), (\tildetheta{}_2,\gamma_2), \tildetheta{}_3\right)\in \Theta$ that correspond to (bounded output) neural networks with widths and depths that are at most $m$ and such that the truncated signature of level $m$ or smaller is used and such that the norms of the weights $\tildetheta{}_i$ and the bounds $\gamma_i$ are bounded by $m$. I.e., $\oldTheta_m := \{ \theta = ((\tildetheta{}_1, \gamma_1), (\tildetheta{}_2, \gamma_2), \tildetheta{}_3 ) \in \hat\oldTheta_m \, | \, |\tildetheta{}_i|_2 \leq m, \gamma_i \leq m \} \subset \oldTheta_m\subset \Theta$, where $\hat \oldTheta_m \subset \Theta$ is defined as the set of possible parameters for the $3$ (bounded output) neural networks, such that their widths and depths are at most $m$ and such that the truncated signature of level $m$ or smaller is used (see \Cref{sec:Recall: the PD-NJ-ODE}). We use the notation\footnote{\add{The definition is less ambiguous, if we write more precisely: 
  $\oldTheta_m^i:=\left\{ \theta_i \,\middle|\, \exists \left(\theta_1', \theta_2', \tildetheta{}_3'\right)\in\oldTheta_m : \theta_i = \theta_i'\right\}$.}} $\oldTheta_m^i:=\left\{ \theta_i \,\middle|\, \left(\theta_1, \theta_2, \tildetheta{}_3\right)\in\oldTheta_m \right\}$ and  $\tilde{\oldTheta}_m^i:=\left\{ \tildetheta{i} \,\middle|\, \left((\tildetheta{}_1,\gamma_1), (\tildetheta{}_2,\gamma_2), \tildetheta{}_3\right)\in\oldTheta_m \right\}$ for the projections of the sets on the weights $\theta_i$ and $\tildetheta{i}$ respectively}
}
\makeCommandsForGLSEntryColoredMandatorySub{\Thetam}{Thetam}{\ensuremath{\oldTheta}}

\makeCommandsForGLSEntryColoredMandatorySub{\tildeThetam}{Thetam}{\ensuremath{\tilde{\oldTheta}}}

\newglossaryentry{ThetamMin}
{
  name={\ensuremath{\oldtheta^{\min}_m\in\oldTheta_m^{\min}}},
  text={\ensuremath{\oldTheta_m^{\min}}},
  description={The set of all minimizers $\oldtheta^{\min}_m \in \oldTheta_m^{\min} := \argmin_{\theta \in \Thetam{m}}\{ \Phi(\theta) \}$ of the objective function $\Phi$ (corresponding to infinitely many training data) under the constraints of \Thetam{m} for any given $m \in \N$ (see \Cref{thm:1})}
}
\makeCommandsForGLSEntryColoredMandatorySub{\ThetamMin}{ThetamMin}{\ensuremath{\oldTheta^{\min}}}
\makeCommandsForGLSEntryColoredMandatorySub{\thetamMin}{ThetamMin}{\ensuremath{\oldtheta^{\min}}}

\newglossaryentry{ThetamNMin}
{
  name={\ensuremath{\oldtheta^{\min}_{m,N} \in\oldTheta_{m,N}^{\min}}},  text={\ensuremath{\oldTheta_{m,N}^{\min}}},
  description={The set of all minimizers $\oldtheta^{\min}_{m,N} \in \oldTheta_{m}^{\min} := \argmin_{\theta \in \Thetam{m}}\{ \hPhi{N}(\theta) \}$ of the objective function $\hPhi{N}$ (corresponding to $N$ training paths) under the constraints of \Thetam{m} for any given $m,N \in \N$ (see \Cref{thm:1})}
}
\makeCommandsForGLSEntryColoredMandatorySub{\ThetamNMin}{ThetamNMin}{\ensuremath{\oldTheta^{\min}}}
\makeCommandsForGLSEntryColoredMandatorySub{\thetamNMin}{ThetamNMin}{\ensuremath{\oldtheta^{\min}}}
\newcommand{\hatthetamNMin}[1]{\hat\oldtheta^{\min}_{#1}}

\let\oldPhi\Phi
\newglossaryentry{Phi}
{
  name={\ensuremath{\oldPhi}},
  description={The objective function~$\oldPhi:\Theta\to\R, \theta \mapsto \oldPhi(\theta) := \Psi(Y^{\theta}(\gX{}))$ on the parameter-space $\Theta$ in expectation (i.e., \enquote{for an infinite amount of training paths}). 
  \inSectionNoiselessButInOtherSection{the old objective function given in \labelcref{equ:Psi,equ:Phi} that only works for noiseless observations is used}{the new objective function given in \labelcref{equ:Psi noisy obs,equ:Phi noisy obs} that also works for noisy observations is used}. $\oldPhi(\theta)$ is the expected value of $\hPhi{N}(\theta)$, where $\hPhi{N}$ is the loss we actually train on based on $N$ training paths%
  }
}
\renewcommand{\Phi}{\ensuremath{\myglslink{Phi}{\oldPhi}}}

\newglossaryentry{hPhi}
{
  name={\ensuremath{\hat{\oldPhi}_N}},
  description={The objective function~$\hat{\oldPhi}_N:\Theta\to\R, \theta \mapsto \hat{\oldPhi}_N(\theta) := \Psi(Y^{\theta}(\gX{}))$ on the parameter-space $\Theta$ for a finite number of $N\in\N$ training paths  (i.e., the Monte Carlo approximation of $\Phi$). 
  \inSectionNoiselessButInOtherSection{$\hat{\oldPhi}_N$ is defined in \Cref{equ:appr loss function} (this old objective function only leads to the correct results for noiseless observations}{$\hat{\oldPhi}_N$ is defined analogously via \labelcref{equ:Psi noisy obs,equ:Phi noisy obs} (this new objective function is different from the old one and even leads to the right result for noisy observations)}.
  Note that these 3 variants of $\hat{\oldPhi}_N$ are the loss functions on which we actually train our parameters in practice with finitely many training data%
  }
}
\makeCommandsForGLSEntryColoredMandatorySub{\hPhi}{hPhi}{\ensuremath{\hat{\oldPhi}}}

\newglossaryentry{boundedOutputNN}
{
  name={\ensuremath{\mathcal{N}}},
  description={The set $\mathcal{N}$ of bounded output neural networks consists of all feed-forward neural networks that have bounded outputs. In particular we assume that the final activation function applied to the output of the neural network is the bounded output activation function $\Gamma_{\gamma} = (\cdot) \min\left(1, \frac{\gamma}{|(\cdot)|_2}\right)$ (see \Cref{sec:Recall: the PD-NJ-ODE}).
  Throughout the paper we assume that the functions $f_{\theta_1}, \rho_{\theta_2} \in \mathcal{N}$ in the PD-NJ-ODE~\eqref{equ:PD-NJ-ODE} are bounded output feedforward neural networks.
  Note that this assumption is not really important in practice but facilitates our theoretical proof}
}

\newcommand{\cN}{\gls{boundedOutputNN}}

\newglossaryentry{classicalNN}
{
  name={\ensuremath{\tilde{\mathcal{N}}}},
  description={The set $\tilde{\mathcal{N}}$ of feedforward neural networks consists of all (classical) feed-forward neural networks.
  We assume that  $\tilde{\mathcal{N}}$ is a set of standard feedforward neural networks with $\operatorname{id} \in \tilde{\mathcal{N}}$ that satisfies the standard universal approximation theorem with respect to the supremum-norm on compact sets, see for example \citet[Theorem~2]{hornik1991approximation}
  Throughout the paper we assume that the functions $\tilde{g}_{\tildetheta{}_1} \in \tilde{\mathcal{N}}$ in the PD-NJ-ODE~\eqref{equ:PD-NJ-ODE} is a feedforward neural networks}
}

\newcommand{\tcN}{\gls{classicalNN}}

\newglossaryentry{normalDistribution}
{
  name={\ensuremath{\mathscr{N}(\mu,\sigma^2)}},
  text={\ensuremath{\mathscr{N}}},
  description={The normal distribution $\mathscr{N}(\mu,\sigma^2)$ (also known as Gaussian distribution) with mean $\mu$ and standard deviation $\sigma$}
}
\newcommand{\normalDistribution}{\gls{normalDistribution}}

\newglossaryentry{Fj}
{
  name={\ensuremath{F_j}},
  description={The Doob-Dynkin Lemma \citep[Lemma 2]{taraldsen2018optimal} implies the existence of measurable functions $F_j : [0,\gls{T}] \times [0,\gls{T}] \times BV^c([0,\gls{T}]) \to \R$ such that $\hX{}_{t,j} = F_j\left(t, \tau(t), \tildeXle{\tau(t)}\right)$, since $\hX{}_{t,j}:=\E[\gX{t,j}|\At{t}]=\E[\gX{t,j}|\tildeXle{\tau(t)}]$ (see \Cref{sec:Recall: the PD-NJ-ODE}). The goal of our model $\gY{}^{\theta}$ is to learn this function $F=(F_j)_{1\leq j \leq \gls{dX}}$, i.e., $\lim_{m\to\infty}\gY{t}^{\thetamNMin{m,N_m}}\left(\tildeXle{\tau(t)}\right)= F\left(t, \tau(t), \tildeXle{\tau(t)}\right)=\hX{}_{t} $, while we are usually not able to write down $F$ explicitly. In settings with noise (e.g., in \Cref{sec:PD-NJ-ODE with Noisy Observations} and in the appendix) we replace $\tildeXle{\tau(t)}$ by $\tildeOle{\tau(t)}$}
}
\makeCommandsForGLSEntryOptionalSub{\Fj}{Fj}{F}

\newglossaryentry{O}
{
  name={\ensuremath{O}},
  description={The noisy observations $O_{\tk{k}} := \gX{}_{\tk{k}} + \epsilon_k$ for $0 \leq k \leq \gn{}$, where $\epsilon_k$ is the noise with known mean (see \Cref{sec:Setting with Noisy Observations}). For the majority of the paper we assume that the noise has zero mean (whenever we use (the MC-approximation of) the loss~\labelcref{equ:Psi noisy obs,equ:Phi noisy obs}), except \Cref{sec:More General Noise Structure Conditional Moments} where we assume any known mean~$\beta_i(\tilde O^{\leq \tau(t)}) := \E[\epsilon_i | \At{\tk{i}-}]$ and thus use (the MC-approximation of) the loss~\eqref{equ:loss noisy obs generalised}. We define $O_{\tk{i} -} := \gX{}_{\tk{i} -} + \epsilon_i$ and therefore also have that $O_{\tk{i} } = O_{\tk{i} -}$} almost surely.
  \inSectionTildeButInOtherSection[0]{$\epsilon_k$ are i.i.d.\ random variables on $(\glsOm{tOmFFP}, \glsF{tOmFFP}, \glsP{tOmFFP})$}{$\epsilon_k$ are i.i.d.\ random variables on $\OmFP{}$}
}
\makeCommandsForGLSEntryOptionalSub{\gO}{O}{O}

\newglossaryentry{tildeO}
{
  name={\ensuremath{\tilde{O}^{\leq t}}},
  description={The interpolated observation process $\tilde O^{\leq t}$ continuously interpolates the noisy observations $\gO{\tk{i}}$ that where observed before time $t$ (see \Cref{sec:Setting with Noisy Observations}). To be precise the first \gls{dX} coordinates of $\tilde O^{\leq t}$ interpolate the noisy observations $\gO{\tk{i}}$, while the next \gls{dX} coordinates of $\tilde O^{\leq t}$ captures explicit information on when which coordinate was observed and the last coordinate is just the time.
  At time $s \in [0,\gls{T}]$, the interpolated observation process $\tilde O^{\leq t}_{s}\omb\in \R^{2\gls{dX}+1}$ is defined analogously to  $\tilde X^{\leq t}_{s}\omb\in \R^{2\gls{dX}+1}$ from \Cref{sec:Recall: the PD-NJ-ODE}, by replacing $\gls{X}$ by $\gls{O}$. 
  \inSectionTildeButInOtherSection[0]{$\tilde O^{\leq t}$ is an adapted stochastic process on $(\glsOm{OmFFP} \times \glsOm{tOmFFP}  , \glsF{OmFFP} \otimes \glsF{tOmFFP}, \glsFF{OmFFP} \otimes \glsFF{tOmFFP}, \glsP{OmFFP} \times \glsP{tOmFFP})$}{$\tilde O^{\leq t}$ is an adapted stochastic process on $\gls{OmFFP}$}%
  }
}
\newcommand{\tildeOle}[1]{\ensuremath{\myglslink{tildeO}{\tilde{O}^{\leq #1}}}}

\newglossaryentry{hO}
{
  name={\ensuremath{\hat{O}}},
  description={The conditional expectation of $\gls{O}$, which is its $L^2$-optimal prediction \citep[Proposition 2.5]{krach2022optimal} given the currently available information, is defined as $\hat{O}_{\tk{i}-} := \E_{\glsP{OmFFP}\times\glsP{tOmFFP}}[{\gO{}_{\tk{i}-} | \At{\tk{i}-}}]=\hX{\tk{i}-}$ (see the proof of \Cref{lem:L2 identity noisy obs setting}). Only directly at observation times, $\hat{O}_{\tk{i}}$ and $\hX{\tk{i}}$ deviate from each other; i.e., in general $\hX{\tk{i},k}=\E[\gX{\tk{i},k} \, | \, \At{\tk{i}}] \ne \gO{\tk{i},k} =\E[\gO{\tk{i},k} \, | \, \At{\tk{i}}]=\hat{O}_{\tk{i},k}$ if $\gM{i,k}=1$ (see \Cref{sec:PD-NJ-ODE with Noisy Observations}). \inSectionTildeButInOtherSection[0]{the expectation is taken with respect to $\glsP{OmFFP}\times\glsP{tOmFFP}$}{the expectation is taken with respect to $\glsP{OmFFP}$}}
}
\makeCommandsForGLSEntryOptionalSub{\hO}{hO}{\hat{O}}

\newglossaryentry{aaa}
{
  name={\ensuremath{aaaaaaa}},
  description={a a a a a a a a a a a a a a a a a a a a a a a a a a a a a a a a a a a a a a a a a a a a a a a (see \Cref{sec:Recall: the PD-NJ-ODE})}
}

\newglossaryentry{bbb}
{
  name={\ensuremath{bbbb}},
  description={b b b b b b (see \Cref{sec:Recall: the PD-NJ-ODE})}
}

\newcommand{\Rp}{\R_{>0}}
\newcommand{\Rpz}{\R_{\geq0}}
\newcommand{\omb}{(\omega)}
\newcommand{\onenorm}[1]{\left|{#1}\right|_1}

\renewcommand{\l}{\left} 
\renewcommand{\r}{\right}
\newcommand{\ind}[1]{\mathbbm{1}_{#1}}

\makeatletter
\newcommand{\expect}[1]{
  \if@display \mathbb{E} \left\lbrack #1 \right\rbrack
  \else \mathbb{E} \lbrack #1 \rbrack
  \fi
}
\makeatother

\makeatletter
\newcommand{\cexpect}[2]{
  \if@display \expect{#1 \; \middle| \; #2}
  \else \expect{#1 \, | \, #2}
  \fi
}
\makeatother

\makeatletter
\newcommand{\abs}[1]{
  \if@display \left| #1 \right|
  \else | #1 |
  \fi
}
\makeatother

\let\add\undefined
\let\del\undefined
\let\com\undefined

\if\viewchanges1
	\newcommand{\add}[1]{{\color{OliveGreen}{#1}}}
	\newcommand{\del}[1]{{\color{red}{#1}}}
	\newcommand{\com}[1]{{\color{orange}{#1}}}
	
\else
	\newcommand{\add}[1]{#1}
	\newcommand{\del}[1]{}
	\newcommand{\com}[1]{}
	
\fi

\newcommand{\code}{\url{https://github.com/FlorianKrach/PD-NJODE}}

\title{{Extending Path-Dependent NJ-ODEs to Noisy Observations and a Dependent Observation Framework}}

\author{%
    \name William Andersson \email anwillia@ethz.ch\\
    \addr Department of Computer Science\\
    ETH Zurich, Switzerland
    \myAND
    \name Jakob Heiss \email jakob.heiss@math.ethz.ch\\
    \addr Department of Mathematics\\ 
    ETH Zurich, Switzerland
    \myAND
    \name Florian Krach \email florian.krach@math.ethz.ch\\
    \addr Department of Mathematics\\ 
    ETH Zurich, Switzerland
    \myAND
    \name Josef Teichmann \email josef.teichmann@math.ethz.ch\\
    \addr Department of Mathematics\\ 
    ETH Zurich, Switzerland
    }

\providecommand{\keywords}[1]{\textbf{{Keywords:}} \textit{#1}}

\def\mymonth{01}  
\def\myyear{2024} 
\def\openreview{\url{https://openreview.net/forum?id=0T2OTVCCC1}} 

\begin{document}

\maketitle

\begin{abstract}
The \emph{Path-Dependent Neural Jump Ordinary Differential Equation (PD-NJ-ODE)} \citep{krach2022optimal} is a model for predicting continuous-time stochastic processes with irregular and incomplete observations. In particular, the method learns optimal forecasts given irregularly sampled time series of incomplete past observations.
So far the process itself and the coordinate-wise observation times were assumed to be independent and observations were assumed to be noiseless.
In this work we discuss two extensions to lift these restrictions and provide theoretical guarantees as well as empirical examples for them.
In particular, we can lift the assumption of independence by extending the theory to much more realistic settings of conditional independence without any need to change the algorithm. Moreover, we introduce a new loss function, which allows us to deal with noisy observations and explain why the previously used loss function did not lead to a consistent estimator.
\end{abstract}

\if\viewkeywords1
\fi

\section{Introduction}\label{sec:Introduction}
While the online prediction\footnote{With \emph{online} prediction we mean that we use the currently available information to predict until we get new information. As soon as new information becomes available, it is part of the available information and therefore taken into account for subsequent predictions.} of regularly observed or sampled time series is a classical machine learning problem that can be solved with recurrent neural networks (RNNs) as proven e.g.\ by \cite{schafer2006recurrent}, the forecasting of continuous-time processes  with irregular observation has long been an unsolved problem.  
The Neural Jump ODE (NJ-ODE) \citep{herrera2021neural} was the first framework with theoretical guarantees to converge to the optimal prediction in this setting.
However, it was restricted to Markovian It\^o-diffusions with irregular but complete (i.e., all coordinates are observed at the same time) observations. This was heavily generalised with the Path-Dependent NJ-ODE (PD-NJ-ODE) \citep{krach2022optimal}, where the convergence guarantees hold for very general (non-Markovian) stochastic processes with irregular and incomplete observations. Still, the process itself and the observation framework were assumed to be independent and observations were assumed to be noisefree.
In practice both of these assumptions are often unrealistic. E.g., for medical patient data collected at a hospital irregularly over time such as \citet{physionet}, measurements are never noise-free and the decision whether to make a measurement depends on the status of the patient.
Therefore, the focus of this work is to lift those two restrictions. A detailed outline is given below.

\subsection{Related Work}\label{sec:Related Work}
GRU-ODE-Bayes \citep{Brouwer2019GRUODEBayesCM} and the latent ODE \citet{ODERNN2019} both use a model very similar to the NJ-ODE model, however, with different training frameworks. While the latent ODE can only be used for offline forecasting, GRU-ODE-Bayes is applicable to online forecasting as the NJ-ODE. However, in comparison to NJ-ODE, no theoretical guarantees exist for GRU-ODE-Bayes.

Neural controlled differential equations (NCDE) \citep{Kidger2020NeuralCD, morrill2022on} and neural rough differential equations \citep{morrill2021neural} also use similar model frameworks, but their primary objective are labelling problems, i.e., the prediction a classification or regression label for the input of an irregularly sampled time series. For example, based on health parameters of a patient these models try to decide whether the patient will develop a certain disease in the future.

As explained in \citet{krach2022optimal}, PD-NJ-ODEs can be used for stochastic filtering. Another well known model class for this problem are particle filters, also called sequential Monte Carlo methods \citep{maddison2017filtering, le2017auto, corenflos2021differentiable, lai2022variational}. 
Particle filtering methods are applied in the context of state-space models (SSM), which are characterized by a discrete latent Markov process $(\gX{t})_{t=1}^T$ and a discrete observation process $(\Zfilter_t)_{t = 1}^T$ defined on a fixed time-grid. 
Particle filters are used to approximate e.g.\ the conditional distribution of $\gX{t}$ given the observations $(\Zfilter_s)_{1\leq s\leq t}$, or the joint distribution of $(\gX{}_s, \Zfilter_s)_{1 \leq s \leq t}$, for any $t\geq 1$, using weighted sequential Monte Carlo samples. 
In our work, we allow for a much more general setting than the SSM. In particular, we allow for a continuous-time (instead of discrete-time), non-Markovian stochastic process. Since our setting allows for jumps of the process, this also includes the SSM case of a discrete-time Markov process. 
Moreover, in our setting, the underlying process can be observed at random, irregularly sampled, discrete observation times and the framework allows for incomplete observations, where some coordinates might not be observed. 
The primary goal of the PD-NJ-ODE method is to make optimal forecasts for $\gX{t}$ given all observations of $\gX{}$ prior to time $t$. As a special case (since the framework can deal with incomplete observations), this includes the filtering problem, however, in a more general setting, allowing for example to predict $\gX{t}$ while only having (discrete) observations of $\Zfilter$ at randomly sampled observation times until time $s < t$. 

For further related work we refer the interested reader to the respective sections in \cite{herrera2021neural} and \cite{krach2022optimal}.

\subsection{Outline of the Work}\label{sec:Outline}
We introduce two extensions of the PD-NJ-ODE \citep{krach2022optimal} that can be used separately or jointly. To highlight the needed adjustments for each of the extensions, we first recall the setup, model and results from \cite{krach2022optimal} (Section~\ref{sec:Recall: the PD-NJ-ODE}) and then introduce the respective changes in the assumptions and proofs for noisy observations (Section~\ref{sec:PD-NJ-ODE with Noisy Observations}) and dependence between the underlying process and the observation framework (Section~\ref{sec:PD-NJ-ODE with Dependence betweenXand Observation Framework}) separately. 
We focus on re-proving the main results \cite[Theorem~4.1 and Theorem~4.4]{krach2022optimal} in the new settings, by giving the arguments which need to be adjusted while skipping those which remain unchanged.
In Appendix~\ref{sec:Combining the Two Extensions With Full Proof} we give the full proof for the most general result with both extensions, making the paper self-contained.
We remark here that also the results for the conditional variance and for stochastic filtering \cite[Section~5 and 6]{krach2022optimal} follow in these extended settings similarly as the main results. Due to the similarity, we do not elaborate on this but leave the details to the interested reader.
In Section~\ref{sec:Practical Implications of the Convergence Result} we discuss the practical implications of our main convergence result.
Finally, in Section~\ref{sec:Experiments} we show empirically that the PD-NJ-ODE performs well in these generalised settings.

\section{Recall: the PD-NJ-ODE}\label{sec:Recall: the PD-NJ-ODE}
In this section we recall the PD-NJ-ODE framework together with the main result from \citet{krach2022optimal}. 
The PD-NJ-ODE is a prediction model that can be used to learn the $L^2$-optimal prediction of a stochastic process (continuously in time), given its discrete, irregular and incomplete observations in the past. 
This means that the PD-NJ-ODE learns to compute the conditional expectation, which is the $L^2$-optimal prediction. Importantly, only data samples are needed to train the model. In particular, no knowledge about the dynamics of the underlying process is needed.
We first given an intuitive example for the application of this model, which will be reused throughout the paper, and then discuss its technical details.

\subsection{Intuitive Example Application}
Suppose we have a dataset of $N$ patients $1\leq j \leq N$. Any patient $j$ has $\gls{dX}$ medical values $\left(\gX{t, k}^{(j)}\right)_{1\leq k \leq \gls{dX}}$ (such as body temperature or blood pressure) at any time $t\in[0,\gls{T}]$. However, we only have (noisy) measurements of some of these coordinates at some irregular times $\tk{i}^{(j)}$ (e.g., we measure the body temperature on one day and the blood pressure on another day and in-between we do not measure anything). Our training dataset consists of all these (noisy) measurements including their time-stamps. Based on this training dataset we train a PD-NJ-ODE which then allows us to make online forecasts for new patients based on their noisy incomplete irregularly observed measurements. For example the Physionet dataset \citep{physionet} is exactly such a dataset, but of course our method can also be applied in many other situations.

\subsection{Technical Background and Mathematical Notation}
We start by recalling the most relevant parts of the problem setting of the PD-NJ-ODE. For more details, please refer to \citet{krach2022optimal}.

For $\gls{dX} \in \N$ and $\gls{T} > 0$ we consider a filtered probability space  $\glsLong{OmFFP}$ with an adapted c\`adl\`ag  stochastic process $\gX{} :={(\gX{t})}_{t \in [0,\gls{T}]}$ taking values in $\R^{\gls{dX}}$. 
The main goal in this paper is the predict $\gX{}$ optimally for future times, based on discrete observations of it in the past.
We denote its running maximum process by \gls{Xs} (i.e., $\glst{Xs}:=\sup_{s \in [0,t]}\onenorm{\gX{s}}$) and the random set of its jump times by $\gls{J}$. The \emph{random observation framework} is defined independently of $\gX{}$ on another filtered probability space  $\glsLong{tOmFFP}$ by
\begin{itemize}
\item $\gn{}: \glsOm{tOmFFP}  \to \N_{\geq 0}$, an $\glsF{tOmFFP} $-measurable random variable, the random number of observations, 
\item $\gls{K} := \sup \left\{k \in \N \, | \, \glsP{tOmFFP}(\gn{} \geq k) > 0 \right\} \in \N \cup\{\infty\}$, the maximal value of $\gn{}$, 
\item  $\tk{i}: \glsOm{tOmFFP}  \to [0,\gls{T}] \cup \{ \infty \}$ for $0 \leq i \leq \gls{K}$, {sorted} stopping times, which are the random observation times, with $\tk{i}(\tilde{\omega}) := \infty$ if $\gn{}(\tilde{\omega}) < i$ and $\tk{i} < \tk{i+1}$ for all $0 \leq i < \gn{}$,
\item $\tau : [0,\gls{T}] \times \glsOm{tOmFFP}  \to [0,\gls{T}], \; (t, \tilde\omega) \mapsto \tau(t, \tilde\omega) := \max\{ \tk{i}(\tilde\omega) | 0 \leq i \leq \gn{}(\tilde\omega), \tk{i}(\tilde\omega) \leq t \}$, the  last observation time before a certain time $t$, and
\item $\gM{} = (\gM{k})_{0 \leq k \leq \gls{K}}$, the observation mask, which is a sequence of random variables on  $(\glsOm{tOmFFP} , \glsF{tOmFFP}, \glsP{tOmFFP} )$ taking values in $\{ 0,1 \}^{\gls{dX}}$ such that $\gM{}_k$ is $\tFt{\tk{k}}$-measurable.
The $j$-th coordinate of the $k$-th element of the sequence $\gM{}$, i.e., $\gM{k,j}$, signals whether $\gX{}_{\tk{k}, j}$, denoting the $j$-th coordinate of the stochastic process at observation time $\tk{k}$ is observed. By abuse of notation we also write $\gM{\tk{k}} := \gM{k}$. 
\end{itemize}
In the following we consider the filtered product probability space $(\glsOm{OmFFP} \times \glsOm{tOmFFP}  , \glsF{OmFFP} \otimes \glsF{tOmFFP}, \glsFF{OmFFP} \otimes \glsFF{tOmFFP}, \glsP{OmFFP} \times \glsP{tOmFFP})$ and the filtration of the currently available information $\bA{} := (\At{t})_{t \in [0,\gls{T}]}$ defined by 
\begin{equation*}
\At{t} := \boldsymbol{\sigma}\left(\gX{}_{\tk{i}, j}, \tk{i}, \gM{\tk{i}} | \tk{i} \leq t,\, j \in \{1 \leq l \leq \gls{dX} | \gM{\tk{i}, l} = 1  \} \right),
\end{equation*} 
where $\boldsymbol\sigma(\cdot)$ denotes the generated $\sigma$-algebra. 
We note that $\At{t} = \At{\tau(t)}$ for all $t \in [0,\gls{T}]$.
The conditional expectation process of $\gX{}$, which is its $L^2$-optimal prediction \citep[Proposition 2.5]{krach2022optimal} and therefore the process we would like to compute, is defined as $\hX{} = (\hX{t})_{0 \leq t \leq \gls{T}}$, with $\hX{}_t := \E_{\glsP{OmFFP}\times\glsP{tOmFFP}}[\gX{t} | \At{t}]$.
Even though this solution to our prediction problem is abstractly well understood, it is not clear how to actually compute it, especially if one only observes data samples without knowing the distribution of $\gX{}$. 
The PD-NJ-ODE is designed to do exactly this.
To compute the conditional expectation, it uses the powerful framework of signatures to convert all information available from the past observations into a tractable feature-vector. This is needed, since the underlying process is not assumed to be Markovian. Since the signature is defined for (continuous) paths, the first step is to construct a continuous interpolation of the discrete observations, the \emph{interpolated observation process}, in such a way that it carries the same information as the discrete observations.
In particular, for any $0 \leq t \leq \gls{T}$ the $j$-th coordinate of the interpolated observation process $\tildeXle{t} \in \R^{2\gls{dX}+1}$  at time $0 \leq s \leq \gls{T}$ is defined by
\begin{equation*}
\tildeXle{t}_{s,j} := \begin{cases}
	\gX{}_{\tk{a(s,t)},j} \frac{\tk{b(s,t)} - s}{\tk{b(s,t)} - \tk{b(s,t)-1}} + \gX{}_{\tk{b(s,t)},j} \frac{s - \tk{b(s,t)-1}}{\tk{b(s,t)} - \tk{b(s,t)-1}}, & \text{if }  \tk{b(s,t)-1}< s \leq \tk{b(s,t)} \text{ and}  \\ & 1 \leq j \leq \gls{dX}, \\
	\gX{}_{\tk{a(s,t)},j}, & \text{if } s \leq \tk{b(s,t)-1}\text{ and } 1 \leq j \leq \gls{dX}, \\
    \tu{\tk{a(s,t)},j-\gls{dX}}  +  \frac{s - \tk{b(s,t)-1}}{\tk{b(s,t)} - \tk{b(s,t)-1}}, & \text{if }  \tk{b(s,t)-1}< s \leq \tk{b(s,t)} \text{ and } \\ & \gls{dX} < j \leq 2 \gls{dX}, \\
	\tu{\tk{a(s,t)},j-\gls{dX}}, & \text{if } s \leq \tk{b(s,t)-1}\text{ and }    \gls{dX} < j \leq 2 \gls{dX} , \\
    s, & \text{if } j = 2 \gls{dX}+1,
\end{cases}
\end{equation*}
where $\tu{t,j} := \sum_{k=0}^{\gls{K}} \gM{k, j} \1_{\tk{k} \leq t}$ is the jump process that counts the coordinate-wise observations and
\begin{equation*}
\begin{split}
a(s,t) &:= a(s,t, j):=  \max\{ 0 \leq a \leq \gn{} \vert \tk{a} \leq \min(s,t), \gM{\tk{a},j}  = 1 \}, \\
b(s,t) &:= b(s,t,j) :=  \inf\{ 1 \leq b \leq \gn{} \vert s \leq \tk{b} \leq t, \gM{\tk{b},j}  = 1 \},
\end{split}
\end{equation*}
with $\tk{\infty} := \infty$. 
Simply put, $\tildeXle{t}$ is a continuous version (without information leakage and with time-consistency) of the rectilinear interpolation of the observations of $\gX{}$ and of $\tu{}$.
The paths of $\tildeXle{t}$ belong to $BV^c([0,\gls{T}])$, the set of continuous $\R^{\gls{dX}}$-valued paths of bounded variation on $[0,\gls{T}]$.
Most importantly, the way it is defined ensures that no information about the next observation is leaked (through the forward-looking interpolation) until after the last observation time prior to it and that $\tildeXle{t}$ is $\At{t}$-measurable. Moreover, it is time-consistent in the sense that for all $r \geq t$ and $s \leq \tau(t)$ we have $\tildeXle{t}_s = \tildeXle{r}_s$.
Clearly, $\gX{}_{\tk{i}, j}, \tk{i}, \gM{\tk{i}}$ can be reconstructed from the coordinates of $\tildeXle{t}$ for all $\tk{i} \leq t$ and $j \in \{1 \leq l \leq \gls{dX} | \gM{\tk{i}, l} = 1  \}$, hence, $\At{t} = \sigmab(\tildeXle{t})$. Moreover, it is easy to see that $\tildeXle{t} = \tildeXle{\tau(t)}$, consequently $\hX{t}$ is $\sigmab(\tildeXle{\tau(t)})$-measurable.
Therefore, the Doob-Dynkin Lemma \citep[Lemma 2]{taraldsen2018optimal} implies the existence of measurable functions $\Fj{j} : [0,\gls{T}] \times [0,\gls{T}] \times BV^c([0,\gls{T}]) \to \R$ such that $\hX{}_{t,j} = \Fj{j}(t, \tau(t), \tildeXle{\tau(t)}) $. 
In \citet{krach2022optimal} convergence of the PD-NJ-ODE model to the conditional expectation $\hX{}$  was shown under the following assumptions. 
\begin{assumption} \label{assumption:F}
We assume that: 
\begin{enumerate}[label=(\roman*)]
\item For every $1\leq k, l \leq \gls{K}$, $\gM{}_k$ is independent of  $\tk{l}$ and of $\gn{}$,  $\glsP{tOmFFP} (\gM{k,j} =1 ) > 0$ and $\gM{0,j}=1$ for all  $1 \leq j \leq \gls{dX}$  (every coordinate can be observed at any observation time and $\gX{}$ is completely observed at $0$) and $|\gM{}_k|_1 > 0$ for every $1 \leq k \leq \gls{K}$ $\glsP{tOmFFP}$-almost surely (at every observation time at least one coordinate is observed). \label{ass_main_1}
\item The probability that any two observation times are closer than $\epsilon>0$ converges to $0$ when $\epsilon$ does, i.e., if $\delta(\tilde \omega) := \min_{0 \leq i < \gn{}(\tilde \omega)} |\tk{i+1}(\tilde \omega) - \tk{i}(\tilde \omega)|$ then $\lim_{\epsilon \to 0} \glsP{tOmFFP} (\delta < \epsilon) = 0$.\label{ass_main_2}
\item Almost surely $\gX{}$ is  not observed at a jump, i.e., $(\glsP{OmFFP} \times \glsP{tOmFFP})( \tk{i} \in \gls{J} | i \leq \gn{} ) = (\glsP{OmFFP} \times \glsP{tOmFFP})( \Delta \gX{}_{\tk{i}} \neq 0 | i \leq \gn{}) = 0$ for all $ 1 \leq i \leq \gls{K}$. \label{ass_main_3}
\item $\Fj{j}$ are  continuous and differentiable in their first coordinate $t$ such that their partial derivatives with respect to $t$, denoted by $f_j$, are again continuous and there exists a $B >0$ and $p \in \N_{\geq 1}$ such that for every $t \in [0,\gls{T}]$ the functions $f_j, \Fj{j}$ are polynomially bounded in $\gls{Xs}$, i.e., 
\[|\Fj{j}(\tau(t), \tau(t), \tildeXle{\tau(t)})| +  | f_j(t, \tau(t), \tildeXle{\tau(t)})  | \leq B (\glst{Xs} +1)^p . \] \label{ass_main_4}
\item $\gls{Xs}$ is $L^{2 p}$-integrable, i.e., $\E[(\glsT{Xs})^{2 p}] < \infty$. \label{ass_main_5}
\item The random number of observations $\gn{}$ is integrable, i.e., $\E_{\glsP{tOmFFP}}[\gn{}] < \infty$. \label{ass_main_6}
\end{enumerate}
\end{assumption} 
\begin{rem}\label{rem:at least one coord observed not needed}
    The assumption that $|\gM{}_k|_1 > 0$ for every $1 \leq k \leq \gls{K}$ $\glsP{tOmFFP}$-almost surely is not needed in the proof. It is just added such that the denotation ``observation time'' is meaningful. When removing it, some of the observation times might be ``pseudo'' observation times, where no coordinate is actually observed.
\end{rem}

\begin{rem}\label{rem:ass2 not needed}
    It turns out that \Cref{assumption:F}~\ref{ass_main_2} is always satisfied in the described setting, hence, we can also leave it away.
    Indeed, $\lim_{\epsilon \to 0} \glsP{tOmFFP} (\delta < \epsilon) = 0$ holds, if $\glsP{tOmFFP} (\delta = 0) = 0$, which is satisfied by  definition of the stopping times to be in strictly increasing order.
\end{rem}
Moreover, relaxations on the assumption of observing $\gX{}_0$ completely were discussed in \citet[Remark 2.2]{krach2022optimal}.
Convergence is defined with respect to the following distance.
\begin{definition}\label{def:indistinguishability}
    Let $c_0 := c_0(k) := (\glsP{tOmFFP} (\gn{}\geq k))^{-1}$.
    A distance between c\`adl\`ag $\bA{}$-adapted processes $Z, \xi  : [0,\gls{T}] \times (\Om{} \times \glsOm{tOmFFP}) \to \R^{r}$ is defined through the pseudo metrics 
\begin{equation}\label{equ:pseudo metric dk}
    \dk{k} (Z, \xi) = c_0(k)\,  \E_{\glsP{OmFFP} \times \glsP{tOmFFP}}\left[ \1_{\{\gn{} \geq k\}} | Z_{\tk{k}-} - \xi_{\tk{k}-} |_2 \right], 
\end{equation}
for $1 \leq k \leq \gls{K}$ and two processes are called indistinguishable, if $\dk{k} (Z, \xi) = 0$ for all $1 \leq k \leq \gls{K}$.
\end{definition}
In particular, this distance compares two c\`adl\`ag processes at their left limits\footnote{%
For a c\`adl\`ag process $Z$ we have (by right-continuity) that $Z_t = \lim_{\epsilon \downarrow 0} Z_{t+\epsilon}$, while the left limit $Z_{t-} = \lim_{\epsilon \downarrow 0} Z_{t-\epsilon}$ exists but does not need to coincide with $Z_t$. In particular, $Z_t = Z_{t-}$ if and only if the jump of the process at $t$, $\Delta Z_t = Z_t - Z_{t-}$, is $0$.} %
at the observation times $\tk{k}$.

The path-dependent generalisation of the Neural Jump ODE model \citep{herrera2021neural} uses the truncated signature transformation $\pim{m}$ \citep[Definition 3.4]{krach2022optimal}.
The signature of a path is an (infinite) tuple of features, which allows to approximate any continuous function of the path by a linear combination of the signature terms. Therefore, it is a very powerful framework in learning theory. It is defined as follows.
\begin{definition}\label{def:signature}
    Let $J$ be a closed interval in $\R$ and $\mathbfX{}:J\rightarrow\R^{d}$ be a continuous path with finite variation.
    The signature of $\mathbfX{}$ is defined as the collection of iterated integrals
    \begin{equation*}
        S(\mathbfX{}) = \left(1, \mathbfX{}_J^1, \mathbfX{}_J^2, \dots\right),
    \end{equation*}
    where, for each $m\geq 1$,
    \begin{equation*}
        \mathbfX{}_J^m = \int_{\substack{\uBlack{}_1<\dots<\uBlack{}_m \\ \uBlack{}_1,\dots,\uBlack{}_m\in J}} d\mathbfX{}_{\uBlack{}_1}\otimes\dots\otimes d\mathbfX{}_{\uBlack{}_m} \in (\R^d)^{\otimes m} .
    \end{equation*}
    The  truncated signature of $\mathbfX{}$ of order $m \in \N$ is defined as 
    \begin{equation*}
        \pim{m}(\mathbfX{}) = \left(1,\mathbfX{}_J^1,\mathbfX{}_J^2,\dots,\mathbfX{}_J^m\right),
    \end{equation*}
    i.e., the first $m+1$ terms (levels) of the signature of $\mathbfX{}$. 
    For a path of dimension $d$, the dimension of the truncated signature of order $m$ is
\begin{equation*}\label{eq:sig_nb_terms}
\begin{cases}
m+1, & \text{if } d =1, \\
 \frac{d^{m+1}-1}{d-1}, & \text{if } d >1. 
\end{cases}   
\end{equation*}
\end{definition}
Moreover, PD-NJ-ODE uses bounded output neural networks $f_{(\tildetheta{}, \gamma)}$, where $\tildetheta{}$ are the weights of the standard neural network and $\gamma > 0$ is the trainable parameter of the bounded output activation function \citep[Definition 3.12]{krach2022optimal}
\begin{equation*}
\Gamma_{\gamma} : \R^d \to \R^d, x \mapsto x \cdot \min\left(1, \frac{\gamma}{|x|_2}\right),
\end{equation*}
applied to the output of the standard neural network. By $\cN{}$ we denote the set of all bounded output neural networks based on a set $\tcN{}$ of standard neural networks. 
In the following we assume that  $\tcN{}$ is a set of standard feedforward neural networks with Lipschitz continuous activation functions, with $\operatorname{id} \in \tcN{}$ that satisfies the standard universal approximation theorem with respect to the supremum-norm on compact sets, see for example \citet[Theorem~2]{hornik1991approximation}.

\begin{definition}\label{def:Sig-NJ-ODE}
    The  \textit{Path-Dependent Neural Jump ODE (PD-NJ-ODE)} model is given by
    \begin{equation}\label{equ:PD-NJ-ODE}
    \begin{split}
    H_0 &= \rho_{\theta_2}\left(0, 0, \pim{m} (0), \gX{}_0 \right), \\
    dH_t &= f_{\theta_1}\left(H_{t-}, t, \tau(t), \pim{m} (\tildeXle{\tau(t)} -\gX{}_0 ), \gX{}_0 \right) dt  \\
    & \quad + \left( \rho_{\theta_2}\left( H_{t-}, t, \pim{m} (\tildeXle{\tau(t)}-\gX{}_0 ), \gX{}_0 \right) - H_{t-} \right) d\gu{}_t, \\
    \gY{}_t &= \tilde g_{\tildetheta{}_3}(H_t).
    \end{split}
    \end{equation}
    The functions $f_{\theta_1}, \rho_{\theta_2} \in \cN{}$ are bounded output feedforward neural networks and $\tilde g_{\tildetheta{}_3} \in \tcN{}$ is a feedforward neural network  with trainable parameters $\theta = (\theta_1, \theta_2, \tildetheta{}_3) \in \Theta$, where $\theta_i = (\tildetheta{}_i, \gamma_i)$ for $i \in \{1,2 \}$ and $\Theta$ is the set of all possible weights for the PD-NJ-ODE; $m \in \N$ is the signature truncation level and $\gu{}$ is the jump process counting the observations defined as $\gu{}_t := \sum_{k=1}^{\gls{K}} \1_{\tk{k} \leq t}$.
\end{definition}
In the PD-NJ-ODE model, $H_t$ has the role of a latent process and $\gY{t}$ is the model output (similar to the latent variable and model output of an RNN). The goal of the model output $\gY{t}$ is to approximate the true conditional expectation $\hX{t}$. $H_t$ evolves continuously between any two observations, according to the \emph{neural ODE} defined by $f_{\theta_1}$. At an observation time, when there is a ``jump'' in the available information, $H_t$ jumps according to $\rho_{\theta_2}$, which can be interpreted as an \emph{RNN-cell}. The latent process $H_t$ is mapped to the output process $\gY{t}$ through the \emph{readout map} $\tilde{g}_{\tildetheta{3}}$. 
The existence and uniqueness of a solution $(H,\gls{Y})$ of \eqref{equ:PD-NJ-ODE} for a fixed $\theta$ is implied by \citet[Theorem~16.3.11]{cohen2015stochastic}\footnote{$f_{\theta_1}$ and $\rho_{\theta_2}$ are Lipschitz continuous as neural networks with Lipschitz continuous activation functions, hence, the stochastic functions $((\omega, \tilde{\omega}), t, H) \mapsto f_{\theta_1}(H_{t-}, t, \tau(t, \tilde{\omega}), \pim{m} (\tildeXle{\tau(t)}-\gX{}_0 )(\omega, \tilde{\omega}), \gX{}_0(\omega))$, and similarly for $\rho_{\theta_2}$, are uniformly Lipschitz according to \citet[Definition~16.3.2]{cohen2015stochastic}. Moreover, it is immediate to see that these functions are \emph{coefficients} according to \citet[Definition~16.0.3]{cohen2015stochastic}, since they are continuous, hence predictable, and since we integrate with respect to finite variation processes. In particular, integrability \citep[Definition~12.3.10]{cohen2015stochastic} is trivially satisfied (path-wise), because any continuous bounded function is Stieltjes-integrable.}. To emphasize the dependence of the PD-NJ-ODE output $\gY{}$ on $\theta$ and $\gX{},(\tk{i})_{1\leq i\leq K}$ and $\gM{}$ we write $\gY{}^{\theta}(\tildeXle{\tau(\cdot)})$ (since $\tildeXle{\tau(\cdot)}$ summarizes $\gX{},(\tk{i})_{1\leq i\leq K}$ and $\gM{}$).

The objective function (cf.\ \emph{equivalent objective function} from Remark~4.8 \& Appendix D.1.5 of \cite{krach2022optimal}) for the training of the PD-NJ-ODE is defined as
\begin{align}
\Psi: \, &\bD{} \to \R, \; Z \mapsto \Psi(Z) := \E_{\glsP{OmFFP}\times\glsP{tOmFFP}}\left[ \frac{1}{\gn{}} \sum_{i=1}^{\gn{}}  \left(  \left\lvert \gM{}_i \odot ( \gX{}_{\tk{i}} - Z_{\tk{i}} ) \right\rvert_2 + \left\lvert \gM{}_i \odot (\gX{}_{\tk{i}} - Z_{\tk{i}-} ) \right\rvert_2 \right)^2 \right], \label{equ:Psi} \\
\Phi : \, &\Theta \to \R, \; \theta \mapsto \Phi(\theta) := \Psi(\gY{}^{\theta}(\tildeXle{\tau(\cdot)})), \label{equ:Phi}
\end{align}
where $\odot$ is the element-wise multiplication (Hadamard product) and $\bD{}$ the set of all c\`adl\`ag $\R^{\gls{dX}}$-valued $\bA{}$-adapted processes on the product probability space $\Om{} \times \glsOm{tOmFFP} $.
The two terms in \eqref{equ:Phi} compare the distances between $\gX{\tk{k}}$ and $Z$ after and before its jump at observation times $\tk{k}$. Intuitively speaking, to minimize the second term, $Z_{\tk{k}-}$ has to be the best possible prediction of $\gX{\tk{k}}$ given the information available before the new observation becomes available; and to minimize the first term, $Z$ has to jump to the new observation $\gX{\tk{k}}$ after it became available.

For $N \in \N$ the number of training paths and for every $1 \leq j \leq N$, let $(\gX{}^{(j)}, \gM{}^{(j)}, \gn{(j)}, \tk{1}^{(j)}, \dotsc, \tksup{\gn{(j)}}{(j)}) \sim (\gX{}, \gM{}, \gn{}, \tk{1}, \dotsc, \tk{\gn{}})$ be independent and identically distributed (i.i.d.) random processes and variables with the same distribution as our  initially introduced process $\gX{}$ together with its random observation framework.
Then the Monte Carlo approximation of \eqref{equ:Phi} is 
\begin{equation}\label{equ:appr loss function}
\hPhi{N}(\theta) := \frac{1}{N} \sum_{j=1}^N  \frac{1}{\gn{(j)}}\sum_{i=1}^{\gn{(j)}} \left(  \left\lvert \gM{i}^{(j)} \odot \left( \gX{}_{\tksup{i}{(j)}}^{(j)} - \gY{}_{\tksup{i}{(j)}}^{\theta, j } \right) \right\rvert_2 + \left\lvert \gM{i}^{(j)} \odot \left( \gX{}_{\tksup{i}{(j)}}^{(j)} - \gY{}_{\tksup{i}{(j)}-}^{\theta, j } \right) \right\rvert_2 \right)^2,
\end{equation}
where $\gY{}^{\theta, j} := \gY{}^{\theta }(\tildeXleSup{\tau(\cdot)}{(j)})$.

Based on these loss functions, the following convergence guarantees can be derived, where $\Thetam{m} \subset \Theta$ is defined as the set of possible parameters for the $3$ (bounded output) neural networks, such that their widths and depths are at most $m$ and such that the truncated signature of level $m$ or smaller is used and such that the norms of the weights $\tildetheta{}_i$ and the bounds $\gamma_i$ are bounded by $m$ (i.e., $|\tildetheta{}_i|_2 \leq m, \gamma_i \leq m$). Thus, $\Thetam{m}$ is a compact subset of $\Theta$.

\begin{theorem}\label{thm:1}
Let $\thetamMin{m} \in \ThetamMin{m} := \argmin_{\theta \in \Thetam{m}}\{ \Phi(\theta) \}$ for every $m \in \N$. If Assumption~\ref{assumption:F} is satisfied, then, for $m \to \infty$, the value of the loss function $\Phi$ converges to the minimal value of $\Psi$  which is uniquely achieved by $\hX{}$ up to indistinguishability, i.e.,
\begin{equation*}
\Phi(\thetamMin{m}) \xrightarrow{m \to \infty} \min_{Z \in \bD{}} \Psi(Z) = \Psi(\hX{}).
\end{equation*}
Furthermore, for every $1 \leq k \leq \gls{K}$ we have that $\gY{}^{\thetamMin{m}}$ converges to $\hX{}$ in the metric $\dk{k}$  as $m \to \infty$.

Let $\thetamNMin{m,N} \in \ThetamNMin{m,N} := \argmin_{\theta \in \Thetam{m}}\{ \hPhi{N}(\theta)\}$ for every $m, N \in \N$. 
Then, for every $m \in \N$, $(\glsP{OmFFP}\times\glsP{tOmFFP})$-a.s. 
\begin{equation*}
\hPhi{N} \xrightarrow{N \to \infty} \Phi \quad \text{uniformly on } \Thetam{m}.
\end{equation*}
Moreover, for every $m \in \N$, we have a.s.,
\begin{equation*}
\Phi(\thetamNMin{m,N}) \xrightarrow{N \to \infty} \Phi(\thetamMin{m}) \quad \text{and} \quad \hPhi{N}(\thetamNMin{m,N}) \xrightarrow{N \to \infty} \Phi(\thetamMin{m}).
\end{equation*}
In particular, one can define an increasing random sequence $(N_m)_{m \in \N}$ in $\N$ such that for every $1 \leq k \leq \gls{K}$ we have a.s.\ that $\gY{}^{\thetamNMin{m,N_{m}}}$ converges to $\hX{}$ 
in the metric $\dk{k}$  as $m \to \infty$.
\end{theorem}
This result shows that, given we find the minimizer~$\thetamNMin{m,N_m}$ of the loss function, the output~$\gY{}^{\thetamNMin{m,N_m}}$ of the PD-NJ-ODE model converges to the optimal prediction, i.e., to the true conditional expectation $\hX{}$, in the metrics $\gls{dk}$. This convergence holds for almost every realization of the training data, which is used to derive the minimizer~$\thetamNMin{m,N_m}$, when evaluated on independent test data.
In particular, this result verifies that the model can approximate $\hX{}$ arbitrarily well and that minimizing the loss function is expedient to find such an approximation of $\hX{}$. 
In Section~\ref{sec:Practical Implications of the Convergence Result}, we further discuss the practical implication of this convergence result.
In this work, we do not focus on the task of finding the minimizer for the loss function, which is an independent and well studied problem on its own. Different optimization schemes exists, which yield convergence to global or local optima. These can be combined with our results as was further discussed in \citet[Appendix~E.2]{herrera2021neural}. In our experiments we use Adam \citep{adam}, an stochastic gradient descent (SGD) version, which yields good empirical results.

\section{PD-NJ-ODE with Noisy Observations}\label{sec:PD-NJ-ODE with Noisy Observations}
So far the PD-NJ-ODE model was only applicable when having noise-free observations of $\gX{}$. In particular, we demonstrate that the PD-NJ-ODE with the original loss yields incorrect predictions in the presence of measurement noise, i.e., i.i.d.\ noise terms that are added to each observation of $\gX{}$. 
Using the stochastic filtering approach described in \citet[Section~6]{krach2022optimal} would be a possibility to include (discrete or continuous) noise. However, this requires strong additional assumptions (i.e., the knowledge of the joint distribution of the process and the noise or equivalently training samples split up into the noise-free observations and the noise terms) which are not satisfied easily. Therefore, we want to adapt our framework, such that it can be applied to noisy observations, while only imposing weak assumptions that are easily satisfied. 
In particular, we introduce a method that can deal with noisy observations, even if we have never seen any noise-free observation during training.  In contrast to classical filtering, there is no need to have a strong prior knowledge of the underlying dynamics.

In this section, we introduce observation noise (e.g., measurement-noise), i.e., i.i.d.\ noise terms $\epsilon_i$ that are added to the process $\gX{}$ at each observation time $\tk{i}$, leading to the noisy observations $\gO{\tk{i}} := \gX{}_{\tk{i}} + \epsilon_i$. Even though we only observe the $\gO{\tk{i}}$, the goal still is to predict $\gX{}$, in particular to compute $\E[\gX{t} \, | \, \gO{\tk{0}},\dots,\gO{\tau(t)}]$.

Inspecting the PD-NJ-ODE model and its loss function (replacing $\gX{}$ by $\gO{}$), we notice two things. First, that nothing in the model's architecture prevents it from learning this modified objective, which should still have the same properties. Second, that the loss function needs to be modified. Indeed, the first term of the loss function would train the model to jump to the noisy observation $\gO{\tk{i}}$. This would be incorrect, since in general $\E[\gX{}_{\tk{i}} \, | \, \gO{\tk{0}}, \dotsc, \gO{\tk{i}}] \ne \gO{\tk{i}}$ because the conditional expectation of $\gls{X}$ filters out the noise as well as possible. We therefore drop the first term of the loss function.

On the other hand, it is easy to see that the conditional expectations of $\gX{}$ and $\gO{}$ coincide in between observation times\footnote{Note that, in general we have $\E[\gX{}_{\tk{i}} \, | \, \gO{\tk{0}}, \dotsc, \gO{\tk{i}}]\neq\E[\gO{\tk{i}} \, | \, \gO{\tk{0}}, \dotsc, \gO{\tk{i}}]=\gO{\tk{i}}$ at observation times, in contrast to their equality between observation times.} if the observation noise $\epsilon_i$ is independent of the observations and has mean $0$. Therefore, the second term of the loss function is minimised if the model learns the conditional expectation $\E[\gX{}_{t} \, | \, \gO{\tk{0}}, \dotsc, \gO{\tau(t)}]$ between observation times.

Along these lines, it turns out that it suffices to omit the first term of the loss function to recover the original results of Theorem~\ref{thm:1} under noisy observations. In particular, to optimize the loss, the model learns to jump to the conditional expectation of $\gX{}$ at observation times even without the respective loss term. Indeed, since it evolves continuously after an observation, it would otherwise be different from the optimal prediction right after the observation time and therefore would not optimize the loss. 
In the following this is formalised.

\subsection{Setting with Noisy Observations}\label{sec:Setting with Noisy Observations}
The process $\gX{}$ as well as the $\gn{}, \gls{K}, \tk{i}, \tau, \gM{}$ are defined as in Section~\ref{sec:Recall: the PD-NJ-ODE}. Additionally, we define
\begin{itemize}
    \item $(\epsilon_k)_{0 \leq k \leq \gls{K}}$, the observation noise, which is a sequence of i.i.d.\ random variables on $(\glsOm{tOmFFP} , \glsF{tOmFFP}, \glsP{tOmFFP} )$ taking values in $\R^{\gls{dX}}$,
    \item $\gO{\tk{k}} := \gX{}_{\tk{k}} + \epsilon_k$ for $0 \leq k \leq \gn{}$, the noisy observation sequence.
\end{itemize}
Since the goal is to predict $\gX{}$ given the observations $\gO{\tk{i}}$, we redefine the filtration of the currently available information via 
\begin{equation*}
\At{t} := \boldsymbol{\sigma}\left(\gO{\tk{i}, j}, \tk{i}, \gM{\tk{i}} | \tk{i} \leq t,\, j \in \{1 \leq l \leq \gls{dX} | \gM{\tk{i}, l} = 1  \} \right),
\end{equation*} 
such that $\hX{}_t = \E_{\glsP{OmFFP}\times\glsP{tOmFFP}}[\gX{t} | \At{t}]$ is the conditional expectation of $\gX{}$ given the noisy observations. We define $\tildeOle{t}$ in the same way as $\tildeXle{t}$ and note that similarly as before there exist measurable functions $\Fj{j}$ such that $\hX{}_{t,j} = \Fj{j}(t, \tau(t), \tildeOle{\tau(t)})$.
We need the following slight modification of Assumption~\ref{assumption:F}.
\begin{assumption}\label{ass:noisy obs setting}
We assume that Assumption~\ref{assumption:F} \cref{ass_main_1,ass_main_2,ass_main_3,ass_main_5} hold and additionally that:
\begin{enumerate}[label=(\roman*)]
\setcounter{enumi}{3}
\item $\Fj{j}$ are  continuous and differentiable in their first coordinate $t$ such that their partial derivatives with respect to $t$, denoted by $f_j$, are again continuous and there exists a $B >0$ and $p \in \N$ such that for every $t \in [0,\gls{T}]$ the functions $f_j, \Fj{j}$ are polynomially bounded in $\gls{Xs}$, i.e., 
$$|\Fj{j}(\tau(t), \tau(t), \tildeOle{\tau(t)})| +  | f_j(t, \tau(t), \tildeOle{\tau(t)})  | \leq B (\glst{Xs} +1)^p + B \sum_{i=0}^{\gn{}} |\epsilon_i|. $$ \label{ass_noisy_4}
\setcounter{enumi}{5}
\item $\gn{}$ is square-integrable, i.e.,\ $\E_{\glsP{tOmFFP}}[|\gn{}|^{2}] < \infty$. \label{ass_noisy_7}
\item The i.i.d.\ random noise variables $\epsilon_k$ are independent of $\gX{}, \gn{}, \gM{}, (\tk{i})_{1 \leq i \leq \gls{K}}$, are centered and square-integrable, i.e.,\ $\E_{\glsP{tOmFFP}}[\epsilon_k] = 0$ and $\E_{\glsP{tOmFFP}}[|\epsilon_k|^{2}] < \infty$.  \label{ass_noisy_6}
\end{enumerate}
\end{assumption}
\begin{rem}
    The relaxations on the assumption of observing $\gX{}_0$ completely discussed in \citet[Remark 2.2]{krach2022optimal} can equivalently be applied in this setting here.
\end{rem}
In this setting, the PD-NJ-ODE uses the noisy observations $\gO{\tk{i}}$ and $ \tildeOle{\tau(t)}$ as inputs instead of $\gX{}_{\tk{i}}$ and $\tildeXle{\tau(t)}$. Moreover, we define the new \emph{noise-adapted} objective function as described before as 
\begin{align}
\Psi: \, &\bD{} \to \R,\, Z \mapsto \Psi(Z) := \E_{\glsP{OmFFP}\times\glsP{tOmFFP}}\left[ \frac{1}{\gn{}} \sum_{i=1}^{\gn{}}    \left\lvert \gM{}_i \odot (\gO{\tk{i}} - Z_{\tk{i}-} ) \right\rvert_2^2 \right], \label{equ:Psi noisy obs} \\
\Phi : \, &\Theta \to \R, \, \theta \mapsto \Phi(\theta) := \Psi(\gY{}^{\theta}(\gX{})), \label{equ:Phi noisy obs}
\end{align}
and its Monte Carlo approximation $\hPhi{N}$ accordingly.

\subsection{Convergence Theorem with Noisy Observations}\label{sec:Convergence Theorem with Noisy Observations}
In the setting defined in Section~\ref{sec:Setting with Noisy Observations}, Theorem~\ref{thm:1} holds equivalently as before. 
\begin{theorem}\label{thm:noisy obs}
    If Assumption~\ref{ass:noisy obs setting} is satisfied and using the definitions of Section~\ref{sec:Setting with Noisy Observations}, the claims of \Cref{thm:1} hold equivalently, upon replacing the original loss functions and their Monte Carlo approximations by their noise-adapted versions. In particular, we obtain convergence  of our estimator~$\gY{}^{\thetamMin{m,N_m}}$ to the true conditional expectation $\hX{}$ in $\dk{k}$.
\end{theorem}
To prove this, we first need to adjust the orthogonal projection result \citep[Lemma 4.2]{krach2022optimal} for this setting.
At this point we want to highlight the relevance of this result as the backbone of the proof of \Cref{thm:noisy obs}, which is first used to show that $\hX{}$ is a minimizer of $\Psi$, then to show that it is unique and finally to bound the distance between $\hX{}$ and $\gY{}^{\thetamMin{m}}$ in $\dk{k}$ through the difference of their loss function values.
\begin{lem}\label{lem:L2 identity noisy obs setting}
For any $\bA{}$-adapted process $Z$ it holds that
\begin{multline*}
\E_{\glsP{OmFFP} \times\glsP{tOmFFP}}\left[\tfrac{1}{\gn{}} \sum_{i=1}^{\gn{}} \left\lvert \gM{\tk{i}} \odot ( \gO{\tk{i}} - Z_{\tk{i}-} ) \right\rvert_2^2\right] \\
	= \E_{\glsP{OmFFP} \times\glsP{tOmFFP}}\left[ \tfrac{1}{\gn{}}\sum_{i=1}^{\gn{}} \left\lvert \gM{\tk{i}} \odot ( \gO{\tk{i}} - \hX{}_{\tk{i}-} ) \right\rvert_2^2\right] + \E_{\glsP{OmFFP} \times\glsP{tOmFFP}}\left[\tfrac{1}{\gn{}}\sum_{i=1}^{\gn{}} \left\lvert \gM{\tk{i}} \odot (  \hX{}_{\tk{i}-} - Z_{\tk{i}-}) \right\rvert_2^2\right] .
\end{multline*}
\end{lem}

\begin{proof}
    First note that by Assumption~\ref{ass:noisy obs setting} point \ref{ass_main_3} we have that $\gX{}_{\tk{i}} = \gX{}_{\tk{i}-}$ almost surely and when defining $\gO{\tk{i} -} := \gX{}_{\tk{i} -} + \epsilon_i$ we therefore also have that $\gO{\tk{i} } = \gO{\tk{i} -}$ almost surely.
Similarly as in  \citet[Lemma 4.2]{krach2022optimal}, we can derive for $\hO{\tk{i}-} := \E_{\glsP{OmFFP} \times\glsP{tOmFFP}}[\gO{\tk{i} -} \, | \, \At{\tk{i}-}]$ that
\begin{multline*}
\E_{\glsP{OmFFP} \times\glsP{tOmFFP}}\left[\tfrac{1}{\gn{}}\sum_{i=1}^{\gn{}} \left\lvert \gM{\tk{i}} \odot ( \gO{\tk{i}-} - Z_{\tk{i}-} ) \right\rvert_2^2\right] \\
	 = \E_{\glsP{OmFFP} \times\glsP{tOmFFP}}\left[\tfrac{1}{\gn{}}\sum_{i=1}^{\gn{}} \left\lvert \gM{\tk{i}} \odot ( \gO{\tk{i}-} - \hO{\tk{i}-} ) \right\rvert_2^2\right] + \E_{\glsP{OmFFP} \times\glsP{tOmFFP}}\left[\tfrac{1}{\gn{}}\sum_{i=1}^{\gn{}} \left\lvert \gM{\tk{i}} \odot (  \hO{\tk{i}-} - Z_{\tk{i}-} ) \right\rvert_2^2\right].
\end{multline*}
To conclude the proof, it is enough to note that 
\begin{equation}\label{equ:O hat equal Xhat}
    \hO{\tk{i}-} = \hX{}_{\tk{i}-} + \E[\epsilon_i | \At{\tk{i}-}] = \hX{}_{\tk{i}-} + \E[\epsilon_i] = \hX{}_{\tk{i}-},
\end{equation}
using that $\epsilon_i$ has expectation $0$ and is independent of $\At{\tk{i}-}$. 
\end{proof}

In the following we sketch the proof of Theorem~\ref{thm:noisy obs}, by only outlining those parts of it that need to be changed in comparison with the original proof of \Cref{thm:1} in \citet{krach2022optimal}. 
The main differences are that the loss function needs to be adjusted whenever used, and when showing integrability, we additionally have to account for the noise terms $\epsilon_k$. %
A full proof is given in Appendix~\ref{sec:Combining the Two Extensions With Full Proof}.

\begin{proof}[Sketch of Proof of Theorem~\ref{thm:noisy obs}]
    First, it follows directly from Lemma~\ref{lem:L2 identity noisy obs setting} that $\Psi(\hX{}) = \min_{Z \in \bD{}}\Psi(Z)$, i.e., that $\hX{}$ is a minimizer of the redefined objective function $\Psi$.
    Secondly, again by Lemma~\ref{lem:L2 identity noisy obs setting}, we have for any process $Z \in \bD{}$ that 
    \begin{equation*}
        \Psi(Z) = \Psi(\hX{})  + \E_{\glsP{OmFFP} \times\glsP{tOmFFP}}\left[\tfrac{1}{\gn{}}\sum_{i=1}^{\gn{}} \left\lvert \gM{\tk{i}} \odot (  \hX{}_{\tk{i}-} - Z_{\tk{i}-}) \right\rvert_2^2\right] ,
    \end{equation*} 
    hence $\Psi(Z) > \Psi(\hX{})$ follows as before if $Z$ is not indistinguishable from $\hX{}$, meaning that $\hX{}$ is the unique minimizer of $\Psi$.

    Under Assumption~\ref{ass:noisy obs setting} the approximation of the functions $f_j, \Fj{j}$ by bounded output feedforward neural networks works similarly as before, with the slight adjustment that their differences are now upper bounded by 
    $$U := 3 B\left((\glsT{Xs} +1)^p + \sum_{i=0}^{\gn{}} |\epsilon_i| \right).$$ 
    Defining 
    \begin{equation*}
        c_m := c \, \varepsilon (\gls{T}+1) \gls{dX} +  c (\gls{T}+1) \gls{dX} U \left( \1_{\{ \glsT{Xs} \geq 1/\varepsilon \}} + \1_{\{ \gn{} \geq  1/\varepsilon \}} + \1_{\{ \delta \leq \varepsilon \}} \right)
    \end{equation*}
    it follows that there exists $\theta_m^*\in\Thetam{m}$ such that $\left\lvert \gY{}_t^{\theta_m^*} - \hX{}_t  \right\rvert_2 \leq c_m$ for all $t \in [0,\gls{T}]$.
    Convergence of $\Phi(\theta_m^\star)$ to $\Psi(\hX{})$ then follows similarly as before, when noting that by Assumption~\ref{ass:noisy obs setting}
    \begin{equation}\label{equ:integraility in noisy obs setting}
        \E \left[ (\glst{Xs} +1)^{2p} + \left(\sum_{i=0}^{\gn{}} |\epsilon_i|\right)^2 \right] \leq \E \left[ (\glst{Xs} +1)^{2p} \right] +  \E[\gn{2}] \E \left[|\epsilon_0|^2 \right] < \infty,
    \end{equation}
    using Cauchy--Schwarz and that the $\epsilon_i$ are i.i.d.\ and independent of $\gn{}$ for the first step and the integrability of $\gls{Xs}$, $\epsilon_0$ and $\gn{2}$ for the upper bound.
    Moreover, the convergence of $\dk{k}(\hX{}, \gY{}^{\theta_m^\star}) \to 0$ follows as before.
    
    Finally, the remaining claims of the theorem (including the Monte Carlo convergence) also hold similarly as before upon replacing $\gX{}_{\tk{i}}$ by $\gO{\tk{i}}$ and noting that the integrability of $\sup_{\theta} h(\theta, \xi_j)$ follows from \eqref{equ:integraility in noisy obs setting}.
\end{proof}

\subsection{More General Noise Structure \& Conditional Moments}\label{sec:More General Noise Structure Conditional Moments}
Revisiting the proof in Section~\ref{sec:Convergence Theorem with Noisy Observations}, we see that the noise terms need neither be independent nor centered.
If we assume that the conditional bias of the noise,
\[\beta_i(\tildeOle{\tau(t)}) := \E[\epsilon_i | \At{\tk{i}-}],\]
is a \emph{known} function of the observations (using Doob-Dynkin Lemma \cite[Lemma~2]{taraldsen2018optimal} for its existence), then we can modify the objective function by subtracting it. This leads to
\begin{align}\label{equ:loss noisy obs generalised}
\Psi: \, &\bD{} \to \R,\, Z \mapsto \Psi(Z) := \E_{\glsP{OmFFP}\times\glsP{tOmFFP}}\left[ \frac{1}{\gn{}} \sum_{i=1}^{\gn{}}    \left\lvert \gM{}_i \odot \left(\left(\gO{\tk{i}} - \beta_i(\tildeOle{\tk{i-1}})\right) - Z_{\tk{i}-} \right) \right\rvert_2^2 \right].
\end{align}
Revisiting \eqref{equ:O hat equal Xhat}, which is the only part of the proof where we needed the noise terms to be centered, we see that
\begin{equation}\label{eq:lossNoiseKnownMean}
    \E\left[ \left(\gO{\tk{i}-} - \beta_i(\tildeOle{\tk{i-1}})\right) | \At{\tk{i}-} \right] = \hX{}_{\tk{i}-} + \E[\epsilon_i \mid \At{\tk{i}-}] - \beta_i = \hX{}_{\tk{i}-}.
\end{equation}
This implies that the statement of Lemma~\ref{lem:L2 identity noisy obs setting} holds equivalently under the reduced assumption of a known conditional bias function, when using the adjusted loss \eqref{equ:loss noisy obs generalised}.
Additionally assuming that
$\E\left[ \left(\sum_{i=0}^{\gn{}} |\epsilon_i| \right)^2 \right] < \infty$, the following result follows as before.
\begin{cor}\label{cor:extended noisy obs thm}
    In the setting described in this sub-section (i.e., arbitrary known mean of the noise and no independence assumption on the noise), which is a generalisation of the setting in Section~\ref{sec:Setting with Noisy Observations}, Theorem~\ref{thm:1} holds equivalently as before when using the objective function~\eqref{equ:loss noisy obs generalised}.
\end{cor}
The following remark explains how Corollary~\ref{cor:extended noisy obs thm} can be used to predict conditional higher moments (instead of only the conditional expectation) under certain assumptions. 
\begin{rem}
    This result makes it possible to compute the conditional moments of $\gX{}$ given the noisy observations, which doesn't work in the setting of Section~\ref{sec:Setting with Noisy Observations}.
    In particular, we consider observations $\gO{\tk{i}-} = \gX{}_{\tk{i}-} + \epsilon_i$, where we assume that
    \begin{itemize}
        \item $\epsilon_i$ is independent of $\At{\tk{i}-}$,
        \item $\epsilon_i$ is conditionally independent of $\gX{}_{\tk{i}-}$ given $\At{\tk{i}-}$
        \item and $\epsilon_i$ have known finite moments.
    \end{itemize} 
    Remark that Proposition~\ref{prop:conditional independence prop 6} implies that the first two assumptions are in particular satisfied if $\epsilon_i$ is independent of $\sigmab(\At{\tk{i}-}, \gX{}_{\tk{i}-})$.
    The binomial theorem implies for any $q \in \N$
    \begin{equation*}
        \gO{\tk{i}-}^q = \gX{}_{\tk{i}-}^q + \sum_{j=1}^q \binom{q}{j} \gX{}_{\tk{i}-}^{q-j} \epsilon_i^j.\footnote{Note that $q$, $j$ and $q-j$ denote exponents here rather than superscripts.}
    \end{equation*}
    We interpret the entire sum as the observation noise and accordingly define the conditional bias of the observation noise of the $q$-th moment as
    \begin{equation*}
        \beta_i^q := \E\left[ \sum_{j=1}^q \binom{q}{j} \gX{}_{\tk{i}-}^{q-j} \epsilon_i^j \mid \At{\tk{i}-} \right] = \sum_{j=1}^q \binom{q}{j} \E[ \gX{}_{\tk{i}-}^{q-j} | \At{\tk{i}-} ] \E[\epsilon_i^j],
    \end{equation*}
    where we use the assumptions on $\epsilon_i$ together with Proposition~\ref{prop:conditional independence prop 5} for the second equality.
    
    Then an inductive argument shows that $\beta_i^q$ is a known function of the observations, using the assumption that the moments of $\epsilon_i$ are known.
    Indeed, to compute $\beta_i^q$ the conditional expectations of smaller moments  $\E[ \gX{}_{\tk{i}-}^{q-j} | \At{\tk{i}-} ]$ need to be computed, which can be done according to the induction hypothesis (note that the base case follows directly from Corollary~\ref{cor:extended noisy obs thm} and the assumptions on $\epsilon_i$). Therefore, Corollary~\ref{cor:extended noisy obs thm} implies that we can compute $\E[ \gX{}_{\tk{i}-}^{q} | \At{\tk{i}-} ]$ (assuming that we reach the limit where the PD-NJ-ODE output equals the conditional expectation). In case of an exponential moment assumption $\E[ \exp(\lambda |\gX{}_{\tk{i}-}|) ] < \infty$ for some $ \lambda > 0 $ we can therefore infer the conditional law of $\gX{}_{\tk{i}-}$.
\end{rem}

In the remainder of this section, we discuss the cases of linear and non-linear (noisy) observation models.
\begin{rem}[Linear Observation Model]\label{rem:Linear Observation Model}
    Let us consider a linear observation process 
    \begin{equation*}
        \gO{\tk{k}} := A \gX{\tk{k}} + \epsilon_k,
    \end{equation*}
    for some fixed matrix $A \in \R^{d \times \gls{dX}}$, for $d\in \N$, where we have observations of $\gO{}$, but are ultimately interested in predicting $\gX{}$ without having access to any observations of $\gX{}$. For this to be suitable for our framework, we need that Assumption~\ref{ass:noisy obs setting} is satisfied for $Z:= A \gX{}$.\footnote{We note that by linearity of the (conditional) expectation we have $\E[Z_t \, | \, \At{t}] = A \, \E[ \gX{t} \, | \, \At{t}]$, hence, $Z$ satisfies Assumptions~\ref{ass:noisy obs setting} \ref{ass_main_3} to \ref{ass_main_5}, if $\gX{}$ does.}
    Then we can apply our method to approximate $\hat{Z}_t = \E[Z_t \, | \, \At{t}]$, as discussed before. 
    To be able to infer predictions for $\gX{}$ from $\hat{Z}$, we need to further assume that $A$ defines an injective map, hence, a left inverse $A^{-1}$ with $A^{-1}A = \id$ exists.
    Under this assumption,  $\E[ \gX{t} \, | \, \At{t}] = A^{-1} \, \hat{Z}_t$, by linearity.
    It is important to note that in an incomplete observation setting, it would in general not be possible to apply our method to directly approximate $\gX{}$, since (noisy) observations of $\gX{}$ cannot be computed from (noisy) observations of $Z$, if they are incomplete (i.e., $A^{-1} \gO{}$ might not be well defined). Therefore, the detour via first computing $\hat{Z}$ is essential.
    Moreover, it should be apparent that in the case the map defined by $A$ is not injective (or at least injective on the support of $\gX{}$), information about $\gX{}$ will be lost when only observing $\gO{}$. Nevertheless, any process of the form $BZ$ for another matrix $B$ can be predicted similarly as described above when observing $\gO{}$.
\end{rem}

\begin{rem}[Non-linear Observation Model \& Beyond]
    Let us consider the observation process 
    \begin{equation*}
        \gO{\tk{k}} := \phi( \gX{\tk{k}} ) + \epsilon_k,
    \end{equation*}
    for some measurable function $\phi$, where we again have observations of $\gO{}$ (in particular we assume that $\phi(\gX{})$ satisfies the assumptions), but are ultimately interested in predicting $\gX{}$.
    Then, even if a left inverse $\phi^{-1}$ with $\phi^{-1} \circ \phi = \id$ exists,  the technique from \Cref{rem:Linear Observation Model} does not work, since $\E[ \gX{t} \, | \, \At{t}] \neq \phi^{-1}\left( \E[ \phi(\gX{})_t \, | \, \At{t}] \right)$ in general.
    However, under the stronger assumption that we have access to (noisy) samples of the joint process $\xi := (\gX{},\phi(\gX{}))$ for training and that this $\xi$ satisfies Assumption~\ref{ass:noisy obs setting}, we can compute $\hat{\xi} = (\hX{}, \widehat{\phi(\gX{})})$, which can then also be evaluated for samples where only (noisy) observations of the $\phi(\gX{})$-coordinates are available (cf.\ \citet[Corollary~6.2]{krach2022optimal}).
    Moreover, we note that the same is true, when replacing $\phi(\gX{})$ by a general process $\Zfilter$.
\end{rem}

\subsection{Examples of Processes Satisfying the Assumptions}\label{sec:examples noisy obs}
In principle, all the examples presented in \citet[Section 7]{krach2022optimal} are valid examples for this setting when adding some type of i.i.d.\ observation noise  satisfying our assumptions, as e.g. Gaussian or uniform noise.
However, it is important to note that the (true) conditional expectation is not the same, since we now condition on the noisy observations $\gO{\tk{i}}$ instead of the original observations $\gX{}_{\tk{i}}$.
Therefore, we give one explicit example where we compute the conditional expectation in the noisy observation setting.

    \subsubsection{Brownian Motion with Gaussian Observation Noise}
    \label{sec:Brownian Motion with Gaussian Observation Noise}
    Let $\gX{} := W$ be a standard Brownian motion and let $\epsilon_0=0$, $\epsilon_i \sim \normalDistribution(0, \sigma^2)$ for $i\geq 1$ be the i.i.d.\ noise terms for some $\sigma > 0$. Then $\gO{\tk{i}} = \gX{}_{\tk{i}}+ \epsilon_i$ are the observations. Clearly, all integrability assumptions are satisfied by $\gX{}$ and $\epsilon_i$ (cf. \citet[Section 7.5]{krach2022optimal}). 
    To compute the true conditional expectations we first note that the independent increments property of the Brownian motion imply for $\tk{k} \leq t < \tk{k+1}$
    \begin{equation*}
        \E[\gX{t} | \At{t}] = \E[W_t - W_{\tk{k}} | \At{\tk{k}} ] + \E[ W_{\tk{k}} | \At{\tk{k}} ] = \E[ W_{\tk{k}} | \At{\tk{k}} ] =  \E[ W_{\tk{k}} | \gO{\tk{1}}, \dotsc, \gO{\tk{k}}],
    \end{equation*}
    and therefore, $f(s, \tau(t), \tildeOle{\tau(t)} ) = 0$.
    Since $W$ is a Brownian motion and $\epsilon_i$ are independent i.i.d. Gaussian noise terms, we know that
    $$v := (\gO{}_{\tk{1}}, \dotsc, \gO{}_{\tk{k}}, W_{\tk{k}})^\top \sim \normalDistribution(0, \Sigma)$$
    where
    $$  \Sigma  = \begin{pmatrix}
    \Sigma_{11} &  \Sigma_{12} \\
    \Sigma_{21} &  \Sigma_{22}
    \end{pmatrix} \in \R^{(k+1) \times (k+1)},$$
    with $\Sigma_{11} \in \R^{k\times k}$ and  $(\Sigma_{11})_{i,j} = \min(\tk{i},t_j) + \sigma^2 \1_{\{i=j\}}$, $\Sigma_{12}^\top = \Sigma_{21} = (\tk{1}, \dotsc, \tk{k}) \in \R^{1\times k}$ and $\Sigma_{22} = \tk{k}$.
    Then the conditional distribution of $(W_{\tk{k}} \, | \, \gO{\tk{1}}, \dotsc, \gO{\tk{k}})$ is again normal with mean $\hat \mu :=  \Sigma_{21}  \Sigma_{11}^{-1} (\gO{\tk{1}}, \dotsc, \gO{\tk{k}} )^\top $ and variance $\hat \Sigma :=  \Sigma_{22} -   \Sigma_{21}  \Sigma_{11}^{-1}  \Sigma_{12}$ \citep[Proposition~3.13]{Eaton2007Multi}. 
    In particular we have 
     \begin{equation*}
        \E[\gX{t} | \At{t}] = \E[ W_{\tk{k}} | \gO{\tk{1}}, \dotsc, \gO{\tk{k}}] = \hat{\mu}.
    \end{equation*}

\subsection{A Practical Note on Using the Noise-Adapted Loss Function}\label{sec:A Practical Note on Using the Noisy Loss Function}
We have seen in Section~\ref{sec:Convergence Theorem with Noisy Observations} and~\ref{sec:More General Noise Structure Conditional Moments}, considering the limit where network size~$m$ and training samples~$N$ go to infinity, that there is no disadvantage of the noise-adapted loss function \eqref{equ:loss noisy obs generalised} compared to the original loss function \eqref{equ:Psi}. In particular, the noise-adapted training framework yields the optimal solution with noisy observations, but also if there is no observation noise. Indeed, setting $\epsilon_k=0$ for all $k$, we have $\gO{\tk{k}} = \gX{}_{\tk{k}}$ and therefore the original result also holds with the noise-adapted loss function.
Hence, the question arises, why one should use the original loss function \eqref{equ:Psi} at all anymore?

To answer this question, we first note that in practice we are not in the limit case, but 
we only have access to a finite training set (i.e., we only observe a finite amount of observations $\gn{}^{(j)}$ from finite number of paths $N$). 
Hence, the inductive bias when training the model becomes more important. See \Cref{sec:inductive bias} for more details on the inductive bias. In particular, using the original loss function \eqref{equ:Psi} in a setting without observation noise is preferable, since it directly penalizes the model for not jumping to observations and therefore making it easier for the model to learn this behaviour. Indeed, in the noise-adapted loss function, the penalization for not jumping to an observation is more indirect, since it will only be punished at the following observation time until which the output of the model further evolves through the neural ODE. Therefore, the feedback signal is weaker, i.e., the respective gradients are smaller. 

In a setting with observation noise, the practical relevance of the noise-adapted loss function depends on the size of the variance of the noise, $\operatorname{Var}_{noise}$, compared to the size of the variance of the data, $\operatorname{Var}_{data}$. 
Clearly, if the variance ratio $\frac{\operatorname{Var}_{noise}}{\operatorname{Var}_{data}}$ is small, the noise does not have a large impact on the observations and therefore neither on the training of the model when using the original loss function \eqref{equ:Psi}. Due to the better inductive bias during training, it might therefore be beneficial to use \eqref{equ:Psi} instead of the noise-adapted loss \eqref{equ:loss noisy obs generalised}.
On the other hand, if the variance ratio is large, the impact of the noise on the observations is substantial and therefore the noise-adapted training framework leads to better results. 
This behaviour is well visible in the experiment presented in Section~\ref{sec:Physionet with Observation Noise}.
The turning point, where training with the noise-adapted loss function becomes better, is problem specific (e.g., the number of training samples has a big influence) and, if in doubt, we suggest to train the PD-NJ-ODE model with both loss functions and compare their results.

\section{PD-NJ-ODE with Dependence between \texorpdfstring{$\gX{}$}{X} and the Observation Framework}\label{sec:PD-NJ-ODE with Dependence betweenXand Observation Framework}
Recall that the observation mask process is given by $\gM{}$ and the underlying process by $\gX{}$. In this section, we remove the assumptions that the observation times are independent of $\gX{}$, and that $\gM{}$ is independent of the observation times and of $\gX{}$. In essence, the model is now defined on only one probability space $\glsP{OmFFP}$ and no independence assumptions between the random variables are made. Instead, we need some weaker conditional independence assumptions to recover the results of Theorem~\ref{thm:1}.

\subsection{Intuition on independence assumptions}\label{sec:Intuition on independence assumptions}
In many real-world applications, the independence of the process $\gX{}$ and the observation framework (i.e., $(\tk{i})_{i\in \{1,\dots,\gn{}\}}$ and $\gM{}$) is heavily violated. 
For example, consider the irregular measurements of a patient's health parameters. A nurse or doctor will only take (expensive)  measurements if information on the patient's state, $\gX{}$, hints that the measurement is necessary. 
In practice, different measurements are taken from different patients depending on observations of their state. This motivates the crucial importance of lifting the independence assumption for real-world applications.

However, even in this paper, we cannot completely remove any independence assumption; we still need conditional independence of the process $\gX{}$ and the observation framework, given all past observations (as we make precise in Assumption~\ref{ass:dependence} \cref{ass_condindep,ass:cond ind X tk} and \Cref{prop:dropping independece assumption for n}).
This is a realistic assumption, which we show by continuing our hospital example. If we assume that \emph{every} piece of information the hospital gets from the patient is immediately logged as an observation (perhaps noisy or incomplete) of $\gX{}$, then all information about $\gX{}$ that the hospital has is contained entirely within these observations. Thus measurement decisions are conditionally independent of $\gX{}$, given the past observations (if there are no further hidden confounders).

On the other hand, it is easy to violate this assumption. Imagine that the patient tells the nurse and doctor she feels feverish and they subsequently take her temperature. If they log her temperature, but not her feeling, the assumption is violated, as the hospital has a piece of information that is not logged as an observation of $\gX{}$. In such cases where information is not logged as an observation, our model (and most other classical forecasting methods) would learn a biased forecast of the body temperature. In the extreme case that body temperature is only measured if patients feel very feverish, measurements will record a high temperature, which leads our model to always predict a high body temperature (i.e., the expected body temperature conditioned on feeling feverish and all other past observations) even if the patient does not feel feverish.

There are often ways to mitigate such issues. In our case, for example, by first logging the feverish feeling and then the actual measurement at a later time stamp. Moreover, a non-feverish feeling has to be logged whenever no feverish feeling is reported during a regular nurse visit.
Nevertheless, this discussion shows that it is important to be aware that even our weakened, more realistic assumption of conditional independence is often not fully satisfied in practical situations. Hence, one must be careful when verifying the assumptions and potentially adapt the experimental setting such that they are satisfied.

\subsection{Setting with Dependence}\label{sec:Setting with Dependence}
To allow for dependence, we only consider the probability measure $\glsP{OmFFP}$ and define $\gX{}, \gn{}, \gls{K}, \tk{i}, \tau, \gM{}, \At{t}$ similar as before, but all on the same (filtered) probability space~\gls{OmFFP} associated with $\glsP{OmFFP}$.
For a random variable $Z$ and a family of sets $\mathcal{B}$ we use the natural notation for their smallest jointly generated sigma algebra $\boldsymbol{\sigma}(Z, \mathcal{B}) := \boldsymbol{\sigma}(Z) \lor \boldsymbol{\sigma}(\mathcal{B})$.

Then we need the following assumptions, most of which are borrowed from Assumption~\ref{assumption:F}. Only~\ref{ass_condindep} and~\ref{ass_newass} are new, while \ref{ass_dependence_1} is a strict generalisation of Assumption~\ref{assumption:F}~\ref{ass_main_1}.
\begin{assumption} \label{ass:dependence}
We assume that \Cref{assumption:F} \cref{ass_main_2,ass_main_3,ass_main_4,ass_main_5,ass_main_6} hold, with all instances of $\glsP{tOmFFP}$ and $\glsP{OmFFP} \times \glsP{tOmFFP}$ replaced by $\glsP{OmFFP}$. Additionally, we assume that:
\begin{enumerate}[label=(\roman*)] 
    \item $\gM{0, j} = 1$ for all $1 \le j \le \gls{dX}$ ($\gX{}$ is completely observed at $0$) and $\abs{\gM{}_k}_1 > 0$ for every $1 \le k \le \gls{K}$ $\glsP{OmFFP}$-almost surely (at every observation time at least one coordinate is observed). \label{ass_dependence_1}
    \setcounter{enumi}{6}
    \item For every  $1 \le k \le \gn{}$, $\gX{}_{\tk{k}-}$ is conditionally independent of $\sigmab(\gn{}, \gM{\tk{k}})$ given $\At{\tk{k}-}$. \label{ass_condindep} 
    \item For all $1 \le k \le \gls{K}$, $1 \le j \le \gls{dX}$ there is some $\eta_{k,j} >0$ such that $\glsP{OmFFP} ( \gM{k, j} = 1 \mid \sigmab(\gn{}, \At{\tk{k}-} )) > \eta_{k,j}$ (i.e., given the currently known information and $\gn{}$, for each coordinate the probability of observing it at the next observation time is positive). \label{ass_newass} 
    \item For every $1 \leq k \leq \gls{K}$ the process $\gX{}$ is conditionally independent of $\tk{k}$ given $\At{\tk{k-1}}$. \label{ass:cond ind X tk}
\end{enumerate}
\end{assumption}

\begin{rem}
    The relaxations on the assumption of observing $\gX{}_0$ completely discussed in \citet[Remark 2.2]{krach2022optimal} can equivalently be applied here.
\end{rem}

We can use the original objective function \eqref{equ:Phi} and its Monte Carlo approximation \eqref{equ:appr loss function}.

\subsection{Convergence Theorem with Dependence}
In the setting defined in Section~\ref{sec:Setting with Dependence}, Theorem~\ref{thm:1} holds equivalently. 
\begin{theorem}\label{thm:depndence}
    If Assumption~\ref{ass:dependence} is satisfied and using the definitions of Section~\ref{sec:Setting with Dependence}, the claims of \Cref{thm:1} hold equivalently. In particular, we obtain convergence  of our estimator~$\gY{}^{\thetamMin{m,N_m}}$ to the true conditional expectation $\hX{}$ in $\dk{k}$.
\end{theorem}

The main technical challenge when generalising \Cref{thm:1} to \Cref{ass:dependence} is to replace the arguments using independence and e.g.\ Fubini's theorem, by arguments using conditional independence. This difficulty is  well visible when comparing the proof of the orthogonal projection \Cref{lem:L2 identity dependence} below, with the original proof of \citet[Lemma~4.2]{krach2022optimal}. Before, the claim followed relatively immediately from Fubini's theorem and the standard orthogonal projection result for conditional expectations of random variables. In contrast to this, now we have to condition first on the known information and the additional random variables of the observation framework, then we have to argue by conditional independence that this is the same as conditioning only on the known information and only after that we can apply the conditional version \Cref{prop:newb2} of the orthogonal projection result.
The relevance of \Cref{lem:L2 identity dependence} as the backbone of the proof of \Cref{thm:depndence} was already discussed in Section~\ref{sec:Convergence Theorem with Noisy Observations}.
Finally, in \Cref{lem:ck} we show how Assumption~\ref{ass:dependence}\ref{ass_newass} can replace independence between $\gM{}$ and $\gX{}$, when deriving that the distance between $\hX{\tk{k}-}$ and $\gY{\tk{k}-}$ can be bounded through their distance when multiplied with the observation mask.

First, we prove the extension of the standard $L^2$-orthogonality result that was stated in \citet[Proposition B.2]{herrera2021neural}.

\begin{prop}
\label{prop:newb2}
Let $(\Omblack{}, \Fblack{}, \PBlack)$ be a fixed probability space, and $\mathcal A, \mathcal B$ be sub-$\sigma$-algebras such that $\mathcal B \subseteq \mathcal A \subseteq \Fblack{}$. For some random variable $\Xblack{} \in L^2(\Omblack{}, \Fblack{}, \PBlack)$ we define $\hXblack{} := \cexpect{\Xblack{}}{\mathcal A}$. Then for every random variable $Z \in L^2(\Omblack{}, \mathcal A, \PBlack{})$  with $\PBlack{}(Z \neq \hXblack{}) > 0$ we have
\begin{equation*}
    \cexpect{\abs{\Xblack{} - Z}^2_2}{\mathcal{B}} = \cexpect{\abs{\Xblack{} - \hXblack{}}^2_2}{\mathcal{B}} + \cexpect{\abs{Z - \hXblack{}}^2_2}{\mathcal{B}} 
    \ge \cexpect{\abs{\Xblack{} - \hXblack{}}^2_2}{\mathcal B},
\end{equation*}
with strict inequality with positive probability. 
\end{prop}

The proof is based on \citet[Theorem 5.1.8]{Durrett:2010:PTE:1869916}. We focus on the one-dimensional case, though this can easily be generalised to multiple dimensions via the 2-norm, as in \cite{herrera2021neural}.
\begin{proof}
We begin by expanding the left hand side.
\begin{multline*}
    \cexpect{\abs{\Xblack{} - Z }^2}{\mathcal B} = \cexpect{\abs{(\Xblack{} - \hXblack{}) - (Z - \hXblack{})}^2}{\mathcal B} \\
    = \cexpect{\abs{\Xblack{} - \hXblack{}}^2}{\mathcal{B}} + \cexpect{\abs{Z - \hXblack{}}^2}{\mathcal B} - 2\cexpect{(\Xblack{} - \hXblack{})(Z - \hXblack{})}{\mathcal B}
\end{multline*}
We now analyse the cross term, which expands to $\cexpect{Z(\Xblack{}-\hXblack{})}{\mathcal B} - \cexpect{\hXblack{} (\Xblack{} - \hXblack{})}{\mathcal B}$. Focusing on the first term, we note that since $Z \in  L^2(\Omblack{}, \mathcal A, \PBlack{})$, it holds that $Z \cexpect{\Xblack{}}{\mathcal A} = \cexpect{Z \Xblack{}}{\mathcal A}$. By taking expectation (conditioned on $\mathcal B$) of both sides, we get
\begin{align*}
    \cexpect{Z \hXblack{}}{\mathcal B} = \cexpect{Z \cexpect{\Xblack{}}{\mathcal A}}{\mathcal B} = \cexpect{\cexpect{Z \Xblack{}}{\mathcal A}}{\mathcal B} = \cexpect{Z \Xblack{}}{\mathcal B}
\end{align*}
via the tower property, as $\mathcal B \subseteq \mathcal A$. Hence, $\cexpect{Z(\Xblack{} - \hXblack{})}{\mathcal B} = 0$.
Note that showing this only required that $Z$ is $\mathcal A$-measurable. Since this is also satisfied by $\hXblack{}$ we directly have $\cexpect{\hXblack{}(\Xblack{} - \hXblack{})}{\mathcal B} = 0$ and therefore the cross term vanishes and the equality follows.
The inequality holds since $\cexpect{\abs{Z - \hXblack{}}^2_2}{\mathcal{B}}$ is non-negative. We therefore just need to show that the inequality is strict with positive probability. To this end, assume for the sake of contradiction that $\cexpect{\abs{Z - \hXblack{}}^2_2}{\mathcal{B}} = 0$ $\PBlack{}$-a.s, which implies (by the tower property) that $\expect{\abs{Z - \hXblack{}}^2_2} = 0$. This is only possible if $Z = \hXblack{}$ $\PBlack{}$-a.s., which contradicts the assumption that $\PBlack{}(Z \neq \hXblack{}) > 0$.
\end{proof}

As a next step we show that under \Cref{ass:dependence}\ref{ass:cond ind X tk} we can recover that $\hat{X}$ is the optimal predictor.
\begin{prop}\label{prop:rewriting of hat X}
    Under \Cref{ass:dependence}\ref{ass:cond ind X tk} we have 
    \begin{equation*}
    \hX{}_{\tk{k}-} = \cexpect{\gX{}_{t}}{\At{\tk{k-1}}} \Big\vert_{t=\tk{k}-} = \cexpect{\gX{}_{\tk{k}-}}{\At{\tk{k}-}} .
\end{equation*}
\end{prop}
\begin{proof}
    The result is a direct consequence of Proposition~\ref{prop:conditional independence prop 7}.
\end{proof}

Next we re-prove \citet[Lemma~4.2]{krach2022optimal} under the relaxed assumptions with the help of \Cref{prop:rewriting of hat X}.

\begin{lemma}
\label{lem:L2 identity dependence}
For any $\bA{}$-adapted process $Z \in L^2(\Om{}, \bA{}, \glsP{OmFFP})$ it holds that 
\begin{multline*}
    \expect{\frac{1}{\gn{}} \sum_{i=1}^{\gn{}} \abs{ \gM{\tk{i}} \odot (\gX{}_{\tk{i}} - Z_{\tk{i}-})}^2_2}
    = \ \expect{\frac{1}{\gn{}} \sum_{i=1}^{\gn{}} \abs{ \gM{\tk{i}} \odot (\gX{}_{\tk{i}} - \hX{}_{\tk{i}-}) }^2_2} + \expect{\frac{1}{\gn{}} \sum_{i=1}^{\gn{}} \abs{ \gM{\tk{i}} \odot (\hX{}_{\tk{i}-} - Z_{\tk{i}-} )}^2_2}.
\end{multline*}
\end{lemma}
\begin{proof}
\begin{align*}
    \E & \left[ \frac{1}{\gn{}} \sum_{i=1}^{\gn{}} \abs{\gM{\tk{i}} \odot (\gX{}_{\tk{i}-} - Z_{\tk{i}-})}^2_2  \right]
    = \ \sum_{i = 1}^{\gls{K}} \expect{\frac{1}{\gn{}} \ind{\{i \le \gn{}\}} \abs{\gM{\tk{i}} \odot (\gX{}_{\tk{i}-} - Z_{\tk{i}-})}^2_2} \\
    &= \ \sum_{i = 1}^{\gls{K}} \expect{\cexpect{\frac{1}{\gn{}} \ind{\{i \le \gn{}\}} \abs{\gM{\tk{i}} \odot (\gX{}_{\tk{i}-} - Z_{\tk{i}-})}^2_2}{\sigmab(\gn{}, \gM{\tk{i}}, \At{\tk{i}-})}} \\
    &= \ \sum_{i = 1}^{\gls{K}} \expect{\frac{1}{\gn{}} \ind{\{i \le \gn{}\}} \sum_{j=1}^{\gls{dX}} \gM{\tk{i}, j} \cexpect{ \abs{\gX{}_{\tk{i}-,j} - Z_{\tk{i}-,j}}^2 }{\sigmab(\gn{}, \gM{\tk{i}}, \At{\tk{i}-})}} \\
    &= \ \sum_{i = 1}^{\gls{K}} \expect{\frac{1}{\gn{}} \ind{\{i \le \gn{}\}} \sum_{j=1}^{\gls{dX}} \gM{\tk{i}, j} \cexpect{ \abs{\gX{}_{\tk{i}-,j} - Z_{\tk{i}-,j}}^2 }{\At{\tk{i}-}}} .
\end{align*}
The first step follows by monotone convergence, the last by Lemma~\ref{claim:condindep4.2} below. Now we can conclude by first applying the equality from Proposition~\ref{prop:newb2} with $\mathcal A, \mathcal B = \At{\tk{i}-}$, which yields that $ \cexpect{\gX{}_{\tk{i}-}}{\At{\tk{i}-}}$ minimises the expression, and then using \Cref{prop:rewriting of hat X} to conclude that $\hX{\tk{i}-}$ equals this minimizer.
Finally, we can reverse the above steps to arrive at the desired form. 
\end{proof}

\begin{lemma}\label{claim:condindep4.2}
Assume the context of Lemma~\ref{lem:L2 identity dependence}. Then for all $i, j$ it holds that
\begin{align*}
    \cexpect{ \abs{ \gX{}_{\tk{i}-, j} - Z_{\tk{i}-, j}}^2}{\sigmab(\gn{}, \gM{\tk{i}}, \At{\tk{i}-})} = \cexpect{ \abs{ \gX{}_{\tk{i}-, j} - Z_{\tk{i}-, j}}^2}{\At{\tk{i}-}}.
\end{align*}
\end{lemma}
\begin{proof}
We prove this by showing that if we expand the $\abs{\gX{}_{\tk{i}-, j} - Z_{\tk{i}-, j}}^2$ term, all three resulting terms can just be conditioned on $\At{\tk{i}-}$. This is a valid argument as $\gX{}$ and $Z$ are both assumed to be square-integrable. Note that squaring a random variable plays no role in the information given by the $\sigma$-algebra it is being conditioned on, and so we only need to analyse the terms $Z_{\tk{i}-, j}$, $\gX{}_{\tk{i}-, j}$, and $\gX{}_{\tk{i}-, j} Z_{\tk{i}-, j}$. See Appendix~\ref{app:condindep} for an overview of conditional independence and how it's used here. \\
\\
CASE $Z_{\tk{i}-, j}$:
$Z$ is $\bA{}$-adapted, and so $Z_{\tk{i}-, j}$ is $\At{\tk{i}-}$-measurable. Thus we have 
\begin{align*}
    \cexpect{Z_{\tk{i}-, j}}{\sigmab(\gn{}, \gM{\tk{i}}, \At{\tk{i}-})} = Z_{\tk{i}-, j} = \cexpect{Z_{\tk{i}-, j}}{\At{\tk{i}-}}
\end{align*} as desired. \\
\\
CASE $\gX{}_{\tk{i}-, j}$:
Assumption~\ref{ass:dependence} point \ref{ass_condindep} implies that $\gX{}_{\tk{i}-, j}$ is conditionally independent of $\sigmab(\gn{}, \gM{\tk{i}})$ given $\At{\tk{i}-}$. We therefore have by Proposition~\ref{prop:conditional independence prop 2}
\begin{align*}
    \cexpect{\gX{}_{\tk{i}-, j}}{\sigmab(\gn{}, \gM{\tk{i}}, \At{\tk{i}-})} = \cexpect{\gX{}_{\tk{i}-, j}}{\At{\tk{i}-}}.
\end{align*}
CASE $\gX{}_{\tk{i}-, j} Z_{\tk{i}-, j}$:
We combine the previous two ideas, namely that $Z_{\tk{i}-, j}$ is $\At{\tk{i}-}$-measurable and that $\gX{}_{\tk{i}-, j}$ is conditionally independent of $\sigmab(\gn{}, \gM{\tk{i}})$ given $\At{\tk{i}-}$. Thus 
\begin{multline*}
    \cexpect{\gX{}_{\tk{i}-, j} Z_{\tk{i}-, j}}{\sigmab(\gn{}, \gM{\tk{i}}, \At{\tk{i}-})}
    = \  Z_{\tk{i}-, j} \cexpect{\gX{}_{\tk{i}-, j} }{\sigmab(\gn{}, \gM{\tk{i}}, \At{\tk{i}-})} \\
    = \  Z_{\tk{i}-, j} \cexpect{\gX{}_{\tk{i}-, j} }{\At{\tk{i}-}}
    = \ \cexpect{Z_{\tk{i}-, j} \gX{}_{\tk{i}-, j} }{\At{\tk{i}-}}.
\end{multline*}
Combining these 3 cases proves the claim.
\end{proof}

The following lemma shows how Assumption~\ref{ass:dependence}\ref{ass_newass} can replace the independence between $\gM{}$ and $\gX{}$. 
\begin{lemma}\label{lem:ck}
There exists some $c_2(k) > 0$ such that for any $\bA{}$-adapted process $Z \in L^2(\Om{}, \bA{}, \glsP{OmFFP})$ we have
\begin{align*}
    \expect{\ind{\{\gn{} \ge k\}} \abs{\hX{}_{\tk{k}-} - Z_{\tk{k}-}}_1 } \le \frac{1}{c_2(k)} \expect{\ind{\{\gn{} \ge k\}} \abs{\gM{\tk{k}} \odot (\hX{}_{\tk{k}-} - Z_{\tk{k}-})}_1 }.
\end{align*}
\end{lemma}
\begin{proof}
Assumption~\ref{ass:dependence} point \ref{ass_newass} states that $0 < \eta_{k,j} < \glsP{OmFFP} ( \gM{k, j} = 1 \mid \sigma(\gn{}, A_{\tk{k}-} )) = \cexpect{\gM{\tk{k}, j}}{\sigmab(\gn{}, \At{\tk{k}-})} $ for all $k, j$.
Let $c_2 := c_2(k) := \min_{1 \le j \le \gls{dX}} \eta_{k,j}$, then $c_2 > 0$ and
\begin{multline*}
    \expect{ \ind{\{\gn{} \ge k\}} \abs{\gM{\tk{k}} \odot ( \hX{}_{\tk{k}-} - Z_{\tk{k}-}) }_1 } 
    = \sum_{j=1}^{\gls{dX}} \expect{\ind{\{\gn{} \ge k\}} \gM{\tk{k}, j} \abs{ \hX{}_{\tk{k}-, j} - Z_{\tk{k}-, j} } } \\
    = \sum_{j=1}^{\gls{dX}} \expect{\ind{\{\gn{} \ge k\}} \abs{ \hX{}_{\tk{k}-, j} - Z_{\tk{k}-, j}} \cexpect{\gM{\tk{k}, j}}{\sigmab(\gn{}, \At{\tk{k}-})} } \\
    \ge c_2 \ \expect{\ind{\{\gn{} \ge k\}} \abs{ \hX{}_{\tk{k}-} - Z_{\tk{k}-}}_1 },
\end{multline*}
where we used that $\ind{\{\gn{} \ge k\}} \abs{ \hX{}_{\tk{k}-, j} - Z_{\tk{k}-, j}}$ is $\sigmab(\gn{}, \At{\tk{k}-})$-measurable in the second line and the definition of $c_2$ in the last line.
\end{proof}

With the help of these lemmas we are now ready to prove Theorem~\ref{thm:depndence}.
In the following we again give  a sketch of the proof, by only outlining those parts of it that need to be changed in comparison with the original proof of \Cref{thm:1} in \citet{krach2022optimal}. We refer the interested reader to the full proof in Appendix~\ref{sec:Combining the Two Extensions With Full Proof} for all details.

\begin{proof}[Sketch of Proof of Theorem~\ref{thm:depndence}]
As before it follows that $\hX{}$ is a minimizer of $\Psi$. To show its uniqueness, we first note that 
\begin{equation}
 \E\left[ \1_{\{\gn{} \geq k\}} \left\lvert \hX{}_{\tk{k}-} - Z_{\tk{k}-}  \right\rvert_2 \right] 
 \leq \frac{c_3}{c_2(k)} \E\left[ \1_{\{\gn{} \geq k\}} \left\lvert \gM{\tk{k}} \odot ( \hX{}_{\tk{k}-} - Z_{\tk{k}-} ) \right\rvert_2 \right]
\end{equation}
is implied as before when using Lemma~\ref{lem:ck} instead of the independence.
Then 
\begin{equation*}
\begin{split}
 \E\left[\tfrac{1}{\gn{}}\sum_{i=1}^{\gn{}} \left\lvert \gM{\tk{i}} \odot (  \hX{}_{\tk{i}-} - Z_{\tk{i}-}) \right\rvert_2^2\right]
	\geq  \left( \frac{c_2(k)}{c_0 c_1 c_3} \right)^2 \dk{k}(\hX{}, Z)^2 > 0,
\end{split}
\end{equation*} 
follows as before, implying the uniqueness of $\hX{}$ as minimizer of $\Psi$.

The approximation of the functions $f_j$, $\Fj{j}$ also works as before. With this we have 
\begin{align*}
    \min_{Z \in \bD{}} \Psi(Z) &\le \Phi (\thetamMin{m}) \le \Phi(\theta^\ast_m) \\
    &= \ \expect{\frac{1}{\gn{}}\sum_{i = 1}^{\gn{}} \l(\abs{\gM{\tk{i}} \odot (\gX{}_{\tk{i}} - \gY{}^{\theta^\ast_m}_{\tk{i}} ) }_2 + \abs{\gM{\tk{i}} \odot (\gY{}^{\theta^\ast_m}_{\tk{i}} - \gY{}^{\theta^\ast_m}_{\tk{i}-} ) }_2 \r)^2 } \\
    &\le \ \E \l[ \frac{1}{\gn{}}\sum_{i = 1}^{\gn{}} \l(\abs{\gM{\tk{i}} \odot (\hX{}_{\tk{i}} - \gY{}^{\theta^\ast_m}_{\tk{i}} ) }_2 + \abs{\gM{\tk{i}} \odot (\gY{}^{\theta^\ast_m}_{\tk{i}} - \hX{}_{\tk{i}}) \r. }_2 \r. \\
    &\l. \l. \hspace{60 pt} + \abs{\gM{\tk{i}} \odot ( \hX{}_{\tk{i}} - \hX{}_{\tk{i}-} ) }_2 + \abs{\gM{\tk{i}} \odot ( \hX{}_{\tk{i}-} - \gY{}^{\theta^\ast_m}_{\tk{i}-} ) }_2 \r)^2 \r] \\
    &\le \ \E \l[ \frac{1}{\gn{}}\sum_{i = 1}^{\gn{}} \l(\abs{ \hX{}_{\tk{i}} - \gY{}^{\theta^\ast_m}_{\tk{i}} }_2 + \abs{ \gY{}^{\theta^\ast_m}_{\tk{i}} - \hX{}_{\tk{i}} \r. }_2 \r.  \\ 
    &\l. \l. \hspace{60 pt} + \abs{\gM{\tk{i}} \odot ( \hX{}_{\tk{i}} - \hX{}_{\tk{i}-} ) }_2 + \abs{ \hX{}_{\tk{i}-} - \gY{}^{\theta^\ast_m}_{\tk{i}-} }_2 \r)^2 \r] \\
    &\le \ \expect{\frac{1}{\gn{}}\sum_{i = 1}^{\gn{}} \l( \abs{\gM{\tk{i}} \odot ( \gX{}_{\tk{i}} - \hX{}_{\tk{i}-} ) }_2 + 3 c_m \r)^2 } \\
    &= \ \Psi(\hX{}) + \expect{\frac{1}{\gn{}}\sum_{i = 1}^{\gn{}} \l(6 c_m \abs{\gM{\tk{i}} \odot ( \gX{}_{\tk{i}} - \hX{}_{\tk{i}-} ) }_2 + 9 c^2_m \r) } \\
    &= \ \Psi(\hX{}) + \expect{\frac{1}{\gn{}}\sum_{i = 1}^{\gn{}} 6 c_m \abs{\gM{\tk{i}} \odot ( \gX{}_{\tk{i}} - \hX{}_{\tk{i}-} )}_2} + 9 \expect{c^2_m } \\
    &\le \ \Psi(\hX{}) + 6 \expect{\frac{1}{\gn{}}\sum_{i = 1}^{\gn{}} c_m \abs{ \gX{}_{\tk{i}} - \hX{}_{\tk{i}-} }_2} + 9 \expect{c^2_m } \\
    &\le \ \Psi(\hX{}) + 6 \expect{\frac{1}{\gn{}}\sum_{i = 1}^{\gn{}} c_m \cdot 2 \gls{T} B (\glsT{Xs} + 1)^p } + 9 \expect{c^2_m } \\
    &= \ \Psi(\hX{}) + 12 \gls{T} B \expect{ c_m (\glsT{Xs} + 1)^p } + 9 \expect{c^2_m },
\end{align*}
where the triangle inequality was used in the third line, and we use Assumption~\ref{ass:dependence} point~\ref{ass_main_4} (which implies that $|\hX{}_t| \le \gls{T}B(\gls{Xs} + 1)^p$ for all $t$) to construct a crude bound in the second to last line.
As in \cite{krach2022optimal}, dominated convergence can be used to show 
\begin{align*}
    \min_{\mathbb Z \in \bD{}} \Psi(Z) = \Psi(\hX{}) \le \Phi (\thetamMin{m}) \le \Phi(\theta^\ast_m) \xrightarrow[]{m \to \infty} \min_{\mathbb Z \in \bD{}} \Psi(Z),
\end{align*}
since $2p$ is again the largest power of $\glsT{Xs}$ in both terms (remember $c_m$ has an $({\glsT{Xs}})^p$ term). The convergence of $\dk{k}(\hX{}, \gY{}^{\theta_m^\star}) \to 0$ follows as before, finishing the first part of the theorem. Finally, it is easy to see that  the proof of the convergence of the Monte Carlo approximation is not affected by our more general dependence assumptions such that also the second part of the theorem follows.
\end{proof}

The assumption that  $\gX{}_{\tk{i-}}$ and $\gn{}$ are conditionally independent given $\At{\tk{i-}}$ seems a bit odd, because $\gn{}$ is a quantity that is not known at $\tk{i}$, i.e., not $\At{\tk{i}}$-measurable. This assumption is needed, because we weight the observations in the loss by $1/{\gn{}}$, which means that we weight the terms known at time $\tk{i}$ with a quantity that is in general only known at final time $\gls{T}$.
In line with this, we can drop the assumption if instead we weight the terms with the time step $\Delta \tk{i} := \tk{i} - \tk{i-1}$, which is $\At{\tk{i}-}$-measurable. However, this needs an additional integrability assumption as explained in the following proposition. 

\begin{prop}\label{prop:dropping independece assumption for n}
    If we change the weighting in the loss function to 
    \begin{equation*}
        \Psi(Z) := \E\left[  \sum_{i=1}^{\gn{}} \Delta\tk{i} \left(  \left\lvert \gM{}_i \odot ( \gX{}_{\tk{i}} - Z_{\tk{i}} ) \right\rvert_2 + \left\lvert \gM{}_i \odot (\gX{}_{\tk{i}} - Z_{\tk{i}-} ) \right\rvert_2 \right)^2 \right],
    \end{equation*}
    then \Cref{ass:dependence}~\ref{ass_both_6} can be replaced by the weaker assumption \ref{ass_dep_6p} together with the additional integrability assumption \ref{ass_dep_6pp} below. 
    \begin{enumerate}[label=(vii'-\alph*)]
        \item  For every  $1 \le i \le \gn{}$, $\gX{}_{\tk{i}-}$ is conditionally independent of $\sigmab(\gM{\tk{i}})$ given $\At{\tk{i}-}$. \label{ass_dep_6p}
        \item For every  $1 \le i \le \gls{K}$, $\E\left[ \frac{1}{\Delta \tk{i}}  \right] < \infty$. \label{ass_dep_6pp}
    \end{enumerate}
\end{prop}

\begin{proof}
    First, we note that $\{ i \leq n \} = \{ \tk{i} < \infty \}$ and $\Delta \tk{i}$ are  $\At{\tk{i}-}$-measurable, hence the proofs of \Cref{lem:L2 identity dependence} and \Cref{lem:ck} work equivalently under the weaker assumption \ref{ass_dep_6p}.
    Next we note that \eqref{equ:HI} has to be replaced in the analysis by 
    \begin{equation}\label{equ:HI prime}
        \E\left[ \left\lvert Z \right\rvert_2 \right] 
	   = \E\left[ \frac{\sqrt{\Delta\tk{i}}}{\sqrt{\Delta\tk{i}}} \, \left\lvert Z \right\rvert_2 \right] 
	   \leq c_1' \, \E\left[ \Delta\tk{i} \,  \left\lvert Z \right\rvert_2^2 \right]^{1/2}
    \end{equation}
    where $c_1':= c_1'(i) := \E [ 1 / {\Delta\tk{i}} ] < \infty$ holds by the additional assumption \ref{ass_dep_6pp}.
    Finally, we note that $\sum_{i=1}^{\gn{}} \Delta\tk{i} \leq \gls{T}$, hence integrability of the loss is still satisfied. These are all needed changes in the proof.
\end{proof}

\begin{rem}\label{rem:different weigthings}
    The different weightings in the loss function by either $1/{\gn{}}$ or $\Delta \tk{i}$ put different importance on the observations. In the former case, every observation of one sample path is given equal importance, while in the latter case, observations have more importance if the previous observation is farther away. Although both loss functions yield the same unique optimizer, the choice of weighting can influence the training of the model for finite $m$ or $N$. In particular, the latter choice corresponds to an inductive bias, which makes it more important for the model to predict well after a long time without observations, while it is less important to predict well whenever the frequency of observations is high. Of course other weightings, e.g.~by $1/\E[\gn{}]$, are possible, too, and reasonable in some situations.
\end{rem}

\subsection{Examples of Processes Satisfying the Assumptions}\label{sec:Examples - dependence setting}
Clearly, all the examples from \cite{krach2022optimal} are trivially valid for our generalised settings of Section~\ref{sec:Setting with Dependence} since points~\ref{ass_condindep} \& \ref{ass_newass} of Assumption~\ref{ass:dependence} are implied by independence.
Furthermore, we provide a relatively general class of examples in Section~\ref{sec:Class of Examples Incorporating Dependence} and extend two of the examples from \cite{krach2022optimal} to a setting where independence does not hold in Example~\ref{sec:Extending the homogeneous Poisson point process} and Example~\ref{exa:Black--Scholes with Dependent Observations}.

\subsubsection{Class of Examples Incorporating Dependence}\label{sec:Class of Examples Incorporating Dependence}
A main problem when constructing examples where the observation times $\tk{i}$ have a dependence on the process $\gX{}$,  is that in general this will  also lead to $\gn{}$ having a dependence on $\gX{}$ (since $\gn{}$, as the amount of observations, increases when observations become more likely). This, in turn, might lead to a contradiction of Assumption~\ref{ass:dependence} point \ref{ass_condindep}. 

One way to circumvent this is to define $\gn{}$ conditionally independent of $\gX{}_{\tk{i}-}$ given $\At{\tk{i}-}$ for all $i$ and to allow for pseudo observation times, i.e., observation times at which no coordinate is observed (cf.\ Remark~\ref{rem:at least one coord observed not needed}).  Then we can control whether the process $\gX{}$ is observed at an observation time $\tk{i}$ via the observation mask $\gM{}_i$.
In this way, the times at which $\gX{}$ is actually observed can depend on $\gX{}$ through the observation mask~$\gM{k}$, even though the $\tk{k}$ do not depend on $\gX{}$.
Therefore, the original problem is replaced by having an observation mask depending on $\gX{}$, which will be discussed in detail below.

When using a time grid on which the process $\gX{}$ is sampled, one concrete example of defining $\gn{}$  conditionally independent of $\gX{}_{\tk{i}-}$ given $\At{\tk{i}-}$ is to use the grid points as observation times $\tk{i}$ leading to $\gn{}$ being the number of grid points. In this case, both $\gn{}$ and the $\tk{i}$ are deterministic, hence, they satisfy the conditional independence assumptions.

We need to ensure that Assumption~\ref{ass:dependence} points \ref{ass_condindep} and \ref{ass_newass} are satisfied (assuming that a dummy variable is observed at every observation time), when defining the observation mask.
One way to define $\gM{}_i$ such that point \ref{ass_condindep} is satisfied is to define it as a function of random variables that are $\At{\tk{i}-}$-measurable and random variables that are independent of $\sigmab(\At{\tk{i}-}, \gX{}_{\tk{i}-}, \gn{})$. 

In particular, let $\gM{}_i := g_i\left( (U^i_j)_{j \in J^i_1}, (V^i_j)_{j \in J^i_2} \right)$, where $g_i$ is a measurable function mapping to $\{0,1\}^{\gls{dX}}$; $J^i_1, J^i_2 \subseteq \N$ are index sets; $U^i_j$ is a $\At{\tk{i}-}$-measurable random variable; and $V^i_j$ a random variable independent of $\sigmab(\At{\tk{i}-}, \gX{}_{\tk{i}-}, \gn{})$ for all $j$ in $J^i_1$ and $J^i_2$ respectively.

By Proposition~\ref{prop:conditional independence prop 3} we need to show that for any measurable function $\phi$ we have $\cexpect{\phi(\gX{}_{\tk{i}-})}{\sigmab(\gn{}, \gM{}_i, \At{\tk{i}-})} = \cexpect{\phi(\gX{}_{\tk{i}-})}{ \At{\tk{i}-} }$.
Indeed, $\gM{}_i$ is $\sigmab(\At{\tk{i}-}, (V^i_j)_{j \in J^i_2})$-measurable  by construction, therefore, we have for such a $\phi$ that
\begin{equation*}
    \cexpect{\phi(\gX{}_{\tk{i}-})}{\sigmab(\gn{}, \gM{}_i, \At{\tk{i}-})} 
    = \cexpect{ \cexpect{\phi(\gX{}_{\tk{i}-})}{\sigmab(\gn{}, (V^i_j)_{j \in J_2^i}, \At{\tk{i}-})} }{\sigmab(\gn{}, \gM{}_i, \At{\tk{i}-})},
\end{equation*}
by the tower property.
On the other hand, 
\begin{align*}
    \cexpect{\phi(\gX{}_{\tk{i}-})}{\sigmab(\gn{}, (V^i_j)_{j \in J_2^i}, \At{\tk{i}-})} 
    = \cexpect{\phi(\gX{}_{\tk{i}-})}{\sigmab(\gn{}, \At{\tk{i}-})} 
    = \cexpect{\phi(\gX{}_{\tk{i}-})}{\At{\tk{i}-} },
\end{align*}
using the independence of $V_j^i$ together with Corollary~\ref{cor:conditional independence prop 1.2}  in the first equality and $\gn{}$ being conditionally independent together with Proposition~\ref{prop:conditional independence prop 3} in the second equality.
Together, this implies the claim and therefore Assumption~\ref{ass:dependence} point \ref{ass_condindep}.

For Assumption~\ref{ass:dependence} point \ref{ass_newass}, we note that by \citet[Lemma~6.2.1]{Durrett:2010:PTE:1869916} we have
\begin{align*}
    \glsP{OmFFP} ( \gM{k} = 1 \mid \sigmab(\gn{}, A_{\tk{k}-} )) 
    &= \cexpect{\gM{\tk{k}}}{\sigmab(\gn{}, \At{\tk{k}-})} \\
    &= \cexpect{g_k\left( (U^k_j)_{j \in J^k_1}, (V^k_j)_{j \in J^k_2} \right)}{\sigmab(\gn{}, \At{\tk{k}-})} \\
    &= \tilde{g}_k((U^k_j)_{j \in J^k_1}),
\end{align*}
for $\tilde{g}_k((\uBlack{}^k_j)_{j \in J^k_1}) := \E\left[g_k\left( (\uBlack{}^k_j)_{j \in J^k_1}, (V^k_j)_{j \in J^k_2} \right)\right]$.
Hence, we need to define the $g_k$ and $V_j^k$ such that $\tilde{g}_k > \eta_k$ (coordinate-wise) for some $\eta_k >0$.

\begin{example}[Homogeneous Poisson Point Process with Dependent Observations]\label{sec:Extending the homogeneous Poisson point process}
We use a $1$-dimensional homogeneous Poisson point process $\gX{} = N^\lambda$  \cite[Section 7.3]{krach2022optimal} and defined observations depending on its value following the instructions above. To begin with, we define the observation times to be the grid points of the sampling grid of the process and $\gn{}$ accordingly to be the number of grid points. To permit observation times of the process depending on the process, we define the observation mask as 
\begin{align*}
    \gM{}_i = 
    \begin{cases}
         \ind{\{x_i \ge \lambda \tk{i-1} \}}, \ x_i \sim \normalDistribution{}(N^\lambda_{\tk{i-1}}, \sigma^2) &\quad \text{if $\gM{}^{\gX{}}_{i-1} = 1$}, \\
         \uBlack{}_i \sim \operatorname{Bernoulli}(p) &\quad \text{if $\gM{}^{\gX{}}_{i-1} = 0$}.
    \end{cases}
\end{align*}
for some $\sigma > 0$ and $p \in (0,1)$. Thus the process is more likely to be observed if the previous value was observed and was large (note $\expect{N^\lambda_{\tk{k}}} = \lambda \tk{k}$ for all $k$). Otherwise the mask value is sampled from a Bernoulli distribution. 

To satisfy Assumption~\ref{ass:dependence} point~\ref{ass_dependence_1}, we define $\gM{}_0 := 1$. 
To see that Assumption \ref{ass:dependence} point \ref{ass_condindep} holds, let $V^i_1 \sim \operatorname{Bernoulli}(p)$ and $V^i_2 \sim \normalDistribution(0, \sigma^2)$ be independent random variables independent of $\sigmab(\At{\tk{i}-}, N^\lambda_{\tk{i}-})$. Then 
\[\gM{}_i = \ind{\{\gM{}_{i-1} = 1\}} \ind{\{N^\lambda_{\tk{i-1}} + V^i_2 \ge \lambda \tk{i-1}\}} + V^i_1 \ind{\{\gM{}_{i-1} = 0\}} =: g_i(\gM{}_{i-1}, N^\lambda_{\tk{i-1}}, \tk{i-1}, V^i_1, V^i_2),\]
and the claim follows as explained above.
Moreover, Assumption~\ref{ass:dependence} point \ref{ass_newass} is satisfied because 
\begin{align*}
    \Tilde{g}_i(\gM{}_{i-1}, N^\lambda_{\tk{i-1}}, \tk{i-1}) 
    &= \ind{\{\gM{}_{i-1} = 1\}} \, \E\left[ \ind{\{a + V^i_2 \ge \lambda b \}} \right] \Big\vert_{(a,b)=\left(N^\lambda_{\tk{i-1}}, \tk{i-1}\right)} + \ind{\{\gM{}_{i-1} = 0\}} \, \E\left[ V^i_1 \right] \\
    & \geq \min\left( \glsP{OmFFP}\left[ V^i_2 \ge \lambda b - a \right] \Big\vert_{(a,b)=\left(N^\lambda_{\tk{i-1}}, \tk{i-1}\right)}, p \right) \\
    & \geq  \min\left( \glsP{OmFFP}\left[ V^i_2 \ge \lambda \gls{T} \right], p \right) =: \eta_i > 0,
\end{align*}
using that $\lambda \tk{i-1} - N^\lambda_{\tk{i-1}} \leq \lambda \gls{T}$, since $N^\lambda \geq 0$, $\tk{i-1} \leq \gls{T}$ and $\lambda > 0$, in the last line.
\end{example}

\begin{rem}
    We note that the choice $\gX{} = N^\lambda$ is only explicitly used for the fact that $N^\lambda$ is lower bounded (by $0$). Hence, the Poisson point process could be replaced by any other process that is lower bounded and satisfies Assumption~\ref{assumption:F} points \ref{ass_main_4} \& \ref{ass_main_5}.
\end{rem}

\begin{example}[Black--Scholes with Dependent Observations]
\label{exa:Black--Scholes with Dependent Observations}
    We use a $1$-dimensional Black--Scholes process (geometric Brownian motion) \cite[Example~7.3]{krach2022optimal} with constant drift and volatility $\mu, \sigma$ starting at $x_0$ and again define observations depending on its value. The observation times and $\gn{}$ are defined as in the previous example and we define the actual times when $\gX{}$ is observed via the mask $\gM{}$. We set $\gM{}_0=1$. Moreover, we redefine $\tau$ to be the last time before $t$ at which $\gX{}$ was observed, i.e., $\tau(t) = \max\{ \tk{i} \, | \, \tk{i} < t, \gM{}_i = 1\}$.
    Let $V^i_1 \sim \operatorname{Bernoulli}\left(\frac{\tk{i} - \tk{i-1}}{\tk{i} - \tau(\tk{i})}\right)$, $V^i_2 \sim \normalDistribution{}(0, \eta^2)$ and $V^i_3 \sim \operatorname{Bernoulli}\left(p\right)$ be independent random variables for some $\eta > 0$ and $p \in (0,1)$.
    Then we define 
    \begin{equation*}
        \gM{}_i := V^i_1 \, \1_{\{ \gX{}_{\tau(\tk{i})} + V^i_2 \geq x_0 e^{\mu \tk{i}} \} } + (1-V^i_1) \, V^i_3,
    \end{equation*}
    for all $i \geq 1$. In particular, if $\gX{}$ was observed at the previous observation time $V^i_1=1$ and the probability of observing $\gX{}$ increases with the size of the last observation of $\gX{}$ (compared to the current expected value of $\gX{}$ at the current time). The further the last observation of $\gX{}$ is in the past, the more likely $V_1^i = 0$ in which case $\gX{}$ is observed with probability $p$.
    Upon noting that the $\tk{i}$ are deterministic, it follows as in the previous example that Assumption~\ref{ass:dependence} points \ref{ass_condindep} and \ref{ass_newass} are satisfied.
\end{example}

\section{Practical Implications of the Convergence Result}\label{sec:Practical Implications of the Convergence Result}
In this section we discuss which practically relevant conclusions we can draw from convergence in the metrics $\dk{k}$ for $1\leq k \leq \gls{K}$. We mainly focus on the setting in Section~\ref{sec:Setting with Dependence}, however, the same is also true in the combined setting of Section~\ref{sec:Setting both ext}.
In Section~\ref{sec:Practically Relevant Version of the Conditional Expectation}, we study the practical meaning of \Cref{ass:dependence}\ref{ass:cond ind X tk}. We give an intuitive counterexample, where this assumption is not satisfied and our model does not converge to $\hX{}$.
Secondly, in Section~\ref{sec:From Approximations at Observation Times to Approximations at any Time}, we discuss the implications of convergence in the (pseudo) metrics $\dk{k}$ (which ensures a good approximation at left-limits of observation times) for general times $t$. In particular, we study two practically relevant examples for the conditional distribution of the observation times and show that in these cases, with high probability, our model approximates the conditional expectation well on the entire support of the observation times.

\subsection{Practical meaning of Assumption~\ref{ass:dependence}\ref{ass:cond ind X tk}}\label{sec:Practically Relevant Version of the Conditional Expectation}
If Assumption~\ref{ass:dependence}\ref{ass:cond ind X tk} is not satisfied, this can lead to situations where $\hX{}$ is not the minimizer of $\Psi$ and therefore our model does not converge to it, as we explain in the following example.
Assume that patients in a hospital get asked by a nurse every morning at 8 am, whether they feel feverish. If the answer is yes, then their temperature is measured right away and logged at 8 am; if the answer is no, their temperature is measured and logged at 4 pm. The (non-)feverish feeling is not logged at all. This data satisfies Assumption~\ref{ass:dependence} (since $\gn{}$ is deterministic; $\gls{dX}=1$ and $\gM{}_i=1$) except for \cref{ass:cond ind X tk}. 
It is clear that the model will predict higher temperatures in the morning than in the afternoon for any test sample, even if the patient always has the same temperature, since the model only saw this behaviour in the training data. 
If we assume that on average patients have the same (or even lower) temperature in the morning as in the afternoon, this prediction is not optimal in practice.

The mathematical reason for this discrepancy is that our model learns to approximate $\cexpect{\gX{}_{\tk{k}-}}{\At{\tk{k}-}}$ (see proof of \Cref{lem:L2 identity dependence}), which does not coincide with $\hX{}$, i.e.,
\begin{equation*}
    \cexpect{\gX{}_{\tk{k}-}}{\At{\tk{k}-}}
    \neq
    \cexpect{\gX{}_{t}}{\At{t}} \Big\vert_{t=\tk{k}-} 
    = \cexpect{\gX{}_{t}}{\At{\tk{k-1}}} \Big\vert_{t=\tk{k}-}
    =\hX{}_{\tk{k}-}
\end{equation*}
(surprisingly) holds for this example.
Indeed, \Cref{prop:rewriting of hat X} (which would turn this inequality into an equality) heavily relies on Assumption~\ref{ass:dependence}\ref{ass:cond ind X tk}.
$\cexpect{\gX{}_{\tk{k}-}}{\At{\tk{k}-}}$ minimizes the test error (if the joint distribution of $(\gX{},\tk{k})$ is the same as during training), since it can exploit the knowledge\footnote{Mathematically, $\At{\tk{k}-}=\sigmab(\At{\tk{k-1}},\tk{k})$ includes the information that a measurement is taken at time~$\tk{k}$.} that it is queried at an observation time $\tk{k}$ (i.e., high temperature if $\tk{k}$ is in the morning and low temperature if $\tk{k}$ is in the afternoon).
However, in this example we do \emph{not} obtain the optimal forecast~$\hX{t}=\cexpect{\gX{}_{t}}{\At{\tk{k-1}}}$ for any time $t$, which is the best prediction of $\gX{}$ at any time, given the information available at the previous observation time.

\Cref{ass:dependence}\ref{ass:cond ind X tk} ensures (via \Cref{prop:rewriting of hat X}) that 
our model converges to the practically meaningful version of the conditional expectation~$\hX{\tk{k}}= \cexpect{\gX{}_{t}}{\At{\tk{k-1}}} \big\vert_{t=\tk{k}-}$ at observation times.
In the following, we discuss what this implies for the approximation at any $t \in [0,\gls{T}]$.

\subsection{From Approximations at Observation Times to Approximations at any Time}\label{sec:From Approximations at Observation Times to Approximations at any Time}
Under Assumption~\ref{ass:dependence} we have established that our model output $\gY{t}^{\thetamMin{m}}(\tildeXle{\tk{k}-})$ converges to $\cexpect{\gX{}_{t}}{\At{\tk{k-1}}}$ at every observation time $\tk{k}$.
Both, $\gY{t}^{\thetamMin{m}}(\tildeXle{\tk{k}-})$ and $\cexpect{\gX{}_{t}}{\At{\tk{k-1}}}$ are functions of the random variables summarized in $\At{\tk{k-1}}$ (by definition for our model and by the Doob-Dynkin lemma for the conditional expectation) and therefore conditionally independent of $\tk{k}$ given $\At{\tk{k-1}}$. We note in particular that $\tildeXle{\tk{k}-}$ carries exactly the information of $\At{\tk{k-1}}$, i.e., it has no information about $\tk{k}$ (despite its notation).
Hence, intuitively speaking, the approximation has to be good for every $t$ in the support of $\tk{k}$ for the expectation to converge. In the following we formalize this.

Let $\P_k$ be the probability measure conditioned on the event that $\gn{} \geq k$, i.e. 
\begin{equation*}
    \P_k (\cdot) = \P(\cdot \,  | \{\gn{} \geq k\}),
\end{equation*}
and denote by $\E_k$ the expectation operator with respect to this probability measure.
Then the following lemma is a consequence of the definitions of $\P_k$ and $c_0(k)$.

\begin{lemma}\label{lem:rewriting of dk}
    We have $\dk{k} (\hX{}, \gY{}^{\thetamMin{m}} ) = \E_k \left[  \left| \hX{}_{\tk{k}-} - \gY{}^{\thetamMin{m}}_{\tk{k}-} \right|_2 \right]$ and the equivalent result for $\gY{}^{\thetamNMin{m,N_{m}}}$. 
\end{lemma}
\begin{proof}
    It is enough to note that $\E[ \cdot \, \ind{ \{\gn{} \geq k \}}] = \P(\gn{} \geq k) \, \E_k[\cdot]$. 
\end{proof}

Next, we define for each $1 \leq k \leq \gls{K}$ the (regular) conditional distribution of $\tk{k}$ given $\tildeXle{\tk{k}-}$ as the almost surely unique probability kernel (or random measure) $\mu_{\tk{k}}(\cdot \, ; \tildeXle{\tk{k}-})$ satisfying 
\begin{equation*}
    \mu_{\tk{k}}( B \, ; \tildeXle{\tk{k}-}) = \P_k \left( \tk{k} \in B \, \middle| \, \At{\tk{k-1}} \right) \quad a.s., \quad \forall \, B \in \mathcal{B}([0, \gls{T}]).
\end{equation*}
\citet[Theorem~8.5]{kallenberg1997foundations} ensures its existence (since $[0, \gls{T}]$ is Borel) and implies for any measurable function $\phi$ that almost surely
\begin{equation}\label{equ:writing cond exp as integral wrt cond dist}
    \E_k\left[ \phi\left( \tildeXle{\tk{k}-}, \tk{k} \right) \, \middle\vert \, \At{\tk{k-1}} \right] 
    = \int_0^{\gls{T}} \phi\left( \tildeXle{\tk{k}-}, t \right) \, \mu_{\tk{k}}( dt \, ; \tildeXle{\tk{k}-}).
\end{equation}
In particular, this implies the following result.
\begin{prop}\label{prop:rewriting of convergence in dk}
    Under Assumption~\ref{ass:dependence} we have, with the same notation as in Theorem~\ref{thm:1}, that for every $1 \leq k \leq \gls{K}$
    \begin{equation*}
        \dk{k} (\hX{}, \gY{}^{\thetamMin{m}} ) = \E_k \left[  \int_0^{\gls{T}} \left| \E \left[ \gX{}_{t-} \middle| \tildeXle{\tk{k}-} \right] - \gY{}^{\thetamMin{m}}_{t-}\left(\tildeXle{\tk{k}-}\right) \right|_2 \, \mu_{\tk{k}}( dt \, ; \tildeXle{\tk{k}-}) \right] \xrightarrow{m \to \infty} 0
    \end{equation*}
    and the equivalent result for $\gY{}^{\thetamNMin{m,N_{m}}}$.
\end{prop}
\begin{proof}
    We first apply Lemma~\ref{lem:rewriting of dk} and then use the tower property to get a nested conditional expectation (conditioning on $\At{\tk{k-1}}$) inside the outer expectation, which can be rewritten by \eqref{equ:writing cond exp as integral wrt cond dist} as an integral. 
    Finally, we use Lemma~\ref{prop:rewriting of hat X} and then the Doob-Dynkin lemma \citep[Lemma~2]{taraldsen2018optimal} to rewrite $\E[ {\gX{}_{t}} \, | \, \At{\tk{k-1}} ]$ as a function of $\tildeXle{\tk{k}-}$, which we denote by $\E [ {\gX{}_{t}} \, | \, {\tildeXle{\tk{k}-} } ]$ (and we replace $t$ by $t-$ for the left-continuous version of it).
\end{proof}
Proposition~\ref{prop:rewriting of convergence in dk} implies convergence of a conditional $L^1$-norm (or equivalently, an $L^1$-norm integrated against a random measure) nested inside an $L^1$-norm. 
Since $L^p$-convergence implies convergence in probability, we have in particular that for every $\epsilon > 0$ 
\begin{equation}\label{equ:conv in prop}
    \P_k\left( \int_0^{\gls{T}} \left| \E \left[ \gX{}_{t-} \middle| \tildeXle{\tk{k}-} \right] - \gY{}^{\thetamMin{m}}_{t-}\left(\tildeXle{\tk{k}-}\right) \right|_2 \, \mu_{\tk{k}}( dt \, ; \tildeXle{\tk{k}-}) > \epsilon \right) \xrightarrow{m \to \infty} 0.
\end{equation}
In the following, we study two examples of the conditional distribution $\mu_{\tk{k}}$ that are relevant in practice. We study them for $\gY{}^{\thetamNMin{m}}$ but note that the hold equivalently for $\gY{}^{\thetamNMin{m,N_{m}}}$.
We first remark that in general the support of $\tk{k}$ is a $\At{\tk{k-1}}$-measurable random set (in particular, it often depends at least on $\tk{k-1}$). We denote this set-valued random variable with
\begin{equation*}
    S_{\tk{k}} := \supp \left( \tk{k} \, | \, \At{\tk{k-1}} \right),
\end{equation*}
which is defined to be the smallest closed random subset $S_{\tk{k}} \subset [0, \gls{T}]$ such that $\mu_{\tk{k}}( S_{\tk{k}} \, ; \tildeXle{\tk{k}-}) = 1$ $\P_k$-almost surely.
By Doob-Dynkin's lemma \citep[Lemma~2]{taraldsen2018optimal} we can write $S_{\tk{k}}$ as a function $\tildeXle{\tk{k}-}$, i.e., $S_{\tk{k}} = S_{\tk{k}}(\tildeXle{\tk{k}-})$.
\begin{example}\label{exa:finite grid for tks}
    Assume that $S_k := \cup_{\omega} S_{\tk{k}}\omb$ is finite and that there exists some $\alpha > 0$ such that for all $t \in S_k $ we have $\mu_{\tk{k}}( \{ t \} \, ; \tildeXle{\tk{k}-}) \geq \alpha \ind{\{ t > \tk{k-1}\}}$ $\P_k$-almost surely.
    This is the case if all observation times are sampled from a (finite) grid with positively lower bounded probability to take any value of the grid, which is larger than the previous observation time (as in the synthetic examples in this paper and in \citet{krach2022optimal}).
    Then Proposition~\ref{prop:rewriting of convergence in dk} implies that for every $1\leq k \leq \gls{K}$
    \begin{equation*}
        \sum_{t \in S_k } \E_k \left[ \ind{\{ t > \tk{k-1}\}} \left| \E \left[ \gX{}_{t-} \middle| \tildeXle{\tk{k}-} \right] - \gY{}^{\thetamMin{m}}_{t-}\left(\tildeXle{\tk{k}-}\right) \right|_2  \right] \xrightarrow{m \to \infty} 0
    \end{equation*}
    and \eqref{equ:conv in prop} implies that for every $\epsilon, \delta > 0$ there exists $m_0 \in \N$ such that for $m \geq m_0$ there is a subset $\Omega_m \subset \Omega$ with $\P_k(\Omega_m) > 1-\delta$ and for every $\omega \in \Omega_m$ with $\gn{}\omb \geq k$ and every $t \in S_k$ with $t > \tk{k-1}(\omega)$
    \begin{equation*}
        \left| \E \left[ \gX{}_{t-} \middle| \tildeXle{\tk{k}-}\omb \right] - \gY{}^{\thetamMin{m}}_{t-}\left(\tildeXle{\tk{k}-}\omb\right) \right|_2 < \epsilon.
    \end{equation*}
    In particular, with high probability, our model is close to the optimal prediction at the left limit of every possible grid point if $m$ is large enough. 
    At the same time it is apparent that we cannot infer anything for $t \notin S_k$. In our synthetic examples this is not a problem since we only plot our model and the optimal prediction (and measure the distance between them) on the grid points. 
\end{example}
\begin{rem}[Convergence in the Evaluation Metric]
    In the same setting as in Example~\ref{exa:finite grid for tks}, if there exists one finite grid from which all observation times are sampled, i.e., $S \subset [0,\gls{T}]$ finite, such that $S_k \subset S$ for all $k$ and $T \in S$, then $n \leq |S|$. 
    Revisiting the proof of Theorem~\ref{thm:1} in Appendix~\ref{sec:Combining the Two Extensions With Full Proof}, we note that \eqref{equ:HI} can therefore be replaced by 
    \begin{equation*}
        \E\left[ \left\lvert Z \right\rvert_2^2 \right] 
	   \leq |S| \, \E\left[ \frac{1}{\gn{}} \left\lvert Z \right\rvert_2^2 \right],
    \end{equation*}
    and therefore we can show the convergence stated in Theorem~\ref{thm:1} in the stronger $L^2$-type (pseudo) metric
    \begin{equation}\label{equ:pseudo metric d2k}
    d_{k}^2 (Z, \xi) := c_0(k)\,  \E \left[ \1_{\{\gn{} \geq k\}} | Z_{\tk{k}-} - \xi_{\tk{k}-} |_2^2 \right], 
    \end{equation}
    for which Proposition~\ref{prop:rewriting of convergence in dk} holds equivalently.
    
    We assume that for $\P$-a.e.\ $\omega$ and for every $t \in S$ we have that $\tk{k-1}\omb < t \leq \tk{k}\omb$ implies that $t \in S_{\tk{k}}\omb$.
    Then the evaluation metric (cf.\ Section~\ref{sec:Experiments}) defined on $S$ for $N_{\text{test}}$ i.i.d.\ test samples is 
    \begin{equation*}
        \operatorname{eval}_{S, N_{\text{test}}}(\hX{}, \gY{}^{\thetamMin{m, N_m}}) := \frac{1}{|S|} \sum_{t \in S} \frac{1}{N_{\text{test}}}\sum_{j=1}^{N_{\text{test}}} \left|  \E \left[ \gX{}_{t-} \middle| \tildeXle{t-, (j)} \right] - \gY{}_{t-}^{\thetamMin{m, N_m}}(\tildeXle{t-, (j)}) \right|_2^2
    \end{equation*}
    and by a similar argument as in the second part of the proof of Theorem~\ref{thm:1} (cf.\ \eqref{equ:unif conv 1}, \eqref{equ:unif conv 2}), we have
    \begin{equation*}
        \operatorname{eval}_{S, N_{\text{test}}}(\hX{}, \gY{}^{\thetamMin{m, N_m}}) 
        \xrightarrow[N_{\text{test}} \to\infty]{\P-a.s.} 
        \E\left[ \frac{1}{|S|} \sum_{t \in S}  \left|  \E \left[ \gX{}_{t-} \middle| \tildeXle{t-} \right] - \gY{}_{t-}^{\thetamMin{m, N_m}}(\tildeXle{t-}) \right|_2^2 \right] =: (I).
    \end{equation*}
    If we add pseudo observation times at $\gls{T}$ (by choosing the respective mask to be $0$ in case there is no actual observation at $\gls{T}$, cf.\ Remark~\ref{rem:at least one coord observed not needed}), we have $\tk{n} = \gls{T}$.\footnote{%
    We need to do this because otherwise we have terms where $t > \tk{n}$ which cannot be upper bounded by the distances in $d_k^2$. However, we note that the objective function is the same except for dividing by $\gn{}+1$ instead of $\gn{}$ for the samples which do not have an observation at $\gls{T}$. Hence, also the trained model is nearly the same.%
    } 
    Then, using that $S$ is finite and with our assumption for $S_{\tk{k}}$, we have
    \begin{equation*}
    \begin{split}
        (I) &= \sum_{k=1}^{|S|}
        \E\left[ \ind{\{ \gn{}=k \}} \frac{1}{|S|} \sum_{t \in S}  \left|  \E \left[ \gX{}_{t-} \middle| \tildeXle{t-} \right] - \gY{}_{t-}^{\thetamMin{m, N_m}}(\tildeXle{t-}) \right|_2^2 \right] \\
        &= \frac{1}{|S|}  \sum_{k=1}^{|S|}
        \E\left[  \ind{\{ \gn{}=k \}} \sum_{\kappa = 1}^{k}   \sum_{t \in S_\kappa} \ind{\{ \tk{\kappa-1} < t \leq \tk{\kappa} \}} \left|  \E \left[ \gX{}_{t-} \middle| \tildeXle{\tk{\kappa}-} \right] - \gY{}_{t-}^{\thetamMin{m, N_m}}(\tildeXle{\tk{\kappa}-}) \right|_2^2 \right] \\
        & \leq \frac{1}{|S|}  \sum_{k=1}^{|S|}
        \E\left[ \ind{\{ \gn{} = k \}}  \sum_{\kappa = 1}^{k} \ind{\{ \gn{} \geq \kappa \}}  \sum_{t \in S_\kappa} \ind{\{ \tk{\kappa-1} < t \}} \left|  \E \left[ \gX{}_{t-} \middle| \tildeXle{\tk{\kappa}-} \right] - \gY{}_{t-}^{\thetamMin{m, N_m}}(\tildeXle{\tk{\kappa}-}) \right|_2^2 \right] \\
        & \leq \frac{1}{|S|}  \sum_{k=1}^{|S|}
        \E\left[  \sum_{\kappa = 1}^{|S|} \ind{\{ \gn{} \geq \kappa \}}  \sum_{t \in S_\kappa} \ind{\{ \tk{\kappa-1} < t \}} \left|  \E \left[ \gX{}_{t-} \middle| \tildeXle{\tk{\kappa}-} \right] - \gY{}_{t-}^{\thetamMin{m, N_m}}(\tildeXle{\tk{\kappa}-}) \right|_2^2 \right] \\
        & \leq \sum_{\kappa = 1}^{|S|} \sum_{t \in S_\kappa}
        \E\left[   \ind{\{ \gn{} \geq \kappa \}}   \ind{\{ \tk{\kappa-1} < t \}} \left|  \E \left[ \gX{}_{t-} \middle| \tildeXle{\tk{\kappa}-} \right] - \gY{}_{t-}^{\thetamMin{m, N_m}}(\tildeXle{\tk{\kappa}-}) \right|_2^2 \right] \\
        & \leq \sum_{\kappa = 1}^{|S|} \sum_{t \in S_\kappa}
        \E_\kappa \left[   \ind{\{ \tk{\kappa-1} < t \}} \left|  \E \left[ \gX{}_{t-} \middle| \tildeXle{\tk{\kappa}-} \right] - \gY{}_{t-}^{\thetamMin{m, N_m}}(\tildeXle{\tk{\kappa}-}) \right|_2^2 \right] \xrightarrow{m \to \infty} 0,
    \end{split}
    \end{equation*}
    which converges to $0$ according to Example~\ref{exa:finite grid for tks}.
    In particular, we have $\P$-almost surely
    \begin{equation*}
        \lim_{m\to\infty} \lim_{N_{\text{test}} \to \infty} \operatorname{eval}_{S, N_{\text{test}}}(\hX{}, \gY{}^{\thetamMin{m, N_m}}) = 0,
    \end{equation*}
    i.e., the evaluation metric converges to $0$ when the number of evaluation samples, the number of training samples and the network sizes increase. Moreover, if the number of evaluation samples $N_{\text{test}}<\infty$ is fixed, the evaluation metric~$\operatorname{eval}_{S, N_{\text{test}}}(\hX{}, \gY{}^{\thetamMin{m, N_m}})$ converges to zero in probability as $m$ and $N_m$ tend to infinity.
\end{rem}

\begin{example}\label{exa:mu tk absolutely cont wrt lebesgue}
    Assume that for $1 \leq k \leq \gls{K}$, $\P_k$-almost surely we have that $S_{\tk{k}} = (\tk{k-1}, \gls{T}]$, that $\mu_{\tk{k}}( \cdot \, ; \tildeXle{\tk{k}-})$ is absolutely continuous with respect to the Lebesgue measure on $[0,\gls{T}]$, and that its Radon-Nikodym derivative w.r.t.\ the Lebesgue measure satisfies 
    \begin{equation*}
        \frac{ d \mu_{\tk{k}}( \cdot \, ; \tildeXle{\tk{k}-}) }{ dt } \geq \alpha \ind{(\tk{k-1}, T] }(\cdot)
    \end{equation*}
    for some $\alpha > 0$.
    This is for example the case if the observation time $\tk{k}$ is uniformly or exponentially (with an upper bound) distributed on $(\tk{k-1}, \gls{T} ]$.
    Then Proposition~\ref{prop:rewriting of convergence in dk} implies that
    \begin{equation*}
        \E_k \left[  \int_{\tk{k-1}}^{\gls{T}} \left| \E \left[ \gX{}_{t-} \middle| \tildeXle{\tk{k}-} \right] - \gY{}^{\thetamMin{m}}_{t-}\left(\tildeXle{\tk{k}-}\right) \right|_2 \, dt \right] \xrightarrow{m \to \infty} 0
    \end{equation*}
    and \eqref{equ:conv in prop} implies that for every $\epsilon, \delta > 0$ there exists $m_0 \in \N$ such that for $m \geq m_0$ there is a subset $\Omega_m \subset \Omega$ with $\P_k(\Omega_m) > 1-\delta$ and for every $\omega \in \Omega_m$ with $\gn{}\omb \geq k$ 
    \begin{equation*}
        \int_{\tk{k-1}}^{\gls{T}} \left| \E \left[ \gX{}_{t-} \middle| \tildeXle{\tk{k}-}\omb \right] - \gY{}^{\thetamMin{m}}_{t-}\left(\tildeXle{\tk{k}-}\omb\right) \right|_2 \, dt < \epsilon.
    \end{equation*}
    In particular, with high probability, the $L^1$-distance on $(\tk{k-1}, \gls{T}]$ between our model and the optimal prediction is small if $m$ is large enough. Since our model and the optimal prediction are both continuous in $t$, we know that in the limit they have to be point-wise the same, i.e., for every $t \in (\tk{k-1}, \gls{T}]$. 
\end{example}
To summarize, under \Cref{ass:dependence}, at every time $s$, for every future time $t\in [s,T]$ that was not deterministically excluded from being the next observation time given the previous observations,\footnote{Mathematically this just means, \enquote{for every $t\in S_{\tk{k}}$} given that $\tau(s)=\tk{k-1}$ for some $k$.} our model $\gY{t}^{\thetamNMin{m,N_m}}(\tildeXle{s})$ converges in probability to the correct conditional expectation~$\cexpect{\gX{t}}{\At{s}}$ as $m$ tends to infinity.

\subsection{Implications for a Trained PD-NJ-ODE model}
Finally, we discuss what we can say about the distance between the true conditional expectation $\hX{}$ and the output of the PD-NJ-ODE model $\gY{}^{\theta^\star}$ for parameters $\theta^\star$ which are $\varepsilon$-optimal. 
This is an important practical question which arises, when training the model with some optimization scheme (e.g.\ a version of stochastic gradient descent) yields such parameters $\theta^\star$. We note that the existence of such parameters follows from \Cref{thm:1} or its versions for noisy observations (\Cref{thm:noisy obs}), dependence (\Cref{thm:depndence}) or both extensions (\Cref{thm:depndence&noise}).

\begin{prop}\label{prop:epsilon close parameters theta}
    Assume that $\theta^\star$ are $\varepsilon$-optimal parameters for the PD-NJ-ODE, i.e., 
    \[\Phi(\theta^\star) \leq \inf_{\theta\in\Theta}\Phi(\theta) + \varepsilon,
    \]
    for some $\varepsilon>0$.\footnote{Note that $\inf_{\theta\in\Theta}\Phi(\theta)=\min_{Z \in \mathbb{D}}\Psi(Z)=\Psi(\hat{X})$ according to our main theorem.} Then for every $1 \leq k \leq K$ there exists a constant $c = c(k) >0$ independent of $\theta^\star$ and $\varepsilon$ such that
    $\dk{k} (\hX{}, \gY{}^{\theta^\star} ) \leq c \sqrt{\varepsilon}$.
\end{prop}

\begin{proof}
    The claim follows from applying \eqref{equ:pseudo metric dk}, \eqref{equ:M split both}, \eqref{equ:HI} and finally Lemma~\ref{lem:L2 identity both ext}, which was summarized in \eqref{equ:convergence in L1} in the full proof of the main theorem with both extensions.
\end{proof}

We note that the results from \Cref{sec:From Approximations at Observation Times to Approximations at any Time} equivalently hold here. Therefore, we are able to make statements about the distance between the paths of $\hX{}$ and $\gY{}^{\theta^\star}$.

\begin{rem}
    Under the same assumption as in \Cref{prop:epsilon close parameters theta} as well as \Cref{ass:dependence}, we have that for every $1 \leq k \leq \gls{K}$
    \begin{equation*}
        \dk{k} (\hX{}, \gY{}^{\theta^\star} ) = \E_k \left[  \int_0^{\gls{T}} \left| \E \left[ \gX{}_{t-} \middle| \tildeXle{\tk{k}-} \right] - \gY{}^{\theta^\star}_{t-}\left(\tildeXle{\tk{k}-}\right) \right|_2 \, \mu_{\tk{k}}( dt \, ; \tildeXle{\tk{k}-}) \right] \leq c(k) \sqrt{\varepsilon}.
    \end{equation*}
    Moreover, if additionally the assumptions of \Cref{exa:finite grid for tks} are satisfied, then we have
    \begin{equation*}
        \sum_{t \in S_k } \E_k \left[ \ind{\{ t > \tk{k-1}\}} \left| \E \left[ \gX{}_{t-} \middle| \tildeXle{\tk{k}-} \right] - \gY{}^{\theta^\star}_{t-}\left(\tildeXle{\tk{k}-}\right) \right|_2  \right] \leq \frac{c(k)}{\alpha} \sqrt{\varepsilon}.
    \end{equation*}
    Equivalently, if additionally the assumptions of \Cref{exa:mu tk absolutely cont wrt lebesgue} are satisfied, then we have
    \begin{equation*}
        \E_k \left[  \int_{\tk{k-1}}^{\gls{T}} \left| \E \left[ \gX{}_{t-} \middle| \tildeXle{\tk{k}-} \right] - \gY{}^{\theta^\star}_{t-}\left(\tildeXle{\tk{k}-}\right) \right|_2 \, dt \right] \leq \frac{c(k)}{\alpha} \sqrt{\varepsilon}.
    \end{equation*}
\end{rem}

\section{Experiments}\label{sec:Experiments}
The code with all new experiments and those from  \citet{krach2022optimal} is available at \code. Further details about the experiments can be found in Appendix~\ref{sec:Experimental Details}. In particular, in Appendix~\ref{sec:Differences between the Implementation and the Theoretical Description of the PD-NJ-ODE} we give details on  the slight deviation of the practical implementation from the theoretical description.

As in \citet{krach2022optimal} we use the following evaluation metric to quantify and compare the training success.
\begin{equation*}\label{equ:evaluation metric}
\operatorname{eval}(\hX{}, \gY{}^{\theta}) := \frac{1}{N_{\text{test}}} \sum_{j=1}^{N_{\text{test}}}  \frac{1}{\kappa+1} \sum_{i=0}^{\kappa} \left|  \hX{}_{\frac{i \gls{T}}{\kappa}-}^{(j)} - \gY{}_{\frac{i \gls{T}}{\kappa}-}^{\theta, j } \right|_2^2,
\end{equation*}
where the outer sum runs over the test set of size $N_{\text{test}}$ and the inner sum runs over the equidistant grid points on the time interval $[0,\gls{T}]$.

In Sections~\ref{sec:Noisy Observations -- Brownian Motion with Gaussian Observation Noise} and~\ref{sec:Dependent Observation Framework -- a Black--Scholes Example} we provide two illustrative experiments on easy synthetic datasets for the extensions discussed in this paper. We note that the extension to a dependence between the observation framework and the process $\gX{}$ is purely theoretical, not changing anything in the implementation of the model, hence, the experiments on real world datasets of \citet{krach2022optimal} are representative for this extension as well. In particular, it is very likely that there actually is such a dependence in the Physionet dataset \citep{physionet}, as discussed in the introduction and in Section~\ref{sec:Intuition on independence assumptions}. Hence, our paper provides the theoretical foundation for those empirical results of \citet{krach2022optimal} and, vice versa, those results show that the method discussed in our paper is applicable to complex high-dimensional settings.

In Section~\ref{sec:Physionet with Observation Noise} we provide further experiments on the Physionet dataset to compare the noise-adapted training framework of Section~\ref{sec:PD-NJ-ODE with Noisy Observations} with the original one.

\subsection{Noisy Observations -- Brownian Motion with Gaussian Observation Noise}\label{sec:Noisy Observations -- Brownian Motion with Gaussian Observation Noise}
\begin{figure}[htp!]
\centering
\includegraphics[width=0.49\textwidth]{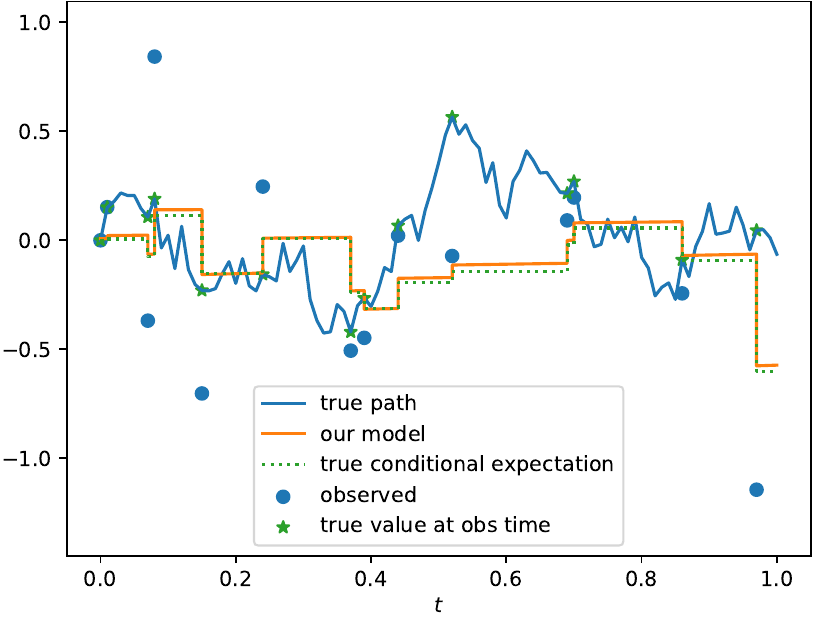}
\includegraphics[width=0.49\textwidth]{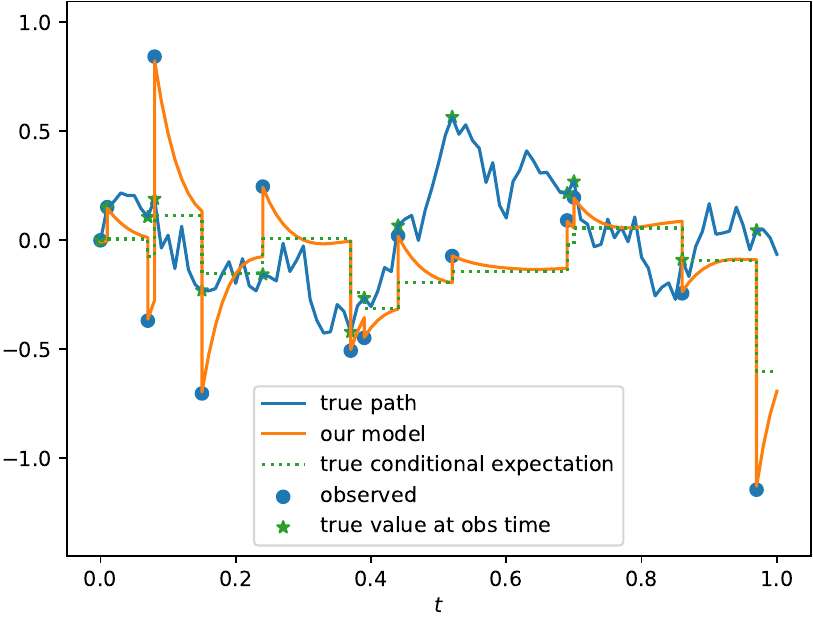}
\caption{A test sample of a Brownian motion $\gX{}$ with noisy observations $\gO{\gls{tk}} = \gX{}_{\gls{tk}} + \epsilon_i$ together with the true and predicted conditional expectation. The PD-NJ-ODE is trained with the noise-adapted loss (left) and the original loss (right).}
\label{fig:NoisyObs}
\end{figure}

We test the PD-NJ-ODE trained with the loss function adapted to noisy observations \eqref{equ:Psi noisy obs} in the context of Section~\ref{sec:Brownian Motion with Gaussian Observation Noise}. In particular, $\gX{}$ is a Brownian motion and we assume to have observation noise of a centered normal distribution with standard deviation $\sigma=0.5$. Moreover, we compare these results to using the original loss function \eqref{equ:Psi} with the noisy observation. PD-NJ-ODE adapted to noisy observations achieves a minimal evaluation metric of $1.1 \cdot 10^{-3}$ while using the original loss function leads to a nearly $20$ times larger evaluation metric of $1.9 \cdot 10^{-2}$. Moreover, in Figure~\ref{fig:NoisyObs} we see that the noise-adapted method learns to correctly jump when new observations become available, while the original method jumps to the noisy observations and afterwards tries to get close to the true conditional expectation quickly. We note that this is the expected behaviour.

\bigskip

\subsection{Dependent Observation Framework -- a Black--Scholes Example}\label{sec:Dependent Observation Framework -- a Black--Scholes Example}
\begin{figure}[bt!]
\centering
\includegraphics[width=0.7\textwidth]{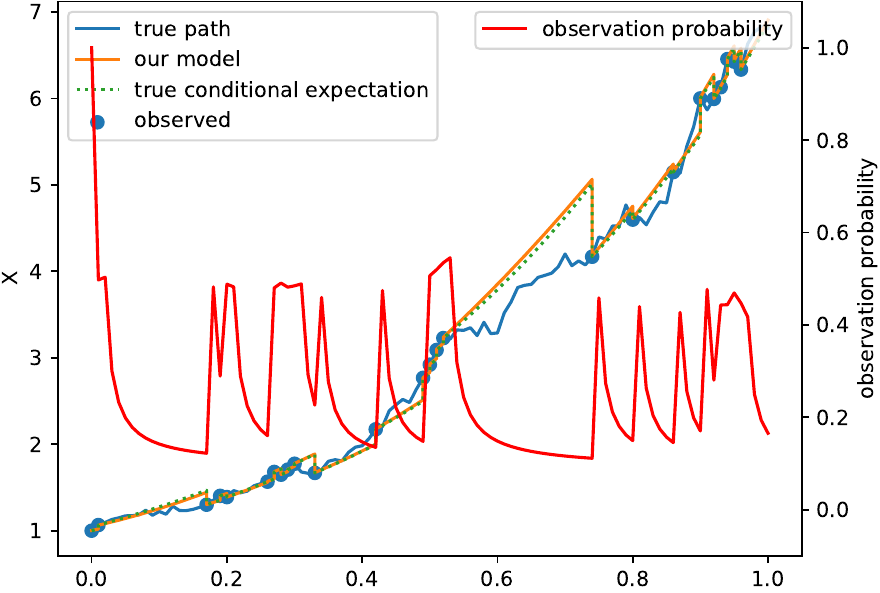}
\caption{A test sample of a geometric Brownian motion $\gX{}$ with observation probability depending on previously observed values of $\gX{}$. Plotted together with the true and predicted conditional expectation (scale on the left) and the observation probability over time (scale on the right).}
\label{fig:DepObs}
\end{figure}

Based on Example~\ref{exa:Black--Scholes with Dependent Observations} we train a PD-NJ-ODE on a $1$-dimensional geometric Brownian motion with drift $\mu=2$ and volatility $\sigma=0.3$ and with observation probability depending on the last observation of $\gX{}$, the time since the last observation and on independent random variables $V_{\{2,3\}}^i$ for which we use the parameters $\eta=3$ and $p=0.1$.
As our theoretical result suggests, our model learns to predict the conditional expectation well with a minimal evaluation metric of $1.1 \cdot 10^{-3}$ which is also visible in Figure~\ref{fig:DepObs}. 

\subsection{Physionet with Observation Noise}\label{sec:Physionet with Observation Noise}
\begin{figure}[bt!]
\centering
\includegraphics[width=0.5\textwidth]{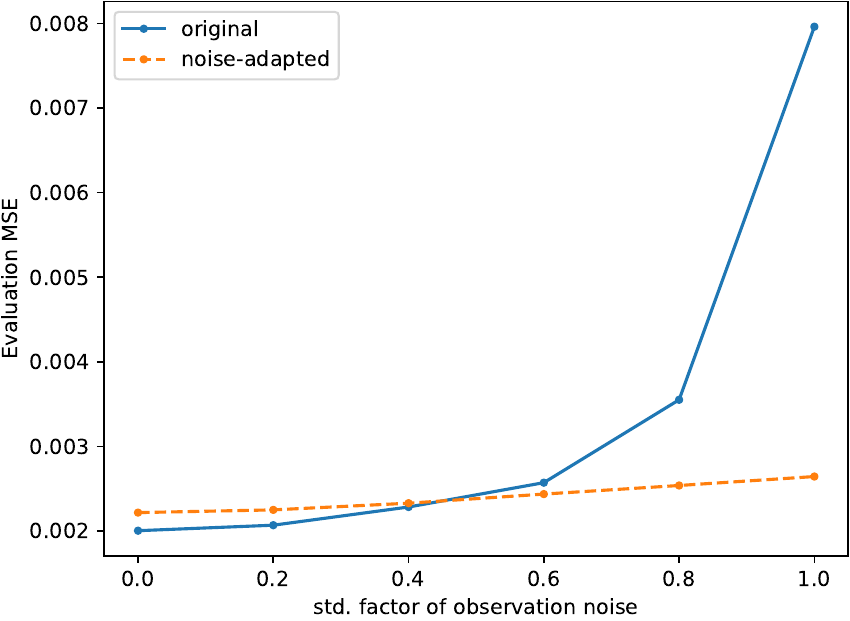}
\caption{Evaluation MSE of the PD-NJ-ODE trained on the Physionet dataset with the original and the noise-adapted loss with different size for the standard deviation factor of the synthetically added observation noise.}
\label{fig:Physionet noisy obs comp}
\end{figure}
To test the new training framework for noisy observations in a 41-dimensional, complex real world setting, we use the Physionet dataset \citep{physionet}. Even though the Physionet dataset, as it is, is likely to have some observation noise, it turns out that this noise is too small to have a big impact on the training (cf.\ Figure~\ref{fig:Physionet noisy obs comp} and Section~\ref{sec:A Practical Note on Using the Noisy Loss Function}). 
Therefore, we add synthetic noise to all observations that are used as input for the PD-NJ-ODE (but not to the observation on which the evaluation MSE is computed) and study its impact on the evaluation MSE depend on the size of its standard deviation. More precisely, we first compute each coordinate's standard deviation $\sigma_{data, j}$ on the training set and then add i.i.d.\ noise samples $\epsilon_{i,j} \sim \normalDistribution(0, \zeta^2 \, \sigma_{data, j}^2)$, for $\zeta \geq 0$, to each observed coordinate $\gX{}_{\tk{i}, j}$. 
In Figure~\ref{fig:Physionet noisy obs comp} we show the minimal evaluation MSE results when training the PD-NJ-ODE  with the original loss~\eqref{equ:appr loss function} and the noise-adapted loss (see \Cref{equ:Psi noisy obs,equ:Phi noisy obs}) on these datasets for increasing standard deviation factor $\zeta$.
Without additional noise ($\zeta=0$), the original loss function leads to slightly better results than the noise-adapted loss function (which is approximately $10\%$ larger), suggesting that in this case the potential intrinsic observation noise has a smaller impact than the inductive bias of the original loss function (cf.\ Section~\ref{sec:A Practical Note on Using the Noisy Loss Function}). The larger the synthetically added noise gets, the larger is its impact. For the noise-adapted loss function the evaluation MSE only grows linearly (with a small slope)  when $\zeta$ increases. In contrast to this, it grows more than linearly for the original loss function. For some $\zeta \in (0.4, 0.5)$ there is the turning point, where the noise-adapted framework becomes more efficient.
Overall, the evaluation MSE  grows less than $20\%$ for the noise-adapted loss (from $\zeta=0$ to $\zeta=1$), while it grows by nearly {$300\%$} for the original loss function. Moreover, the evaluation MSE with $\zeta=1$ is more than {$3$} times larger for the original PD-NJ-ODE than for the noise-adapted one.
This example demonstrates well the importance of the noise-adapted training framework when the observation noise is large.

\section{Conclusion}
In this work we broadened the applicability of the PD-NJ-ODE of \cite{krach2022optimal} by extending the theoretical foundation to allow for the observation framework (i.e., observation times and masks) to depend on previous information and additionally proposed a new loss function that provably leads to optimal predictions even if observations are noisy.
In particular, we showed that any centered i.i.d.\ observation noise satisfying some integrability conditions can be dealt with by switching to the noise-adapted objective function \eqref{equ:Psi noisy obs}. Moreover, we showed that the proof of the main result can be retained when lifting the independence assumption between the process $\gX{}$ and the observation framework, by extensively working with conditional independence. 
Finally, we provided experiments showing empirically that the PD-NJ-ODE works well in those extended settings.

\section*{Acknowledgement}
All authors would like to thank the anonymous reviewers for their detailed feedback and great suggestions that led to significant improvements of the paper.

\if\addackn1
	\section*{Acknowledgement}
	
\fi

\bibliographystyle{tmlr}
\bibliography{references.bib}

\if\inclapp1
	\clearpage
	\appendix
    \section*{Appendix}

\section{Conditional independence}
\label{app:condindep}
Let $(\Omega, \mathcal{F}, \PBlack)$ be a probability space, $\mathcal G, \mathcal H \subseteq \mathcal{F}$ be two sigma-algebras and let $\Xblack{}$ be a random variable.

The assumption that $\Xblack{}$ is independent of $\mathcal{G}$ leads to the natural but incorrect conclusion that $\cexpect{\Xblack{}}{\sigmab(\mathcal G, \mathcal H)} = \cexpect{\Xblack{}}{\mathcal H}$. For this to hold, we actually need a stronger assumption as in the following proposition that is due to \cite{stackexchange_hansen, stackexchange_yoo}.

\begin{prop}\label{prop:conditional independence prop 1}
    If $\sigmab(\Xblack{}, \mathcal H)$ is independent of $\mathcal G$ then $\cexpect{\Xblack{}}{\sigmab(\mathcal G, \mathcal H)} = \cexpect{\Xblack{}}{\mathcal H}$.
\end{prop}

\begin{proof}
We prove the desired equality in the context of basic measure theory. We first assume that $\Xblack{}$ is integrable on $\sigmab(\mathcal G, \mathcal H)$, since otherwise neither of the expectations are valid. Then we recall that conditional expectation is simply a random variable $Z$ that satisfies the following three properties
\begin{enumerate}
    \item $Z$ is $\sigmab(\mathcal G, \mathcal H)$-measurable,
    \item $Z$ is integrable,
    \item $\int_{A} \Xblack{} d \PBlack = \int_{A} Z d \PBlack$ for all $A \in \sigmab(\mathcal G, \mathcal H)$.
\end{enumerate}
By the definition of the conditional expectation we know that $\cexpect{\Xblack{}}{\sigmab(\mathcal G, \mathcal H)}$  satisfies these properties. To show the claim, it is therefore enough to prove that also $Z = \cexpect{\Xblack{}}{\mathcal H}$ satisfies them.
The first two follow trivially since $\mathcal H \subseteq \sigmab(\mathcal G, \mathcal H)$. For the third, we note that by Dynkin's $\pi$-$\lambda$ theorem it is enough to consider the $\cap$-stable generator $P = \{ A \cap B \mid A \in \mathcal G, B \in \mathcal H\}$ of $\sigmab(\mathcal G, \mathcal H)$ and show that
\begin{align*}
    \int_{A \cap B} \Xblack{} d \PBlack = \int_{A \cap B} Z d \PBlack \quad   \forall A \in \mathcal G, B \in \mathcal H,
\end{align*}
or equivalently
\begin{align}\label{equ:cond indep equ prop1}
    \expect{\Xblack{} \cdot \mathbbm{1}_{A \cap B}} = \expect{\cexpect{\Xblack{}}{\mathcal H} \cdot \mathbbm{1}_{A \cap B}} \ \ \  \ \forall A \in \mathcal G, B \in \mathcal H.
\end{align}
Indeed, $P$ is a $\pi$-system (with $\sigmab(P) = \sigmab(\mathcal G, \mathcal H)$) that is trivially included in the smallest Dynkin system $D$ including $P$ (this is the system that includes all sets in $P$, is closed under complements and under countable unions of disjoint sets).
Moreover, if \eqref{equ:cond indep equ prop1} holds for every set in $P$, it is easy to see (by linearity of the integral and dominated convergence) that the corresponding property also holds for every set in $D$, since for a set $A \in D$ we have $\mathbbm{1}_{A^\complement} = 1 - \mathbbm{1}_{A}$ and for disjoint sets $(A_k)_{k\geq 1} \in D$ we have $\mathbbm{1}_{\cup_{k \geq 1} A_k}  = \sum_{k\geq 1} \mathbbm{1}_{A_k}$. Dynkin's $\pi$-$\lambda$ theorem implies that  $\sigmab(P) \subset D$, therefore the third point follows.

To show \eqref{equ:cond indep equ prop1}, note  that $\sigmab(\Xblack{}, \mathcal H)$ being independent of $\mathcal G$ implies that $\mathcal H$ is independent of $\mathcal G$. Therefore,
\begin{align*}
    \expect{\Xblack{} \cdot \mathbbm{1}_{A \cap B}} &= \ \expect{\Xblack{} \cdot \mathbbm{1}_B \mathbbm{1}_A} \\
    &= \ \expect{\Xblack{} \cdot \mathbbm{1}_B} \expect{\mathbbm{1}_A} \tag{$\sigmab(\Xblack{}, \mathcal H)$ indep. of $\mathcal G$} \\ 
    &= \ \expect{\cexpect{\Xblack{} \cdot \mathbbm{1}_{B}}{\mathcal H}} \cdot \expect{\mathbbm{1}_A} \tag{tower property} \\
    &= \ \expect{\cexpect{\Xblack{} \cdot \mathbbm{1}_{B}}{\mathcal H} \cdot \mathbbm{1}_A} \tag{$\mathcal H$ indep. of $\mathcal G$} \\
    &= \ \expect{\cexpect{\Xblack{}}{\mathcal H} \cdot \mathbbm{1}_{A} \mathbbm{1}_{B}} \tag{$\mathbbm{1}_{B}$ is $\mathcal{H}$-measurable} \\
    &= \ \expect{\cexpect{\Xblack{}}{\mathcal H} \cdot \mathbbm{1}_{A \cap B}},
\end{align*}
completing the proof.
\end{proof}

\begin{cor}\label{cor:conditional independence prop 1.2}
    If $\sigmab(\Xblack{}, \mathcal H)$ is independent of $\mathcal G$ then $\cexpect{\phi(\Xblack{})}{\sigmab(\mathcal G, \mathcal H)} = \cexpect{\phi(\Xblack{})}{\mathcal H}$ for all measurable and integrable functions $\phi$.
\end{cor}
\begin{proof}
    This follows directly from Proposition~\ref{prop:conditional independence prop 1} upon replacing $\Xblack{}$ by $\phi(\Xblack{})$ and noting that $\sigmab(\phi(\Xblack{}), \mathcal H) \subseteq \sigmab(\Xblack{}, \mathcal H)$, implying the needed condition.
\end{proof}

It should now be apparent that the assumption that $\Xblack{}$ and $\mathcal G$ are independent is insufficient if we want to show $\cexpect{\Xblack{}}{\sigmab(\mathcal G, \mathcal H)} = \cexpect{\Xblack{}}{\mathcal H}$ (a counterexample is provided in \citet{stackexchange_hansen}).
However, we can actually make a weaker assumption (though still stronger than the assumption that $\Xblack{}$ and $\mathcal G$ are independent) to attain the same result, as shown in the next proposition.

\begin{definition}
    $\Xblack{}$ and $\mathcal G$ are conditionally independent given $\mathcal H$ if for all $x \in \sigmab(\Xblack{})$, $A \in \mathcal G$, 
\begin{align*}
    \cexpect{\ind{x} \ind{A}}{\mathcal H} = \cexpect{\ind{x}}{\mathcal H} \cexpect{\ind{A}}{\mathcal H}.
\end{align*}
\end{definition}

\begin{prop}\label{prop:conditional independence prop 2}
    If $\Xblack{}$ is conditionally independent of $\mathcal G$ given $\mathcal H$ then $\cexpect{\Xblack{}}{\sigmab(\mathcal G, \mathcal H)} = \cexpect{\Xblack{}}{\mathcal H}$.
\end{prop}

\begin{proof}
Following the previous proof it is clear that we only need to show 
$$\expect{\Xblack{} \cdot \mathbbm{1}_{A \cap B}} = \expect{\cexpect{\Xblack{}}{\mathcal H} \cdot \mathbbm{1}_{A \cap B}}$$ 
for all $A \in \mathcal{G}$ and $B \in \mathcal{H}$. We do this by measure-theoretic induction (as in \citet[Proof of Theorem 1.6.9]{Durrett:2010:PTE:1869916}), in particular, we proceed in a four-part case distinction of $\Xblack{}$. In the first case, assume $\Xblack{}$ is an indicator function, i.e., $\Xblack{} = \ind{x}$ for some $x \in \mathcal{F}$. Then 
\begin{align*}
    \expect{\Xblack{} \cdot \ind{A \cap B}} &= \expect{\ind{x} \ind{A \cap B}} \\
    &= \ \expect{\cexpect{\ind{x} \ind{A \cap B}}{\mathcal H}} \tag{tower property} \\
    &= \ \expect{\cexpect{\ind{x} \ind{A}}{\mathcal H} \ind{B} } \tag{$B \in \mathcal H$} \\
    &= \ \expect{\cexpect{\ind{x}}{\mathcal H} \cexpect{\ind{A}}{\mathcal H} \ind{B} } \tag{cond. indep.} \\
    &= \ \expect{ \cexpect{ \cexpect{\ind{x}}{\mathcal H} \cdot \ind{A}}{\mathcal H} \ind{B} } \tag{$\cexpect{\ind{x}}{\mathcal H}$ $\mathcal H$-mbl.} \\
    &= \ \expect{ \cexpect{ \cexpect{\ind{x}}{\mathcal H} \cdot \ind{A \cap B}}{\mathcal H} } \tag{$B \in \mathcal H$} \\
    &= \ \expect{ \cexpect{\ind{x}}{\mathcal H} \cdot \ind{A \cap B}} \tag{tower property} \\
    &= \ \expect{ \cexpect{\Xblack{} }{\mathcal H} \cdot \ind{A \cap B}}.
\end{align*}
Thus in the most basic case we have the required property. In the second case we let $\Xblack{} = \sum_{i = 1}^{k} c_i \ind{x_i}$ be a finite weighted sum of indicator random variables, where $k\in\N$, $c_i \in \R$ and $x_i \in \mathcal{F}$. The result of the first case combined with linearity of expectation immediately shows that the property holds for $\Xblack{}$ in this form too. 

In the third case, we assume $\Xblack{}$ is some non-negative function. We define a random variable $\Xblack{}_k$ that is simple and such that $\Xblack{}_k \uparrow \Xblack{}$ as $k \to \infty$. For example, we can take $\Xblack{}_k = \sum_{i=0}^{k 2^{k} - 1} \frac{i}{2^{k}} \ind{} \{\frac{i}{2^{k}} \le \Xblack{} < \frac{i+1}{2^{k}}\} + k \ind{k \le \Xblack{}}$, as in \citet[Proof of Theorem 1.6.9]{Durrett:2010:PTE:1869916}. Then, by monotone convergence and the previous case, the property also holds when $\Xblack{}$ is an arbitrary non-negative function. 

Finally, in the fourth case, we let $\Xblack{}$ be an arbitrary integrable function. Then we can write $\Xblack{} = \Xblack{}^+ - \Xblack{}^-$ where $f^+(x) := \max\{0, f(x) \}$ and $f^-(x) := \min\{0, f(x)\}$. Integrability of $\Xblack{}$ means that $\Xblack{}^+$ and $\Xblack{}^-$ are themselves integrable. We can use linearity of expectation and the previous cases to conclude that, in this general setting, the property still holds. This concludes the proof by measure-theoretic induction. 
\end{proof}

\begin{prop}\label{prop:conditional independence prop 3}
    $\Xblack{}$ is conditionally independent of $\mathcal G$ given $\mathcal H$ if and only if  $\cexpect{\phi(\Xblack{})}{\sigmab(\mathcal G, \mathcal H)} = \cexpect{\phi(\Xblack{})}{\mathcal H}$ for all measurable and integrable functions $\phi$.
\end{prop}
\begin{proof}
    The ``$\Rightarrow$'' direction is a simple corollary of Proposition~\ref{prop:conditional independence prop 2} that follows with the same proof upon replacing $\Xblack{}$ by $\phi(\Xblack{})$ and noticing that $\phi(\Xblack{})$ is a $\sigmab(\Xblack{})$-measurable random variable.

    The ``$\Leftarrow$'' direction is also easy to see. Let $x \in \sigmab(\Xblack{})$, $A \in \mathcal G$ and define $\phi$ such that $\phi(\Xblack{}) = \ind{x}$.
    Then we have 
    \begin{align*}
        \cexpect{\ind{x} \ind{A}}{\mathcal H} 
        &= \cexpect{\cexpect{\ind{x} \ind{A}}{\sigmab(\mathcal{G}, \mathcal{H})}}{\mathcal{H}} \tag{tower property}\\
        &= \cexpect{\cexpect{\ind{x} }{\sigmab(\mathcal{G}, \mathcal{H})} \ind{A} }{\mathcal{H}} \tag{$A \in \mathcal G \subseteq \sigmab(\mathcal{G}, \mathcal{H})$}\\
        &= \cexpect{\cexpect{\ind{x} }{\mathcal{H}} \ind{A} }{\mathcal{H}} \tag{RHS of claim with $\phi(\Xblack{}) = \ind{x}$}\\
        &= \cexpect{\ind{x} }{\mathcal{H}} \cexpect{ \ind{A} }{\mathcal{H}} \tag{$\cexpect{\ind{x} }{\mathcal{H}}$ $\mathcal{H}$-mbl.}
    \end{align*}
    This completes the proof.
\end{proof}

\begin{prop}\label{prop:conditional independence prop 4}
    If $\Xblack{}$ is conditionally independent of $\mathcal G$ given $\mathcal H$, and $\Yblack$ is independent of $\sigmab(\Xblack{}, \mathcal{G}, \mathcal{H})$ then for any measurable and integrable function $f$ also $f(\Xblack{},\Yblack)$ is conditionally independent of $\mathcal G$ given $\mathcal H$.
\end{prop}
\begin{proof}
    By Proposition~\ref{prop:conditional independence prop 3} we have to show that for any measurable $\phi$ we have $\cexpect{\phi(f(\Xblack{},\Yblack))}{\sigmab(\mathcal G, \mathcal H)} = \cexpect{\phi(f(\Xblack{},\Yblack))}{\mathcal H}$.
    First note that it is enough to show this for $\phi$ being the identity, since both $\phi$ and $f$ are arbitrary measurable  and integrable functions.
    Then we have  for $g(x) := \E_{\Yblack}[{f(\Xblack{},\Yblack)}]$, which is a measurable and integrable function again, that
    \begin{align*}
        \cexpect{f(\Xblack{},\Yblack)}{\sigmab(\mathcal G, \mathcal H)} 
        &= \cexpect{\cexpect{f(\Xblack{},\Yblack)}{\sigmab(\Xblack{}, \mathcal G, \mathcal H)}}{\sigmab(\mathcal G, \mathcal H)} \tag{tower property}\\
        &= \cexpect{g(\Xblack{})}{\sigmab(\mathcal G, \mathcal H)} \tag{\citet[Lemma~6.2.1]{Durrett:2010:PTE:1869916}} \\
        &= \cexpect{g(\Xblack{})}{\mathcal H} \tag{Proposition~\ref{prop:conditional independence prop 3}} \\
        &= \cexpect{f(\Xblack{},\Yblack)}{\mathcal H} \tag{reversing step 1 \& 2}, 
    \end{align*}
    as wanted, proving the claim.
\end{proof}

\begin{prop}\label{prop:conditional independence prop 5}
    If $\Xblack{}$ is conditionally independent of $\Yblack$ given $\mathcal{H}$ then $$\E[\Xblack{}\Yblack \mid \mathcal{H}] = \E[\Xblack{} \mid \mathcal{H}] \, \E[\Yblack \mid \mathcal{H}].$$
\end{prop}
\begin{proof}
    From Proposition~\ref{prop:conditional independence prop 2} we know that $\E[\Xblack{} \mid \sigmab(\Yblack, \mathcal{H})] = \E[\Xblack{} \mid \mathcal{H}] $.
    Therefore, 
    \begin{align*}
        \cexpect{\Xblack{}\Yblack}{\mathcal H} 
        &= \cexpect{\cexpect{\Xblack{}\Yblack}{\sigmab(\Yblack, \mathcal{H})}}{\mathcal{H}} \tag{tower property}\\
        &= \cexpect{\cexpect{\Xblack{} }{\sigmab(\Yblack, \mathcal{H})} \Yblack }{\mathcal{H}} \tag{$\Yblack \in  \sigmab(\Yblack, \mathcal{H})$}\\
        &= \cexpect{\cexpect{\Xblack{} }{\mathcal{H}} \Yblack }{\mathcal{H}} \tag{Proposition~\ref{prop:conditional independence prop 2}}\\
        &= \cexpect{\Xblack{} }{\mathcal{H}} \cexpect{ \Yblack }{\mathcal{H}} \tag{$\cexpect{\Xblack{} }{\mathcal{H}}$ $\mathcal{H}$-mbl.},
    \end{align*}
    as wanted.
\end{proof}

\begin{prop}\label{prop:conditional independence prop 6}
     If $\Xblack{}$ is independent of $\sigmab(\mathcal G, \mathcal H)$ then $\Xblack{}$ is conditionally independent of $\mathcal G$ given $\mathcal H$.
\end{prop}
\begin{proof}
    Follows directly from Proposition~\ref{prop:conditional independence prop 3}.
\end{proof}

\begin{prop}\label{prop:conditional independence prop 7}
    Let $X,Y,Z$ be random variables such that $X$ and $Y$ are conditionally independent given $\sigmab(Z)$.
    Then, for any measurable function $\phi$ we have 
    \begin{equation*}
        \cexpect{\phi(X,Y)}{\sigmab(Z,Y)} = \cexpect{\phi(X,y)}{\sigmab(Z)}\vert_{y=Y}.
    \end{equation*}
\end{prop}

We follow the proof of the somewhat less general result \citet[Lemma~6.2.1]{Durrett:2010:PTE:1869916}.
\begin{proof}
    By its definition, it is clear that the r.h.s.\ is $\sigmab(Y,Z)$-measurable. 
    Hence, by the definition of the conditional expectation, we only need to show that 
    \begin{equation} \label{equ:proof prop A9, 0}
        \expect{\phi(X,Y) \ind{C}} = \expect{\cexpect{\phi(X,y)}{\sigmab(Z)}\vert_{y=Y} \,  \ind{C}},
    \end{equation}
    for all $C \in \sigmab(Y, Z)$.
    
    Assume first that $\phi(x,y) = \ind{A}(x) \ind{B}(y)$ and let $C \in \sigmab(Y,Z)$ be of the form $C = \{ Y \in C_1, Z \in C_2 \}$ for Borel sets $A,B, C_1, C_2$ in the images of $X,Y, Z$, respectively.
    Then,
    \begin{equation}\label{equ:proof prop A9, 1}
        \begin{split}
            \expect{\phi(X,Y) \ind{C}} &=
            \expect{ \cexpect{\ind{A}(X) \ind{B}(Y) \ind{C_1}(Y) \ind{C_2}(Z)}{\sigmab(Z)}} \\
            &= \expect{ \ind{C_2}(Z) \,  \cexpect{\ind{A}(X)}{\sigmab(Z)} \,  \cexpect{\ind{B}(Y) \ind{C_1}(Y) }{\sigmab(Z)} },
        \end{split}
    \end{equation}
    using the tower property in the first and conditional independence in the second equality.
    Moreover, note that 
    $$\cexpect{\phi(X,y)}{\sigmab(Z)} = \cexpect{ \ind{A}(X) \ind{B}(y) }{\sigmab(Z)} = \ind{B}(y) \cexpect{ \ind{A}(X) }{\sigmab(Z)},$$ 
    hence,
    \begin{equation}\label{equ:proof prop A9, 2}
        \begin{split}
            \expect{\cexpect{\phi(X,y)}{\sigmab(Z)}\vert_{y=Y} \,  \ind{C}} 
            &= \expect{ \ind{B}(Y) \, \cexpect{ \ind{A}(X) }{\sigmab(Z)} \,  \ind{C} } \\
            &= \expect{ \cexpect{ \ind{B}(Y) \,  \cexpect{ \ind{A}(X) }{\sigmab(Z)} \, \ind{C_1}(Y) \, \ind{C_2}(Z)  }{\sigmab(Z)} } \\
            &= \expect{ \cexpect{ \ind{A}(X) }{\sigmab(Z)} \, \ind{C_2}(Z) \, \cexpect{ \ind{B}(Y)  \,  \ind{C_1}(Y)  }{\sigmab(Z)} },
        \end{split}
    \end{equation}
    using the tower property in the second and measurability in the third equality.
    \eqref{equ:proof prop A9, 1} and \eqref{equ:proof prop A9, 2} together show that \eqref{equ:proof prop A9, 0} holds for the simple functions and the special $C$.
    Next, we note that by Dynkin's $\pi$-$\lambda$ theorem, the same holds for general $C \in \sigmab(Y,Z)$ (cf.\ proof of Proposition~\ref{prop:conditional independence prop 1}). 
    Then the claim follows for all bounded functions by the monotone class theorem \citep[Theorem~6.1.3]{Durrett:2010:PTE:1869916}, as outlined in the proof of \citet[Lemma~6.2.1]{Durrett:2010:PTE:1869916}. Moreover, it follows for general measurable functions by dominated convergence. 
\end{proof}

\section{Inductive Bias}\label{sec:inductive bias}
In this section we first discuss the inductive bias induced by the old loss function. Afterwards we  discuss the general inductive bias introduced by the PD-NJ-ODE~\eqref{equ:PD-NJ-ODE}.
In this paper, we have shown that under certain assumptions our estimator based on the new loss~\eqref{equ:loss noisy obs generalised} is consistent, and thus asymptotically unbiased as the model complexity $m$ and the number of training paths $N$ tend to infinity.
However in such an infinite hypothesis space it is impossible to obtain an unbiased estimator for any finite number of training paths $N\in \N$. For this reason, it is helpful to have an appropriate inductive bias that guides the estimator in the right direction when there is a limited amount of training data \citep{Mitchell:1980:InductiveBias}.\footnote{Occam's razor suggests, that one should always pick the simplest (or most parsimonious) model that explains the data, i.e., the inductive bias should be directed towards simplicity/parsimony. Different machine learning (ML) models incorporate simplicity/parsimony in various different ways, so one should pick the right model according to the alignment of the inductive bias of the model with one's prior belief.} 

\subsection{Inductive Bias of the Loss}
While the new loss~\eqref{equ:loss noisy obs generalised} is asymptotically unbiased, the old loss~\eqref{equ:Psi} is in general always biased. Even in the limit $m,N\to\infty$ this bias induced by the old loss~\eqref{equ:Psi} does not vanish. On average, models obtained from the old loss will jump too closely to new observation~$\gO{\tk{k}}$ at observation times $\tk{k}$ as can be observed in the right subplot of \Cref{fig:NoisyObs}.
However, in the case of no noise (i.e., $\gX{\tk{k}}=\gO{\tk{k}}$), the correct model should always exactly jump to new observations. Hence, teaching the model explicitly to jump to new observations via the term~$\left\lvert \gM{k} \odot ( \gX{\tk{k}} - Z_{\tk{k}} ) \right\rvert_2$ in the old loss~\eqref{equ:Psi} incorporates helpful prior knowledge which is particularly helpful for small training datasets. Even if the observations~$\gO{\tk{k}}$ are noisy (with a small noise-scale $\sqrt{\E[\epsilon_k^2]}$), the term~$\left\lvert \gM{k} \odot ( \gO{\tk{k}} - Z_{\tk{k}} ) \right\rvert_2$ in the old loss~\eqref{equ:Psi} can be a beneficial inductive bias, if one only has access to a very small training dataset. If one has very little information from other observations about the dynamics of $X$, then it might be a reasonable best guess to jump close to new observations~$\gO{\tk{k}}$ at observation times~$\tk{k}$. For example, if the model $\gY{\tk{k}}^{\thetamNMin{m,N}}(\tildeXle{\tk{k-1}})$ obtained from the new loss~\eqref{equ:loss noisy obs generalised} is further away from the true conditional expectation~$\left.\cexpect{\gX{t}}{\At{\tk{k-1}}}\right|_{t=\tk{k}}$ than the noisy observation~$\gO{\tk{k}}$, the term~$\left\lvert \gM{k} \odot ( \gO{\tk{k}} - Z_{\tk{k}} ) \right\rvert_2$ in the old loss~\eqref{equ:Psi} would be helpful to push the model closer to the truth. This explains why we can see in \Cref{fig:Physionet noisy obs comp} that the old loss~\eqref{equ:Psi} outperforms the new loss~\eqref{equ:loss noisy obs generalised} in the case of very small noise relative to the overall variability of the data.\footnote{In this experiment, we trained also only for $100$ epochs. Additionally to the statistical learning argument given before, there is probably also an algorithmic effect, that the term~$\left\lvert \gM{k} \odot ( \gO{\tk{k}} - Z_{\tk{k}} ) \right\rvert_2$ in the old loss~\eqref{equ:Psi} helps the model to move directly in the right direction starting from the first epochs. At least during the first epochs the model $\gY{\tk{k}}^{\theta}(\tildeXle{\tk{k-1}})$ is usually very far off (at observation times~$\tk{k}$) compared to the noisy~$\gO{\tk{k}}$ observations thus the term~$\left\lvert \gM{k} \odot ( \gO{\tk{k}} - Z_{\tk{k}} ) \right\rvert_2$ in the old loss~\eqref{equ:Psi} pushes the model directly in the right direction at observation times, while the new loss~\eqref{equ:loss noisy obs generalised} only has an indirect affect on the model at the observation times~$\tk{k}$.}

Furthermore, different weighting schemes in the loss also influence the inductive bias (see \Cref{rem:different weigthings}). The term~$\left\lvert \gM{}_k \odot ( \gO{\tk{k}} - Z_{\tk{k}} ) \right\rvert_2$ in the old loss~\eqref{equ:Psi} can be particularly helpful if the loss is weighted via $\Delta\tk{k}$.

\subsection{Inductive Bias of the PD-NJ-ODE architecture}
As shown in our main theorem the PD-NJ-ODE architecture~\eqref{equ:PD-NJ-ODE} is universal and when it is trained with the right loss, such as \eqref{equ:loss noisy obs generalised}, it is asymptotically unbiased. However, for any finite number of training paths $N$, there are infinitely many models (i.e., infinitely $\theta\in\Theta$) that all perfectly fit the data, i.e., achieve $\hPhi{N}(\theta)=0$, but correspond to very different predictions $\gY{t}^{\theta}(\tildeXle{s})$ on unseen test data. Therefore, for $N<\infty$, our architecture (including the training algorithm) necessarily needs to have an appropriate inductive bias to make a reasonable choice among all the models with reasonable low training loss~$\hPhi{N}(\theta)$.\footnote{If the underlying process $\gX{}$ is not deterministic or if the observations are noisy, even the exact conditional expectation does not fit the observations $\gO{\tk{k}}$ exactly (i.e., $\Psi(\hX{})>0$) due to stochasticity. Hence, the model has to learn to average out the noise. Therefore, we do not only need to choose among models that achieve $\hPhi{N}(\theta)=0$, but even among a wider range of models achieving $\hPhi{N}(\theta)<c$.} (Additionally we want this bias to vanish in the limit $N\to\infty$, which is already given by our theory.)
First, we give some intuition on our understanding of the important inductive bias of PD-NJ-ODE~\eqref{equ:PD-NJ-ODE} and afterwards we will give some theoretical arguments that motivate these hypotheses.
Most machine learning models have and inductive bias towards some form of simplicity for the function mapping inputs to predictions. E.g., some NN-architectures have an inductive bias towards low second derivative of this function w.r.t.\ the inputs \citep{savarese2019infinite,ongie2019function,heiss2019ImplReg1,heiss2023ImplReg2,HeissImplReg3,parhi2022kinds} (i.e., the function can still be universal but among all functions with equal loss the one which is \enquote{most linear} is chosen).
The crucial difference of PD-NJ-ODE~\eqref{equ:PD-NJ-ODE} is that its inductive bias favours simplicity of the functions $f_{\theta_{1}}, \rho_{\theta_{2}}, \tilde{g}_{\tildetheta{3}}$ instead of simply favouring simplicity of the map $(\tildeXle{s},t) \mapsto \gY{t}^{\theta}(\tildeXle{s})$.
For many applications of PD-NJ-ODEs, a linear ODE is intuitively one of the most simple/parsimonious/conservative/plausible models possible. The following example illustrates the benefit of the inductive bias of PD-NJ-ODEs.
\begin{example}
    Consider the situation where the true underlying process of $\gX{}$ follows a linear SDE (driven by a Brownian motion, i.e., a linear It\^o-diffusion), then the conditional expectation $\hX{}$ follows a linear ODE (except the jumps at observation times $\tk{k}$, cf.\ \citet[Example~B.2]{krach2022optimal}). For simplicity consider the case of complete noiseless observations $\gX{\tk{k}}$ (i.e., $M=1$ and $\epsilon=0$). Then $\hX{}$ can be perfectly expressed by our PD-NJ-ODE~\eqref{equ:PD-NJ-ODE} with 3 \emph{linear} functions $f_{\theta_{1}}, \rho_{\theta_{2}}, \tilde{g}_{\tildetheta{3}}$ instead of simply favouring simplicity of the map $(\tildeXle{s},t) \mapsto \gY{t}^{\theta}(\tildeXle{s})$. E.g., $f_{\theta_{1}}$ is the right-hand side of the ODE resulting in $\hX{}$, $\rho_{\theta_{2}}$ maps new observations $\gX{\tk{k}}$ simply to $\gX{\tk{k}}$,\footnote{Note that the the first $\gls{dX}$ signatures of level 1 of $\tildeXle{\tk{k}}-\gX{0}$ are exactly $\gX{\tk{k}}-\gX{0}$. Thus, the function~$\rho_{\theta_2}$ defined by $\rho_{\theta_2}\left( H_{t-}, t, \pim{m} (\tildeXle{\tau(t)}-\gX{}_0 ), \gX{}_0 \right)=\gX{\tk{k}}$ is linear.} and $\tilde{g}_{\tildetheta{3}}$ is also simply the identity. As these 3 functions are linear, their second derivative is zero. Hence, for example, the inductive bias of L2-regularized ReLU-NNs~$f_{\theta_{1}}, \rho_{\theta_{2}}, \tilde{g}_{\tildetheta{3}}$ strongly favours such simple linear functions. Therefore, the model can learn these linear dynamics already from a relatively small amount of training paths and extrapolate well far into the future. On the other hand, in this setting the map $(\tildeXle{s},t) \mapsto \hX{t}(\tildeXle{s})$ grows \emph{exponentially} in the input $t$ and thus has huge (second) derivative with respect to $t$, which is considered the opposite of simple by most of the commonly used ML-models.   
    Hence, if one trains, on the same dataset, a feed-forward neural network $\check{F}_{\theta}:(\tildeXle{s},t) \mapsto \check{F}_{\theta}(\tildeXle{s},t)$ that directly approximates $\Fj{}$, 
    the model would find many other \enquote{simpler}/flatter functions~$\check{F}_{\theta}$ that fit the finite training dataset equally well.
    Even for larger $N$, such a model, favouring a low Sobolev-norm of the map $\check{F}_{\theta}$, would always have troubles to make predictions far ahead into the future $t\gg\tau(t)$: For values of $t-\tau(t)$ larger than the largest $\tk{k}-\tk{k-1}$ ever observed in training, such a model has no reason to keep growing exponentially but would rather extrapolate as flat as possible. In contrast, $\gY{t}^{\theta}(\tildeXle{\tau(t)})$, obtained from our PD-NJ-ODE with linear functions~$f_{\theta_{1}}, \rho_{\theta_{2}}, \tilde{g}_{\tildetheta{3}}$, would keep the exponential growth in $t$ for arbitrarily large values of $t-\tau(t)$ even far out of sample and even far out of distribution.\footnote{For our theoretical result, we assumed $f_{\theta_1}$ to be bounded, which would not enable this exponentially growth arbitrarily far into the future, but at least for a very long time if $\gamma_1$ is large enough.}
    In line with this, it is expected that the PD-NJ-ODE leads to better results than the model~$\check{F}_{\theta}$, when evaluated on previously unseen test data. This was confirmed by the experiment in \citet[Appendix~E]{krach2022optimal}. 
\end{example}

To summarize: While the PD-NJ-ODE~\eqref{equ:PD-NJ-ODE} is able to learn arbitrarily complicated dynamics (fulfilling some mild regularity assumptions), it favours models~$\theta$ with \enquote{simple} functions $f_{\theta_{1}}, \rho_{\theta_{2}}, \tilde{g}_{\tildetheta{3}}$ among different models with equal training loss~$\hPhi{N}(\theta)$, where the definition of \enquote{simplicity} depends on the architectures used for $f_{\theta_{1}}, \rho_{\theta_{2}}, \tilde{g}_{\tildetheta{3}}$. E.g., for L2-regularized ReLU-NNs the notion of simplicity relates to the second derivative of $f_{\theta_{1}}, \rho_{\theta_{2}}, \tilde{g}_{\tildetheta{3}}$ \citep{savarese2019infinite,ongie2019function,heiss2019ImplReg1,heiss2023ImplReg2,HeissImplReg3,parhi2022kinds}. This can lead to better generalization performance.

\subsubsection{Inductive Bias towards Multi-task Learning}
If one trains multiple tasks together (i.e., if one wants to forecast a multi-dimensional process $\gX{}$), many deep learning applications have shown that an inductive bias towards functions that use similar features among multiple tasks (rather than using very different features for each of these tasks) can be highly beneficial. This phenomena is known as \emph{multi-task learning} \citep{MultitaskCaruana1997}. One can argue that multi-task learning (highly related to feature learning, representation learning, transfer learning, metric learning) is one of the biggest strengths of deep learning models compared to shallow learning models in terms of generalization \citep{HeissImplReg3}. The inductive bias of our PD-NJ-ODE~\eqref{equ:PD-NJ-ODE} towards feature sharing (i.e., multi-task learning) is twofold:
\begin{enumerate}
    \item Per definition the hidden state $H_t$ in \eqref{equ:PD-NJ-ODE} is shared among all tasks/outputs. I.e., each coordinate of the output $\gY{t}$ is a function of the same hidden state $H_t$. Therefore, the hidden state $H_t$ becomes more parsimonious, if it mainly contains features that are helpful for all coordinates of $\gY{t}$.
    \item If the neural network architectures of $f_{\theta_{1}}, \rho_{\theta_{2}}, \tilde{g}_{\tildetheta{3}}$ contain trainable parameters in multiple hidden layers, then these functions favour feature sharing within their hidden layers, as shown by \citet{HeissImplReg3} in the case of L2-regularization\footnote{In the case of L2-regularization \citet{HeissImplReg3} shows (for certain architectures) that this multi-task learning effect does not vanish when the width of the network increases, while it would vanish with increasing width for some specific other training methods. For most deep learning methods used in practice, there is some multi-task learning effect assumed to be present even if it is theoretically not always as well understood as for L2-regularized wide ReLU-NNs.}. \Citet{MultitaskCaruana1997,HeissImplReg3} provide more intuition on multi-task learning.
\end{enumerate}

\subsubsection{Implicit Regularization}
Even without any explicit L2-regularization and even for very large theoretical model complexity $m$ (i.e., large $\Theta_m$ in our setting), it is widely assumed that (stochastic) gradient descent (initialized close to zero) induces an implicit inductive bias towards simplicity \citep{neyshabur2017implicit}. In some settings this implicit regularization is extremely similar (or even exactly identical) to explicit L2-regularization \citep{heiss2019ImplReg1} and in other settings it can be (slightly) different. In our setting of training PD-NJ-ODEs, implicit regularization of the gradient descent does most probably not exactly correspond to explicit L2-regularization, but on a high level the qualitative behaviour might be similar.\footnote{Note that a classical gradient flow is always identical to a gradient flow with respect to the L2 norm and thus \emph{locally} moves as little w.r.t.\ the L2-distance as possible. Therefore, it seems plausible that gradient descent does not increase the L2-norm of $\theta$ \emph{much} more than necessary to achieve a certain training loss.}

In theory, $m$ could also be too large in relation to $N$, since too large $m$ allows for over-fitting. In contrast to this, $N$ can never be too large. In particular, if one picks an $\thetamNMin{m,N}\in\ThetamNMin{m,N}$ with unnecessarily large L2-norm over-fitting can be a serious problem. However, in practice, too many neurons are usually not a problem due to implicit regularization of (stochastic) gradient descent as suggested empirically by \citet[Figure~6]{herrera2021neural} and theoretically by \citet{neyshabur2017implicit,savarese2019infinite,ongie2019function,heiss2019ImplReg1,heiss2023ImplReg2,HeissImplReg3,parhi2022kinds}. 

\subsubsection{Mathematical Theory on the Inductive Bias of PD-NJ-ODEs}
For implicit regularization it seems difficult to get precise global results for PD-NJ-ODEs. However, for explicit L2-regularization of $\theta$, there is much more hope to get nice theoretical results on the inductive bias of the PD-NJ-ODEs~\eqref{equ:PD-NJ-ODE}. (Note that the bound $m$ on the L2-norm of $\theta$ in the definition of $\Thetam{m}$ corresponds to L2-regularization in a Lagrangian sense.\footnote{The L2-regularization regularization parameter~$\lambda_{\text{reg}}$ can be seen as a KKT-multiplier of the constraint $\vert\theta\rvert_2<m$. However, if $m$ is too large the constraint is not active and $\thetamNMin{m,N}$ is highly undetermined. For L2-regularization with $\lambda_{\text{reg}}>0$, the \emph{parameters}~$\thetamNMin{m,N}$ are usually also not uniquely defined, but the solution $\gY{}^{\thetamNMin{m,N}}$ can be unique.}) The results by \citet{savarese2019infinite,ongie2019function,heiss2019ImplReg1,heiss2023ImplReg2,HeissImplReg3,parhi2022kinds} strongly support the hypothesis that $\argmin_{\theta} \left(\hPhi{N}(\theta)+\lambda_{\text{reg}} \left\lvert\theta\right\rvert_2^2\right)$ exactly converges to $\argmin_{f,\rho,\tilde{g}} \left(\hat{\Psi}_{\!N}(\Yblack^{f,\rho,\tilde{g}}) + \lambda_{\text{reg}} \left(\vphantom{\big(}P_1(f)+P_2(\rho)+P_3(\tilde{g})\right)\right)$, where $P_i$ describe some regularization functionals (such as weighted $L^p$ norms of the second derivative) depending on the exact architectures used for $f,\rho,\tilde{g}$ given in \citet{savarese2019infinite,ongie2019function,heiss2019ImplReg1,heiss2023ImplReg2,HeissImplReg3,parhi2022kinds}. However, (for rather technical reasons) their proofs do not directly apply to PD-NJ-ODEs. The formulation of the the main theorem in \cite{heiss2023ImplReg2} could be applied more directly to PD-NJ-ODE by considering the map $(f,\rho,\tilde{g}) \mapsto \hat{\Psi}_{\!N}(\Yblack^{f,\rho,\tilde{g}})$ as the loss functional.\footnote{One would need to check if this loss functional meets the assumptions of \cite{heiss2023ImplReg2} for a rigorous proof.} Proofing such results rigorously would be interesting future work.

\section{Combining the Two Extensions with Full Proof}
\label{sec:Combining the Two Extensions With Full Proof}
In this section we prove convergence of the PD-NJ-ODE to the optimal prediction in the most general setting that allows for non-Markovian processes with irregular incomplete noisy observations, where dependence between the observation framework and the process is possible.

\subsection{Setting}\label{sec:Setting both ext}
As in Section~\ref{sec:Setting with Dependence} we consider only the probability space $\gls{OmFFP}$ on which we define  $\gX{}, \gn{}, \gls{K}, \tk{i}, \tau, \gM{}$.  As in Section~\ref{sec:Setting with Noisy Observations}, we additionally define the observation noise $(\epsilon_k)_{0 \leq k \leq \gls{K}}$, the noisy observations $\gO{\tk{k}} := \gX{}_{\tk{k}} + \epsilon_k$ for $0 \leq k \leq \gn{}$, the filtration of the currently available information via
\begin{equation*}
\At{t} := \boldsymbol{\sigma}\left(\gO{\tk{i}, j}, \tk{i}, \gM{\tk{i}} | \tk{i} \leq t,\, j \in \{1 \leq l \leq \gls{dX} | \gM{\tk{i}, l} = 1  \} \right),
\end{equation*} 
$\tildeOle{t}$ and the functions $\Fj{j}$ such that $\hX{}_{t,j} = \Fj{j}(t, \tau(t), \tildeOle{\tau(t)})$.
Then we make the following assumptions.
\begin{assumption} \label{ass:noise&dependence}
We assume that: 
\begin{enumerate}[label=(\roman*)] 
    \item $\gM{0, j} = 1$ for all $1 \le j \le \gls{dX}$ ($\gX{}$ is completely observed at $0$) and $\abs{\gM{}_k}_1 > 0$ for every $1 \le k \le \gls{K}$ $\glsP{OmFFP}$-almost surely (at every observation time at least one coordinate is observed). \label{ass_both_1}
    
    \item The probability that any two observation times are closer than $\epsilon > 0$ converges to 0 when $\epsilon$ does, i.e., if $\delta(\omega) := \min_{0 \le i \le \gn{}(\omega)} \abs{\tk{i+1}(\omega) - \tk{i}(\omega)}$ then $\lim_{\epsilon \to 0} \glsP{OmFFP}(\delta < \epsilon) = 0$.\label{ass_both_2}
    
    \item Almost surely $\gX{}$ is not observed at a jump, i.e., $\glsP{OmFFP}(\tk{i} \in \gls{J} \mid i \le \gn{}) = \glsP{OmFFP}(\Delta \gX{}_{\tk{i}} \neq 0 \mid i \le \gn{}) = 0$ for all $1 \le i \le \gls{K}$. \label{ass_both_3}
    
    \item $\Fj{j}$ are  continuous and differentiable in their first coordinate $t$ such that their partial derivatives with respect to $t$, denoted by $f_j$, are again continuous and there exists a $B >0$ and $p \in \N$ such that for every $t \in [0,\gls{T}]$ the functions $f_j, \Fj{j}$ are polynomially bounded in $\gls{Xs}$, i.e., 
    $$|\Fj{j}(\tau(t), \tau(t), \tildeOle{\tau(t)})| +  | f_j(t, \tau(t), \tildeOle{\tau(t)})  | \leq B (\glst{Xs} +1)^p + B \sum_{i=0}^{\gn{}} |\epsilon_i|. $$\label{ass_both_4}
    
    \item $\gls{Xs}$ is $L^{2p}$-integrable, i.e., $\expect{(\glsT{Xs})^{2p}} < \infty$.\label{ass_both_5}

    \item $\gn{}$ is square-integrable, i.e., $\E[\gn{2}]<\infty$. \label{ass_both_9}
    
    \item The i.i.d.\ $\epsilon_k$ are independent of $\gX{}, \gn{}, \gM{}, (\tk{i})_{1 \leq i \leq \gls{K}}$, are centered and square-integrable, i.e., $\E[\epsilon_k] = 0$ and $\E[|\epsilon_k|^{2}] < \infty$. \label{ass_both_6}
    
    \item For every  $1 \le i \le \gn{}$, $\gX{}_{\tk{i}-}$ is conditionally independent of $\sigmab(\gn{}, \gM{\tk{i}})$ given $\At{\tk{i}-}$.\footnote{More precisely, one should formulate this assumption as \enquote{For every  $1 \le i \le \gls{K}$, $\gX{}_{\tk{i}-}\1_{\{i\leq \gn{}\}}$ is conditionally independent of $\sigmab(\gn{}, \gM{\tk{i}})$ given $\At{\tk{i}-}$.}} \label{ass_both_7}

    \item For all $1 \le k \le \gls{K}$, $1 \le j \le \gls{dX}$ there is some $\eta_{k,j} >0$ such that $\glsP{OmFFP} ( \gM{k, j} = 1 \mid \sigmab(\gn{}, \At{\tk{k}-} )) > \eta_{k,j}$ (i.e., given the currently known information and $\gn{}$, for each coordinate the probability of observing it at the next observation time is positive). \label{ass_both_8}

    \item We assume that for every $1 \leq k \leq \gls{K}$ the process $\gX{}$ is conditionally independent of $\tk{k}$ given $\At{\tk{k-1}}$. \label{ass_both_10}
\end{enumerate}
\end{assumption}
\begin{rem}
All the relaxations and extensions discussed in \Cref{sec:Recall: the PD-NJ-ODE,sec:PD-NJ-ODE with Noisy Observations,sec:PD-NJ-ODE with Dependence betweenXand Observation Framework,sec:Practical Implications of the Convergence Result} hold also in this combined setting.
\end{rem}
As in Section~\ref{sec:Setting with Noisy Observations}, the PD-NJ-ODE uses the noisy observations $\gO{\tk{i}}$ and $ \tildeOle{\tau(t)}$ as inputs instead of $\gX{}_{\tk{i}}$ and $\tildeXle{\tau(t)}$ and is trained with the objective functions \eqref{equ:Psi noisy obs} respectively \eqref{equ:Phi noisy obs} and their Monte Carlo approximations.

\subsection{Convergence Theorem}\label{sec:Convergence Theorem both ext}
\begin{theorem}\label{thm:depndence&noise}
    If Assumption~\ref{ass:noise&dependence} is satisfied and using the definitions of \Cref{sec:Setting both ext}, the claims of \Cref{thm:1} hold equivalently, upon replacing the original loss functions and their Monte Carlo approximations by their noise-adapted versions\footnote{In particular, we replace the original empirical loss~$\hPhi{N}(\theta) = \text{\eqref{equ:appr loss function}}$ by the noise-adapted empirical loss function,
    \begin{equation}\label{equ:appr noise-adapted loss function}
\hPhi{N}(\theta) := \frac{1}{N} \sum_{j=1}^N  \frac{1}{\gn{(j)}}\sum_{i=1}^{\gn{(j)}} \left\lvert \gM{i}^{(j)} \odot \left( \gO{}_{\tksup{i}{(j)}}^{(j)} - \gY{}_{\tksup{i}{(j)}-}^{\theta, j } \right) \right\rvert_2^2.
\end{equation}
    }. In particular, we obtain convergence  of our estimator~$\gY{}^{\thetamMin{m,N_m}}$ to the true conditional expectation $\hX{}$ in $\dk{k}$.\footnote{For a convergence result at times $t$ that are not observation times, see \Cref{sec:From Approximations at Observation Times to Approximations at any Time}.}
\end{theorem}

We start with the following result which is a combination of Lemma~\ref{lem:L2 identity noisy obs setting} and Lemma~\ref{lem:L2 identity dependence}.

\begin{lem}\label{lem:L2 identity both ext}
For any $\bA{}$-adapted process $Z$ it holds that
\begin{equation*}
    \E \left[\tfrac{1}{\gn{}} \sum_{i=1}^{\gn{}} \left\lvert \gM{\tk{i}} \odot ( \gO{\tk{i}} - Z_{\tk{i}-} ) \right\rvert_2^2\right]
	= \E \left[ \tfrac{1}{\gn{}}\sum_{i=1}^{\gn{}} \left\lvert \gM{\tk{i}} \odot ( \gO{\tk{i}} - \hX{}_{\tk{i}-} ) \right\rvert_2^2\right] + \E \left[\tfrac{1}{\gn{}}\sum_{i=1}^{\gn{}} \left\lvert \gM{\tk{i}} \odot (  \hX{}_{\tk{i}-} - Z_{\tk{i}-}) \right\rvert_2^2\right] .
\end{equation*}
\end{lem}

\begin{proof}
    First note that by Assumption~\ref{ass:noise&dependence} point \ref{ass_both_3} we have that $\gX{}_{\tk{i}} = \gX{}_{\tk{i}-}$ almost surely and when defining $\gO{\tk{i} -} := \gX{}_{\tk{i} -} + \epsilon_i$ we therefore also have that $\gO{\tk{i} } = \gO{\tk{i} -}$ almost surely.
    Next notice that Assumption~\ref{ass:noise&dependence} point \ref{ass_both_6} \& \ref{ass_both_7} together with Proposition~\ref{prop:conditional independence prop 4} imply that $\gO{\tk{i}-}$ is conditionally independent of $\sigmab(\gn{}, \gM{}_i)$ given $\At{\tk{i}-}$.
    Hence, for $\hO{\tk{i}-} := \E[\gO{\tk{i} -} \, | \, \At{\tk{i}-}]$ it follows as in Lemma~\ref{lem:L2 identity dependence} that
    \begin{equation*}
        \E \left[\tfrac{1}{\gn{}}\sum_{i=1}^{\gn{}} \left\lvert \gM{\tk{i}} \odot ( \gO{\tk{i}-} - Z_{\tk{i}-} ) \right\rvert_2^2\right] 
	    = \E \left[\tfrac{1}{\gn{}}\sum_{i=1}^{\gn{}} \left\lvert \gM{\tk{i}} \odot ( \gO{\tk{i}-} - \hO{\tk{i}-} ) \right\rvert_2^2\right] + \E \left[\tfrac{1}{\gn{}}\sum_{i=1}^{\gn{}} \left\lvert \gM{\tk{i}} \odot (  \hO{\tk{i}-} - Z_{\tk{i}-} ) \right\rvert_2^2\right].
    \end{equation*}
    Then we can conclude the proof as in Lemma~\ref{lem:L2 identity noisy obs setting}, by noting that with \Cref{prop:rewriting of hat X} we obtain,
    \begin{equation*}
    \hO{\tk{i}-} = \hX{}_{\tk{i}-} + \E[\epsilon_i | \At{\tk{i}-}] = \hX{}_{\tk{i}-} + \E[\epsilon_i] = \hX{}_{\tk{i}-},
    \end{equation*}
    using that $\epsilon_i$ has expectation $0$ and is independent of $\At{\tk{i}-}$.
\end{proof}

In the following, we use the notation  $\Thetam{m}^i$ and  $\tildeThetam{m}^i$  if we speak of the projections of the sets on the weights $\theta_i$ and $\tildetheta{i}$ respectively.

\begin{proof}[Proof of Theorem~\ref{thm:depndence&noise} -- Part 1]
    We start by showing that $\hX{} \in \bD{}$ is the unique minimizer of $\Psi$ up to indistinguishability (as defined in Definition~\ref{def:indistinguishability}). Note that for every $\tk{i}$ we have $\gM{\tk{i}} \odot \hX{}_{\tk{i}} = \gM{\tk{i}} \odot \gX{}_{\tk{i}}$ and that $\gX{}_{\tk{i}} = \gX{}_{\tk{i}-}$ if $\tk{i} \notin \gls{J}$, hence with probability 1. It follows directly from Lemma~\ref{lem:L2 identity both ext} that $\hX{}$ is a minimizer of $\Psi$, since
    \begin{equation}\label{equ:Psi of Z split}
        \Psi(Z) = \Psi(\hX{})  + \E \left[\tfrac{1}{\gn{}}\sum_{i=1}^{\gn{}} \left\lvert \gM{\tk{i}} \odot (  \hX{}_{\tk{i}-} - Z_{\tk{i}-}) \right\rvert_2^2\right].
    \end{equation} 
    Before we can show uniqueness of $\hX{}$, we need some additional results. For those, let $Z \in \bD{}$.
    Let $c_1 := \E\left[ \gn{} \right]^{1/2} \in (0, \infty)$, then  the H\"older inequality, together with the fact that $\gn{} \geq 1$, yields 
    \begin{equation}\label{equ:HI}
        \E\left[ \left\lvert Z \right\rvert_2 \right] 
	   = \E\left[ \frac{\sqrt{\gn{}}}{\sqrt{\gn{}}} \left\lvert Z \right\rvert_2 \right] 
	   \leq c_1 \, \E\left[ \frac{1}{\gn{}} \left\lvert Z \right\rvert_2^2 \right]^{1/2}.
    \end{equation}
    
    By Lemma~\ref{lem:ck} and by the equivalence of $1$- and $2$-norm we have for some constant $c_3 > 0$ and for any $1 \leq k \leq \gls{K}$ 
    \begin{equation}\label{equ:M split both}
        \E\left[ \1_{\{\gn{} \geq k\}} \left\lvert \hX{}_{\tk{k}-} - Z_{\tk{k}-}  \right\rvert_2 \right] 
        \leq \frac{c_3}{c_2(k)} \E\left[ \1_{\{\gn{} \geq k\}} \left\lvert \gM{\tk{k}} \odot ( \hX{}_{\tk{k}-} - Z_{\tk{k}-} ) \right\rvert_2 \right].
    \end{equation}

    To see that $\hX{}$ is the unique minimiser up to indistinguishability, let $Z \in \bD{}$ be a process which is not indistinguishable from $\hX{}$. Hence, there exists some $1 \leq k \leq \gls{K}$ such that $\dk{k}(\hX{}, Z) > 0$. We have
    \begin{equation}\label{equ:thm1-positive expectation}
    \begin{split}
         \E\left[\tfrac{1}{\gn{}}\sum_{i=1}^{\gn{}} \left\lvert \gM{\tk{i}} \odot (  \hX{}_{\tk{i}-} - Z_{\tk{i}-}) \right\rvert_2^2\right]
	   &=  \E\left[\tfrac{1}{\gn{}}\sum_{i=1}^{\gls{K}} \1_{\{ \gn{} \geq i \}} \left\lvert \gM{\tk{i}} \odot (  \hX{}_{\tk{i}-} - Z_{\tk{i}-}) \right\rvert_2^2\right] \\
	   & \geq \E\left[\tfrac{1}{\gn{}}  \1_{\{ \gn{} \geq k \}} \left\lvert \gM{\tk{k}} \odot (  \hX{}_{\tk{k}-} - Z_{\tk{k}-}) \right\rvert_2^2\right] \\
	   & \geq c_1^{-2} \E\left[ \1_{\{ \gn{} \geq k \}} \left\lvert \gM{\tk{k}} \odot (  \hX{}_{\tk{k}-} - Z_{\tk{k}-}) \right\rvert_2\right]^2 \\
	   & \geq \left( \frac{c_2}{c_1 c_3} \right)^2 \E\left[ \1_{\{ \gn{} \geq k \}} \left\lvert  \hX{}_{\tk{k}-} - Z_{\tk{k}-}  \right\rvert_2\right]^2 \\
	   & = \left( \frac{c_2}{c_0 c_1 c_3} \right)^2 \dk{k}(\hX{}, Z)^2 > 0,
    \end{split}
    \end{equation} 
    where we used \eqref{equ:HI} for the 3rd, \eqref{equ:M split both} for the 4th and \eqref{equ:pseudo metric dk} for the last line. Together with \eqref{equ:Psi of Z split} this implies $\Psi(Z) > \Psi(\hX{})$.

    Next we show that \eqref{equ:PD-NJ-ODE} can approximate $\hX{}$ arbitrarily well. Since the dimension $d_H$ can be chosen freely, let us fix it to $d_H := \gls{dX}$. 
    Furthermore, let us fix  $\tildetheta{}_3^{\star}$ such that $ \tilde g_{\tildetheta{3}^{\star}} = \id$, which is possible since we assumed that $\id \in \tcN{}$. 
    Let $\varepsilon > 0$,   $N_\varepsilon := \lceil 2 (\gls{T}+1) \varepsilon^{-4} \rceil  $ (implying that $\lim_{\varepsilon \to 0} N_\varepsilon = \infty$) and $\mathcal{P}_\varepsilon\subseteq \text{BV}^c([0,\gls{T}])$ be the closure of the set $A_{ N_\varepsilon  }$ of \citet[Remark~3.11]{krach2022optimal}, which is compact.
    For any $1 \leq j \leq \gls{dX}$, the function $f_j$ is continuous by Assumption~\ref{ass:noise&dependence} and can (by abuse of notation) equivalently be written as a (continuous) function $ f_j(t, \tau(t), \tildeOle{\tau(t)} -\gO{}_0, \gO{}_0 )$.
    Therefore, \citet[Proposition~3.8]{krach2022optimal} implies that there exists an $m_0 = m_0(\varepsilon) \in \N$ and a continuous function $\hat f_j$ such that
    \begin{equation*}
        \sup_{(t, \tau, \Xblack{}) \in [0,\gls{T}]^2\times \mathcal{P}_\varepsilon } \left| f_j(t, \tau, \Xblack{} ) - \hat f_j(t, \tau, \pim{m_0}( \Xblack{} -\Xblack{}_0 ), \Xblack{}_0 )\right| \leq \varepsilon/2.
    \end{equation*}
    Since the variation of functions in $\mathcal{P}_\varepsilon$ is uniformly bounded by a  finite constant,  the set of their truncated signatures $\pim{m_0}(\mathcal{P}_\varepsilon)$ is a bounded subset in $\R^d$ for some $d \in \N$ (depending on $\gls{dX}$ and $m_0$), hence its closure, denoted by $\Pi_\varepsilon$, is compact. 
    Therefore, the universal approximation theorem for neural networks  \citep[Theorem 2.4]{10.5555/70405.70408} implies that there exists an $m_1 = m_1(\varepsilon) \in \N$ and neural network weights $\tildetheta{}_1^{\star, m_1} \in \tildeThetam{m_1}^1$ such that for every $1 \leq j \leq \gls{dX}$ the function $\hat f_j$ is approximated up to $\varepsilon/2$ by the $j$-th coordinate of the neural network $\tilde f_{\tildetheta{}_1^{\star, m_1}} \in \tcN{}$ (denoted by $\tilde f_{\tildetheta{}_1^{\star, m_1}, j}$) on the compact set $[0,\gls{T}]^2\times \Pi_\varepsilon$. Hence, combining the two approximations we get (by triangle inequality)
    \begin{equation*}
        \sup_{(t, \tau, \Xblack{}) \in [0,\gls{T}]^2\times \mathcal{P}_\varepsilon } \left| f_j(t, \tau, \Xblack{} ) -  \tilde f_{\tildetheta{}_1^{\star, m_1}, j}  (t, \tau, \pim{m_0}( \Xblack{} -\Xblack{}_0 ), \Xblack{}_0 )\right| \leq \varepsilon.
    \end{equation*}
    Obviously, extending the input of the neural network does not make the approximation worse, by simply setting the corresponding weights to $0$, hence, also $H_{t-}$ can be used as additional input. 
    Similarly we get that  there exists an $m_2 = m_2(\varepsilon) \in \N$ and neural network weights $\tildetheta{}_2^{\star, m_2} \in \tildeThetam{m_2}^2$ such that for every $1 \leq j \leq \gls{dX}$
    \begin{equation*}
        \sup_{(t, \Xblack{}) \in [0,\gls{T}] \times \mathcal{P}_\varepsilon } \left| \Fj{j}(t, t, \Xblack{} ) -  \tilde \rho_{\tildetheta{}_2^{\star, m_2}, j}  (t, \pim{m_1}( \Xblack{} -\Xblack{}_0 ), \Xblack{}_0 )\right| \leq \varepsilon.
    \end{equation*}
    As before, $H_{t-}$ can be used as additional input without worsening the approximation.
    
    Next we define the bounded output neural networks based on these neural networks. For this let us define 
    \begin{equation*}
    \gamma_1 := \max_{(t, \tau, \Xblack{}) \in [0,\gls{T}]^2\times \mathcal{P}_\varepsilon } \left|  \tilde f_{\tildetheta{}_1^{\star, m_1}}  (t, \tau, \pim{m_0}( \Xblack{} -\Xblack{}_0 ), \Xblack{}_0 )\right|
    \end{equation*}
    and $\gamma_2$ equivalently for $\tilde \rho_{\tildetheta{}_2^{\star, m_2}}$. Since the neural networks are continuous functions they take a finite maximum on the compact sets, hence $\gamma_1, \gamma_2$  are finite.
    Then we define the bounded output neural networks  $f_{\theta_1^{\star, m_1}}, \rho_{ \theta_2^{\star, m_2}} \in \cN{}$ with $\theta_i^{\star, m_i} := (\tildetheta{}_i^{\star, m_i}, \gamma_i)$.
    Clearly, these bounded output neural networks coincide with the neural networks on the compact sets. Therefore, they satisfy the same $\varepsilon$-approximation and since $\Fj{j}, f_j$ are bounded by $U:=B\left((\glsT{Xs} +1)^p+\sum_{i=0}^{\gn{}} |\epsilon_i|\right)$ (Assumption~\ref{ass:noise&dependence}\ref{ass_both_4}), it follows that $f_{\theta_1^{\star, m_1}}, \rho_{ \theta_2^{\star, m_2}}$ are bounded by $U + \varepsilon$.
    In particular, we have for $\epsilon<B$ the global bounds $|f_j - f_{\theta_1^{\star, m_1}, j} |_\infty \leq 3U$ and $|\Fj{j} - \rho_{\theta_2^{\star, m_2}, j}|_\infty \leq 3U$.
    Setting $m := \max(m_0, m_1, m_2, \gamma_1, \gamma_2, |\tildetheta{}_i^{\star, m_2}|_2, |\tildetheta{}_2^{\star, m_2}|_2)$, it follows that $\theta^\star_m := (\theta^{\star, m_1}_1, \theta^{\star, m_2}_2, \tildetheta{}^\star_3 ) \in \Thetam{m}$.

    Now we can bound the distance between $\gY{}_t^{\theta_m^\star}(\tildeOle{t})$ and $\hX{t}$. Whenever $\glsT{Xs} < 1/\varepsilon$,  the number of observations satisfies $\gn{} < 1/\varepsilon$, the minimal difference between any two consecutive observation times $\delta > \varepsilon$ and all noise terms satisfy $| \epsilon_i |_2 < 1/\varepsilon$ we know that the corresponding path $\tildeOle{\tau(t)} - \gO{}_0$ is an element of $A_{N_\varepsilon}$ and therefore the neural network approximations up to $\varepsilon$ hold. Otherwise, one of those conditions is not satisfied and the global upper bound can be used. 
    Hence, if $t \in \{ \tk{1}, \dotsc, \tk{\gn{}} \}$, we have for $\Fj{} = (\Fj{j})_{1 \leq j \leq \gls{dX}}$ and $f = (f_j)_{1 \leq j \leq \gls{dX}}$
    \begin{equation*}
    \begin{split}
        \left\lvert \gY{}_t^{\theta_m^*} - \hX{}_t  \right\rvert_1 
	   & = \left\lvert  \rho_{\theta_2^{\star, m_2}}\left(H_{t-}, t,\pim{m}( \tildeOle{t} - \gO{}_0 ), \gO{}_0 \right) - \Fj{} \left(t, t, \tildeOle{t} \right) \right\rvert_1  \\
	   & \leq \varepsilon \gls{dX} \1_{\{ \glsT{Xs} < 1/\varepsilon \}}  \1_{\{ \gn{} < 1/\varepsilon \}}  \1_{\{ \delta > \epsilon \}} \1_{\{ \forall 0 \leq i \leq \gn{}: | \epsilon_i |_2 < 1/\varepsilon \} } \\
	   & \quad +  \gls{dX} 3 U \left( \1_{\{ \glsT{Xs} \geq 1/\varepsilon \}} + \1_{\{ \gn{} \geq  1/\varepsilon \}} + \1_{\{ \delta \leq \epsilon \}} + \1_{\{ \exists 0 \leq i \leq \gn{}: | \epsilon_i |_2 \geq 1/\varepsilon \}}  \right),
    \end{split}
    \end{equation*}
    and if $t \notin \{ \tk{1}, \dotsc, \tk{\gn{}} \}$,
    \begin{equation*}
    \begin{split}
        \left\lvert \gY{}_t^{\theta_m^*} - \hX{}_t  \right\rvert_1  
	   &\leq \left\lvert \gY{}_{\tau(t)}^{\theta_m^*} - \hX{}_{\tau(t)}  \right\rvert_1 
    +  \int_{\tau(t)}^t \left\lvert f_{\theta_1^{\star, m_1}}\left(H_{s-}, s, \tau(t), \pim{m} (\tildeOle{\tau(t)} -\gO{}_0 ), \gO{}_0 \right) 
		- f(s, \tau(t), \tildeOle{\tau(t)})  \right\rvert_1 ds\\
	   & \leq \varepsilon (\gls{T}+1) \gls{dX} \1_{\{ \glsT{Xs} < 1/\varepsilon \}}  \1_{\{ \gn{} < 1/\varepsilon \}}  \1_{\{ \delta > \varepsilon \}} \1_{\{ \forall 0 \leq i \leq \gn{}: | \epsilon_i |_2 < 1/\varepsilon \} }  \\
	   & \quad + (\gls{T}+1) \gls{dX} 3U \left( \1_{\{ \glsT{Xs} \geq 1/\varepsilon \}} + \1_{\{ \gn{} \geq  1/\varepsilon \}} + \1_{\{ \delta \leq \varepsilon \}} + \1_{\{ \exists 0 \leq i \leq \gn{}: | \epsilon_i |_2 \geq 1/\varepsilon \}} \right).
    \end{split}
    \end{equation*}
    Moreover, by  equivalence of the $1$- and $2$-norm, there exists a constant $c > 0$ such that for all $t \in [0,\gls{T}]$
    \begin{equation*}
    \begin{split}
        \left\lvert \gY{}_t^{\theta_m^*} - \hX{}_t  \right\rvert_2 
	   & \leq c \, \varepsilon (\gls{T}+1) \gls{dX} 
     +  c (\gls{T}+1) \gls{dX} 3 U \left( \1_{\{ \glsT{Xs} \geq 1/\varepsilon \}} + \1_{\{ \gn{} \geq  1/\varepsilon \}} + \1_{\{ \delta \leq \varepsilon \}} + \1_{\{ \exists 0 \leq i \leq \gn{}: | \epsilon_i |_2 \geq 1/\varepsilon \}} \right) =: c_m.
    \end{split}
    \end{equation*}

    So far, we have fixed an $\varepsilon >0$ and argued that there exists some $ m \in \N$ such that the neural network approximation bounds hold with $\varepsilon$-error. 
    However, what we actually need to show is that this error converges to $0$ when increasing the  truncation level and network size $m$.
    Therefore, we define $\varepsilon_m \geq 0$ to be the smallest number such that the above bounds hold with error $\varepsilon_m$ when using an architecture with signature truncation level  $m \in \N$ and weights in $\Thetam{m}$. Since increasing $m$ can only make the approximations better (by setting the new weights to $0$, the same approximation error as before is achieved, but potentially there exists a better choice), we have $\varepsilon_{m_1} \geq \varepsilon_{m_2}$ for any $m_1 \leq m_2$. In particular $(\varepsilon_m)_{m \geq 0}$ is a a decreasing sequence, hence, our derivations before proof that $\lim_{m \to \infty} \varepsilon_m = 0 $. In the following we denote by $\theta^\star_m \in \Thetam{m}$ the best choice for the weights within the set $\Thetam{m}$ to approximate the functions $\Fj{j}, f_j$.

   Since $\thetamMin{m} \in \argmin _{\theta \in \Thetam{m}}\{ \Phi(\theta) \}$  (note that at least one minimum exists in the compact set $\Thetam{m}$ since $\Phi$ is continuous), we get with Lemma~\ref{lem:L2 identity both ext} that
    \begin{equation}\label{equ:phi convergence 1 both}
    \begin{split}
        \min_{Z \in \bD{}} \Psi(Z) &\le \Phi (\thetamMin{m}) \le \Phi(\theta^\ast_m)
        =  \expect{\frac{1}{\gn{}}\sum_{i = 1}^{\gn{}} \abs{\gM{\tk{i}} \odot (\gO{\tk{i}} - \gY{}^{\theta^\ast_m}_{\tk{i}} ) }_2^2 } \\
        &= \Psi(\hX{}) + \E \l[ \frac{1}{\gn{}}\sum_{i = 1}^{\gn{}} \abs{\gM{\tk{i}} \odot ( \hX{}_{\tk{i}-} - \gY{}^{\theta^\ast_m}_{\tk{i}-} ) }_2^2 \r]
        \le \Psi(\hX{}) + \E \l[ c_m^2 \r].
    \end{split}
    \end{equation}
    Integrability of $|\glsT{Xs}|_2$, $|\gn{}|$ and $|\epsilon_i|$ together with Assumption~\ref{ass:noise&dependence}\ref{ass_both_2} on $\delta$ imply that 
    $$\1_{\{ \glsT{Xs} \geq 1/\varepsilon_m \}} + \1_{\{ \gn{} \geq  1/\varepsilon_m \}} + \1_{\{ \delta \leq \varepsilon_m \}} + \1_{\{ \exists 0 \leq i \leq \gn{}: | \epsilon_i |_2 \geq 1/\varepsilon_m \}} \xrightarrow[m \to \infty]{\glsP{OmFFP}-a.s.} 0.$$ 
    Therefore, we have for a suitable constant $c>0$ (not depending on $\varepsilon_m$ and $m$),
    \begin{equation*}
    \begin{split}
        \E\left[ c_m^2 \right] 
	   \leq c \varepsilon_m^2+ &  c  \E\left[U^2 \left(  \1_{\{ \glsT{Xs} \geq 1/\varepsilon_m \}} + \1_{\{ \gn{} \geq  1/\varepsilon_m \}} 
        + \1_{\{ \delta \leq \varepsilon_m \}} + \1_{\{ \exists 0 \leq i \leq \gn{}: | \epsilon_i |_2 \geq 1/\varepsilon_m \}} \right) \right] \xrightarrow{m \to \infty} 0,
    \end{split}
    \end{equation*}
    by dominated convergence, since $U$ is $L^{2}$-integrable. Indeed, 
    \begin{equation}\label{equ:integraility in both ext}
        \E[U^2] 
        \leq 8B^2 \E \left[ (\glst{Xs} +1)^{2p} + \gn{} \sum_{i=0}^{\gn{}} |\epsilon_i|^2 \right]
        = 8B^2 \l( \E \left[ (\glst{Xs} +1)^{2p} \right] +  \E[\gn{2}] \E \left[|\epsilon_0|^2 \right] \r) < \infty,
    \end{equation}
    using Cauchy-Schwarz inequality for the first step, that the $\epsilon_i$ are i.i.d. and independent of $\gn{}$ for the equality and the integrability of $\gls{Xs}$, $\epsilon_0$ and $\gn{2}$ for the upper bound.    
    Using this and $\Psi(\hX{}) = \min_{Z \in \bD{}} \Psi(Z)$, we get from \eqref{equ:phi convergence 1 both}
    \begin{equation}\label{equ:convergence theoretical loss function proof}
        \min_{Z \in \bD{}} \Psi(Z) 
	   \leq \Phi(\thetamMin{m}) \leq \Phi(\theta_m^{*}) \xrightarrow{m \to \infty} \min_{Z \in \bD{}} \Psi(Z).
    \end{equation}
    Finally, we show that $\lim_{m\to\infty} \dk{k} \left( \hX{},  \gY{}^{\thetamMin{m}} \right) = 0$ for all $1 \leq k \leq \gls{K}$.
    Applying \eqref{equ:HI}, \eqref{equ:M split both} and \eqref{equ:pseudo metric dk} in  reverse order than it was done in  \eqref{equ:thm1-positive expectation} and finally Lemma~\ref{lem:L2 identity both ext}, yields
    \begin{equation}\label{equ:convergence in L1}
    \begin{split}
        \dk{k} \left( \hX{} , \gY{}^{\theta_m^{*}}  \right)
	   & \leq \frac{c_0 \, c_1 \, c_3}{c_2} \, \E\left[ \frac{1}{\gn{}} \sum_{i=1}^{\gn{}} \left\lvert \gM{\tk{i}} \odot ( \hX{}_{\tk{i}-} - \gY{}^{\theta_m^{*}}_{\tk{i}-} ) \right\rvert_2^2 \right]^{1/2}  =   \frac{c_0 \, c_1 \, c_3}{c_2} \, \left( \Phi(\theta_m^{*}) - \Psi(\hX{}) \right)^{1/2}  \xrightarrow{m \to \infty} 0,
    \end{split}
    \end{equation}
    which completes the first part of the proof.
\end{proof}

Next we assume the size $m$ of the neural network and of the signature truncation level is fixed and we study the convergence of the Monte Carlo approximation when the number of training paths $N$ increases.
The convergence analysis is based on \citet[Chapter 4.3]{lapeyre2019neural} and follows \citet[Theorem E.13]{herrera2021neural}.
We define the separable Banach space $\mathcal{S} := \{ x = (x_i)_{ i \in \N} \in \ell^1(\R^{d}) \; \vert \; \lVert x \rVert_{\ell^1} < \infty \}$ for a suitable $d$ (see below) with the norm $\lVert x \rVert_{\ell^1} := \sum_{i \in \N} \lvert x_i \rvert_2$, the function
\begin{equation*}
G(x,y,m) :=  \left\lvert m \odot ( x - y ) \right\rvert_2
\end{equation*}
and $\xi_j := (\xi_{j, 0}, \dotsc, \xi_{j, \gn{(j)}}, 0, \dotsc)$, where $\xi_{j,k} := (\tksup{k}{(j)}, \gO{\tksup{k}{(j)}}^{(j)}, \gM{\tksup{k}{(j)}}^{(j)}, \pim{m}(\tildeOle{\tksup{k}{(j)}, {(j)}})) \in \R^d$ and $\tksup{k}{(j)}$, $\gM{\tksup{k}{(j)}}^{(j)}$ and $\gO{}^{(j)}_{t^{(j)}_i}$ (with $0$ entries for coordinates which are not observed) are random variables describing the $j$-th realization of the training data (cf. Section~\ref{sec:Recall: the PD-NJ-ODE}).
Let $\njNotNumberOfObservations{j}(\xi_j) := \max_{k \in \N}\{ \xi_{j,k} \neq 0\}$, $\tk{k}(\xi_j):= \tksup{k}{(j)}$, $\gO{}_k (\xi_j) := \gO{\tksup{k}{(j)}}^{(j)}$ and $\gM{}_k(\xi_j) := \gM{\tksup{k}{(j)}}^{(j)}$.
By this definition we have  $\gn{(j)} = \njNotNumberOfObservations{j}(\xi_j)$ $\glsP{OmFFP}$-almost-surely. Moreover, we have that $\xi_j$ are i.i.d. random variables taking values in $\mathcal{S}$.
Let us write $\gY{}_t^{\theta}(\xi)$ to make the dependence of $\gY{}$ on the input and the weight $\theta$ explicit.
Then we define
\begin{equation*}
h(\theta, \xi_j) := \frac{1}{\njNotNumberOfObservations{j}(\xi_j)}\sum_{i=1}^{\njNotNumberOfObservations{j}(\xi_j)}  G \left( \gO{}_i(\xi_j), \gY{}^{\theta}_{\tk{i}(\xi_j)-}(\xi_j), \gM{}_i(\xi_j) \right)^2.
\end{equation*}

The following lemma is known from \cite{krach2022optimal}.
\begin{lemma}\label{lem:properties for MC conv thm}
Almost-surely the random function $\theta\in \Thetam{m} \mapsto \gY{}_{t}^{\theta}$ is uniformly continuous for every $t \in [0,\gls{T}]$.
\end{lemma}
Now we are ready to prove the second part of our main theorem.

\begin{proof}[Proof of Theorem \ref{thm:depndence&noise} -- Part 2.]
First we note that, $\gY{}^{\theta}_t$ is the (integration over the) output of (bounded output) neural networks and therefore bounded in terms of the input, the weights (which are bounded by $m$), $\gls{T}$ and some constant depending on the architecture and the activation functions of the neural network. 
In particular we have that $|\gY{}^{\theta}_t(\xi_j)| \leq \tilde B \leq \tilde B \left((\gXs{\gls{T}}{(j)} +1)^p+\sum_{i=0}^{\njNotNumberOfObservations{j}(\xi_j)} |\epsilon_i^{(j)}|\right)$ for all $t \in [0,\gls{T}]$ and $\theta \in \Thetam{m}$ for some constant $\tilde B$ (possibly depending on $m$), where $\gXs{}{(j)}, \epsilon_i^{(j)}$ corresponds to the input $\xi_j$.
Hence,
\begin{multline*}
    G \left( \gO{}_i(\xi_j), \gY{}^{\theta}_{\tk{i}(\xi_j)-}(\xi_j), \gM{}_i(\xi_j) \right)^2
	=  \left\lvert \gM{}_i(\xi_j) \odot ( \gO{i}(\xi_j) - \gY{}^{\theta}_{\tk{i}(\xi_j)-}(\xi_j) ) \right\rvert_2^2\\
	\leq \left((B+\tilde B)\left((\glsT{Xs} +1)^p+\sum_{i=0}^{\njNotNumberOfObservations{j}(\xi_j)} |\epsilon_i|\right)\right)^{2} 
    = \left( \frac{B+\tilde B}{B} U^{(j)}\right)^2,
\end{multline*}
where $U^{(j)}$ is as defined before corresponding to the input $\xi_j$.
Hence,
\begin{equation}
\label{equ:dominating bound loss function}
\E \left[\sup_{\theta \in \Thetam{m}} h(\theta, \xi_j)\right] 
\leq   \E \left[\frac{1}{\njNotNumberOfObservations{j}(\xi_j)}\sum_{i=1}^{\njNotNumberOfObservations{j}(\xi_j)} \left( \frac{B+\tilde B}{B} U^{(j)}\right)^2 \right]  < \infty,
\end{equation}
by Assumption~\ref{assumption:F} and \eqref{equ:integraility in both ext}.
By Lemma~\ref{lem:properties for MC conv thm}, the function $\theta \mapsto h(\theta, \xi_1)$ is continuous, hence, we can apply \citet[Lemma~4.6]{krach2022optimal}, yielding that almost-surely for $N \to \infty$ the function 
\begin{equation}\label{equ:unif conv 1}
\theta \mapsto \frac{1}{N} \sum_{j=1}^{N} h(\theta, \xi_j) = \hPhi{N}(\theta)
\end{equation}
converges uniformly on $\Thetam{m}$ to 
\begin{equation}\label{equ:unif conv 2}
\theta \mapsto \E [h(\theta, \xi_1)] = \Phi(\theta).
\end{equation} 
Moreover, we deduce from \citet[Lemma~4.6]{krach2022optimal} that $d(\thetamNMin{m,N},\ThetamMin{m})\to 0$ a.s. when $N\to \infty$. 
Then there exists a sequence $(\hatthetamNMin{m,N})_{N \in \N}$ in $\ThetamMin{m}$ such that $\lvert \thetamNMin{m,N} - \hatthetamNMin{m,N} \rvert_2 \to 0$ a.s. for $N \to \infty$.
The uniform continuity of the random functions $\theta \mapsto \gY{}_{t}^{\theta}$ on $\Thetam{m}$ implies that for any fixed deterministic and bounded $\xi_0$ (taking values in the same space as the $\xi_j$),
$$\lvert \gY{}_{t}^{\thetamNMin{m,N}}(\xi_0) - \gY{}_{t}^{\hatthetamNMin{m,N}}(\xi_0) \rvert_2 \to 0 \text{ a.s. for all  } t \in [0,\gls{T}] \text{ as } N \to \infty.$$ 
By continuity of $G$ this yields $\lvert h(\thetamNMin{m,N}, \xi_0) - h(\hatthetamNMin{m,N}, \xi_0) \rvert \to 0$ a.s.\ as $N \to \infty$.
Let $\xi_0$ now be a random variable which is independent of and identically distributed as the $\xi_j$ defined on a copy $(\Omega_0, \mathbb{F}_0, \mathcal{F}_0, \P_0)$ of the filtered probability space $(\Omega, \mathbb{F}, \mathcal{F}, \P)$. Then the above statements hold for $\xi_0(\omega_0)$ for $\P_0$-a.e.\ fixed $\omega_0$.
Hence, we have for $\P$-a.e.\ fixed $\omega$, that $\lvert h(\thetamNMin{m,N}\omb, \xi_0) - h(\hatthetamNMin{m,N}\omb, \xi_0) \rvert \to 0$ $\P_0$-a.s.\ as $N \to \infty$.
With \eqref{equ:dominating bound loss function} we can apply dominated convergence which yields
\begin{equation*}
\lim_{N \to \infty} \E_{\xi_0} \left[ \lvert h(\thetamNMin{m,N}\omb, \xi_0) - h(\hatthetamNMin{m,N}\omb, \xi_0) \rvert \right] = 0 \text{ for $\P$-a.e. } \omega.
\end{equation*}
Since for every integrable random variable $Z$ we have $0 \leq \lvert \E[Z] \rvert \leq \E[\lvert Z \rvert] $ and since $\hatthetamNMin{m,N}\in \ThetamMin{m}$ we can deduce that for $\P$-a.e.\ fixed $\omega$,
\begin{equation}
\label{equ: MC convergence}
\lim_{N \to \infty} \Phi(\thetamNMin{m,N}\omb) = \lim_{N \to \infty} \E_{\xi_0} \left[  h(\thetamNMin{m,N}\omb, \xi_0) \right] = \lim_{N \to \infty} \E_{\xi_0} \left[  h(\hatthetamNMin{m,N}\omb, \xi_0) \right] = \Phi(\thetamMin{m}).
\end{equation}
Now by triangle inequality, for $\P$-a.e.\ fixed $\omega$, we have for $\hPhi{\tilde{N}}$ and $\Phi$ defined through  test samples $\tilde{\xi}_j$ on $\Omega_0$, i.e., independent of and identically distributed as the training samples $\xi_j$ yielding  $\thetamNMin{m,N}\omb$,
\begin{equation}\label{equ: MC convergence 2}
\lvert \hPhi{\tilde{N}}(\thetamNMin{m, N}\omb) -  \Phi(\thetamMin{m}) \rvert \leq \lvert \hPhi{\tilde{N}}(\thetamNMin{m, N}\omb) -  \Phi(\thetamNMin{m, N}\omb) \rvert + \lvert \Phi(\thetamNMin{m, N}\omb) -  \Phi(\thetamMin{m}) \rvert.
\end{equation}
\eqref{equ:unif conv 1} and \eqref{equ:unif conv 2} imply that the first term on the right hand side converges to 0 when $\tilde{N} \to \infty$
and \eqref{equ: MC convergence} implies that the second term on the right hand side converges to 0 a.s.\ when $ N \to \infty$. 
Moreover, the uniform convergence in \eqref{equ:unif conv 1} and \eqref{equ:unif conv 2} yields the same result when setting $\tilde N = N$. Furthermore, \citet[Lemma~4.6]{krach2022optimal} yields the same result for $\hPhi{N}(\thetamNMin{m, N})\omb$, i.e., when $\hPhi{N}$ and $\Phi$ are defined through the $\xi_j$ on the probability space corresponding to the training data. This finishes the proof of the convergence with respect to $N$.

Finally, we want to show the joint convergence.
We define $N_0 := 0$ and for every $m \in \N$
\begin{equation*}
N_m\omb := \min\left\{ N \in \N \; \vert \; N > N_{m-1}\omb, \lvert \Phi(\thetamNMin{m,N}\omb) - \Phi(\thetamMin{m}) \rvert  \leq \tfrac{1}{m} \right\},
\end{equation*}
which is possible due to \eqref{equ: MC convergence} for $\P$-a.e. $\omega$. Then \eqref{equ:convergence theoretical loss function proof} implies that for $\P$-a.e. $\omega$
\begin{equation*}
\lvert \Phi(\theta^{\min}_{m,N_m\omb}\omb) - \Psi(\hX{}) \rvert  \leq \tfrac{1}{m} +  \lvert \Phi(\thetamMin{m}) - \Psi(\hX{}) \rvert \xrightarrow{m \to \infty} 0.
\end{equation*}
Therefore, we can apply the same arguments as in the first part of the proof (cf. \eqref{equ:convergence in L1}) to show that 
\begin{equation*}
\dk{k} \left(  \hX{} , \gY{}^{\theta^{\min}_{m,N_m\omb}\omb}  \right)
\leq   \frac{c_0 \, c_1 \, c_3}{c_2} \, \left( \Phi(\theta^{\min}_{m,N_m\omb}\omb) - \Psi(\hX{}) \right)^{1/2}  \xrightarrow{m \to \infty} 0,
\end{equation*}
for every $1 \leq k \leq \gls{K}$ and for $\P$-a.e. $\omega$.
\end{proof}

The following corollary follows as in \cite{krach2022optimal}.
\begin{cor}\label{cor:1}
In the setting of Theorem~\ref{thm:depndence&noise}, we also have that $\glsP{OmFFP}$-a.s.
\begin{equation*}
\Phi(\theta^{\min}_{m,N_m}) \xrightarrow{m \to \infty} \Psi(\hX{}) \quad \text{and} \quad \hPhi{\tilde{N}_m}(\theta^{\min}_{m,\tilde{N}_m}) \xrightarrow{m \to \infty} \Psi(\hX{}),
\end{equation*}
for a suitable increasing random sequence $(\tilde{N}_m)_{m \in \N}$ in $\N$.
\end{cor}

\section{Experimental Details}\label{sec:Experimental Details}

Our experiments are based on the implementation used by \cite{krach2022optimal}, which is available at \url{https://github.com/FlorianKrach/PD-NJODE}. Therefore, we refer the reader to its Appendix for any details that are not provided here.

\subsection{Differences between the Implementation and the Theoretical Description of the PD-NJ-ODE}\label{sec:Differences between the Implementation and the Theoretical Description of the PD-NJ-ODE}
Since we use the same implementation of the PD-NJ-ODE, all differences between the implementation and the theoretical description listed in \citet[Appendix~D.1.1]{krach2022optimal} also apply here. In particular, we use standard neural networks for $f_{\theta_1}$ and $\rho_{\theta_2}$ with additional inputs. 
In contrast to the original objective function, we do not need to add a regularizing constant for the noise-adapted objective functions \eqref{equ:Psi noisy obs} and \eqref{equ:Phi noisy obs}, since no square-root needs to be computed.
We do not use the $\tu{}$-coordinates of $\tildeXle{t}$ (i.e., we only use the first $\gls{dX}$ coordinates and omit the remaining coordinates from $\gls{dX}+1$ to $2\gls{dX}$) in the implementation, since we anyways use the observation times and masks as additional inputs for $\rho_{\theta_2}$.

\subsection{Details for Noisy Observations}\label{sec:Details for Noisy Observations}
\textbf{Dataset.} 
We sample paths from a standard $1$-dimensional Brownian motion on the interval $[0,1]$, i.e., with $\gls{T}=1$, and discretisation time grid with step size $0.01$. At each time point we observe the process with probability $p=0.1$. Whenever the process is observed, an independent observation noise is sampled from a centered normal distribution with standard deviation $\sigma=0.5$ and added to the observation. The model never sees the observation of the original process, but only the noisy observation. We sample $20'000$ paths of which $80\%$ are used as training set and the remaining $20\%$ as test set.

\textbf{Architecture.}
We use the PD-NJ-ODE with the following architecture. The latent dimension is $d_H = 100$, the readout network is a linear map  and the other 2 neural networks have the same structure of 1 hidden layer with $\operatorname{ReLU}$ activation function and $100$ nodes. The signature is used up to truncation level $3$, the encoder is recurrent and the decoder uses a residual connection.

\textbf{Training.}
We use the Adam optimizer with the standard choices $\beta = (0.9, 0.999)$, weight decay of $0.0005$ and learning rate $0.001$. Moreover, a dropout rate of $0.1$ is used for every layer and training is performed with a mini-batch size of $200$ for $200$ epochs.
The model is trained once with the noise-adapted objective function \eqref{equ:Psi noisy obs} and once with the original one \eqref{equ:Psi}.

\subsection{Details for Dependent Observation Framework}\label{sec:Details for Dependent Observation Framework}
\textbf{Dataset.}
We use the Euler scheme to sample paths from a $1$-dimensional Black--Scholes model (geometric Brownian motion) with drift $\mu = 2$, volatility $\sigma = 0.3$, and starting value $\gX{}_0 = 1$.
At each time point we observe the process with the probability $\glsP{OmFFP}(\gM{}_i = 1 \, | \, \At{\tk{i}-}) = \E[\gM{}_i \, | \, \At{\tk{i}-} ]$, where $\gM{}_i$ is described in Example~\ref{exa:Black--Scholes with Dependent Observations}, using $\eta=3$ and $p=0.1$.
We use the same discretisation grid and dataset sizes as in Section~\ref{sec:Details for Noisy Observations}.

\textbf{Architecture.}
We use the PD-NJ-ODE with the following architecture. The latent dimension is $d_H = 50$ and all 3 neural networks have the same structure of 2 hidden layers with $\tanh$ activation function and $50$ nodes. 
The signature is used up to truncation level $3$, the encoder is recurrent and the decoder does not use a residual connection.

\textbf{Training.}
Same as in Section~\ref{sec:Details for Noisy Observations} but only trained with original loss function \eqref{equ:Psi}.

\subsection{Details for Physionet with Observation Noise}
\label{sec:Details for Physionet with Observation Noise}

\textbf{Dataset.}
Details on the standard Physionet dataset are given in \citet[Appendix F.5.3]{herrera2021neural}.
In particular, we use a train-test split of $80\%$ to $20\%$ (i.e., $N=6\,400$ training paths, and $N_{\text{test}}=1\,600$ test paths). On the train set, the entire path amounting to 48 hours is used as input to the model, while on the test set only the first 24 hours are used as input and the second 24 hours are used to compute the evaluate MSE. In particular, the model predicts starting from 24 hours until 48 hours and at every time point where there is an observation, the squared error is computed between the observation and the prediction. Importantly, the observation is not used as an input to the model afterwards. I.e., we evaluate $\gY{}_{\tk{i}}^{\thetamMin{m, N}}(\tildeXle{24\,\text{hours}})$, while in theory we could get even better results by evaluating $\gY{}_{\tk{i}}^{\thetamMin{m, N}}(\tildeXle{\tk{i}-})$.\footnote{In practice both predictions $\gY{}_{\tk{i}}^{\thetamMin{m, N}}(\tildeXle{24\,\text{hours}})$ and $\gY{}_{\tk{i}}^{\thetamMin{m, N}}(\tildeXle{\tk{i}-})$ can be useful depending on the situation. One has to use $\gY{}_{t}^{\thetamMin{m, N}}(\tildeXle{24\,\text{hours}})$ if one is at the situation that one has observed the patient for exactly 24 hours and wants to forecast the next 24 hours. On the other hand, at any time $s$ one should use all the currently available information to make the optimal prediction $\gY{}_{t}^{\thetamMin{m, N}}(\tildeXle{s})=\gY{}_{t}^{\thetamMin{m, N}}(\tildeXle{\tau(s)})$. While our model is flexible enough to make predictions $\gY{}_{t}^{\thetamMin{m, N}}(\tildeXle{s})$ for every combination $t>s$, we only evaluate $\gY{}_{\tk{i}}^{\thetamMin{m, N}}(\tildeXle{24\,\text{hours}})$, since some competing methods compared to in \cite{herrera2021neural} are limited to this restricted setting of one fixed value of $s=24\,\text{hours}$. However, especially if one is  interested in now-casting, the performance of $\gY{}_{\tk{i}}^{\thetamMin{m, N}}(\tildeXle{\tk{i}-})$ would be the more relevant metric (in which we expect our model to outperform its competitors even more clearly).}
For the datasets with synthetically added noise, the variance of the noise is chosen relative to the variance of the data. Since the $\gls{dX}=41$ different coordinates of the Physionet dataset describe very different health parameters on different scales, we first compute each coordinate's standard deviation $\sigma_{data, j}$ on the training set. Then we add i.i.d.\ noise samples $\epsilon_{i,j} \sim \normalDistribution(0, \zeta^2 \, \sigma_{data, j}^2)$, for $\zeta \geq 0$, to each observed coordinate $\gX{}_{\tk{i}, j}$ of input samples for the model, i.e., to each observation of the training set as well as to each observation on the first 24 hours of the test set (but not to any of the observations on the second 24 hours of the test set, which are exclusively used for evaluating the model).
This is done for $\zeta \in \{ 0, 0.2, 0.4, 0.6, 0.8, 1 \}$, where the same seed is used for each of them, to make the results more comparable.

\textbf{Architecture.}
We use the PD-NJ-ODE with the following architecture. The latent dimension is $d_H = 50$ and all 3 neural networks have the same structure of 1 hidden layer with $\tanh$ activation function and $50$ nodes. 
The signature is used up to truncation level $2$, the encoder is recurrent and the decoder uses a residual connection.

\textbf{Training.}
For each dataset the model is trained with mini-batch size $50$ for $100$ epochs, once with the noise-adapted objective function \eqref{equ:Psi noisy obs} and once with the original one \eqref{equ:Psi}. 
Otherwise the same as in Section~\ref{sec:Details for Noisy Observations}.
We always report the minimal evaluation MSE among all trained epochs, as was done in prior works.
Due to the shorter training compared to the $175$ epochs in \citet{krach2022optimal}, the evaluation MSE of the original PD-NJ-ODE on the dataset without synthetic noise is slightly larger ($2 \cdot 10^{-3}$ compared to $1.93 \cdot 10^{-3}$).
We could imagine that for our new loss, one should actually use even more epochs rather than less epochs, since the additional term that was only present in the old loss pushed the jump-network~$\rho_{\theta_{2}}$ faster into the right direction directly from the start of the training, while the new loss trains the jump-network only indirectly and thus probably slower.

\section{Pseudocode}

\begin{figure*}[!p]
    \begin{algorithm}[H]
    \caption{Forward pass of the PD-NJ-ODE.
	A small step size $\Delta t$ is fixed and we denote $\tk{\gn{}+1} := \gls{T}$. 
    $\hbox{ODESolve}(f, x, (a,b))$ numerically solves the 1st-order ODE defined by $f$, taking inputs $x$, for $t \in (a,b)$.
	}
   \label{algo:1}
\begin{algorithmic}
   \STATE {\bfseries Input:} Data points {with} timestamps and masks $\{(\gX{}_i, \tk{i}, \gM{}_i)\}_{i=0\dots {\gn{}}}$ 
   \STATE {\bfseries Output:} Prediction process $\gY{}$
   
   \STATE set $H_{0-} = 0$
   \FOR{$i=0$ {\bfseries to} {$\gn{}$}} 
        \STATE construct $\tildeXle{\tk{i}}$ from data
        \STATE $S_i = \pim{m}(\tildeXle{\tk{i}} - \gX{}_0)$  \hfill $\triangleright$ compute truncated signature
		\STATE ${H_{\tk{i}}} = {\rho_{\theta_2}(H_{\tk{i}-}, \tk{i}, S_i, \gX{}_0})$ \hfill $\triangleright$ Update hidden state given next observation $x_i$
		\STATE {$\gY{}_{\tk{i}} = \tilde g_{\tildetheta{}_3}(H_{\tk{i}})$}  \hfill $\triangleright$ compute output
		\STATE $s \leftarrow \tk{i}$
		
		\WHILE {$s + \Delta t < \tk{i+1}$}
        		\STATE ${H_{s + \Delta t}} = \hbox{ODESolve}(f_{\theta_1},  (H_{s}, s, \tk{i}, S_i, \gX{}_0), (s, s + \Delta t))$ \hfill $\triangleright$ get next hidden state
        		\STATE {$\gY{}_{s + \Delta t} = \tilde g_{\tildetheta{}_3}(H_{s + \Delta t})$}  \hfill $\triangleright$ compute output
        		\STATE $s \leftarrow s + \Delta t$
        \ENDWHILE
        \STATE ${H_{\tk{i+1}-}} = \hbox{ODESolve}(f_{\theta_1},  (H_{s-}, s, \tk{i}, S_i, \gX{}_0), (s, \tk{i+1}))$
        \STATE {$\gY{}_{\tk{i+1}-} = \tilde g_{\tildetheta{}_3}(H_{(s + \Delta t)-})$}
   \ENDFOR
\end{algorithmic}
\end{algorithm}
\vspace{-0.5cm}
\end{figure*}
\begin{figure*}[!p]
    \begin{algorithm}[H]
    \caption{Computation of the standard loss for PD-NJ-ODE.
	A small $\epsilon > 0$ is used for numerical stability.
	}
   \label{algo:2}
\begin{algorithmic}
   \STATE {\bfseries Input:} $N$ samples of data points with timestamps and masks $\{(\gX{}_i^{(j)}, \tksup{i}{(j)}, \gM{}_i^{(j)})\}_{i=0\dots {\gn{(j)}}}$ and the respective prediction process $\gY{}^{(j)}$ of PD-NJ-ODE for $1 \leq j \leq N$
   \STATE {\bfseries Output:} Loss $L$
   
   \STATE $L = \frac{1}{N} \sum_{j=1}^N  \frac{1}{\gn{(j)}}\sum_{i=1}^{\gn{(j)}} \left(  \left\lvert \gM{i}^{(j)} \odot \left( \gX{}_{i}^{(j)} - \gY{}_{\tk{i}}^{(j)}  \right)  + \epsilon \right\rvert_2 + \left\lvert \gM{i}^{(j)} \odot \left( \gX{}_{i}^{(j)} - \gY{}_{\tk{i}-}^{(j) } \right) + \epsilon \right\rvert_2 \right)^2$
\end{algorithmic}
\end{algorithm}
\vspace{-0.5cm}
\end{figure*}
\begin{figure*}[!p]
    \begin{algorithm}[H]
    \caption{Computation of the noisy observation loss for PD-NJ-ODE.
	}
   \label{algo:3}
\begin{algorithmic}
   \STATE {\bfseries Input:} $N$ samples of noisy data points with timestamps, masks and conditional means of the noise $\{(\gO{}_i^{(j)}, \tksup{i}{(j)}, \gM{}_i^{(j)}, \beta_i^{(j)}) \}_{i=0\dots {\gn{(j)}}}$ and the respective prediction process $\gY{}^{(j)}$ of PD-NJ-ODE for $1 \leq j \leq N$
   \STATE {\bfseries Output:} Loss $L$
   
   \STATE $L = \frac{1}{N} \sum_{j=1}^N  \frac{1}{\gn{(j)}}\sum_{i=1}^{\gn{(j)}} \left\lvert \gM{i}^{(j)} \odot \left( \left( \gO{i}^{(j)} - \beta_i^{(j)} \right) - \gY{}_{\tk{i}-}^{(j)}  \right) \right\rvert_2^2$
\end{algorithmic}
\end{algorithm}
\vspace{-0.5cm}
\end{figure*}
\begin{figure*}[!p]
    \begin{algorithm}[H]
    \caption{Training of PD-NJ-ODE. 
	}
   \label{algo:4}
\begin{algorithmic}
   \STATE {\bfseries Input:} step size $\alpha$; number of epochs $E$; $N$ samples of noisy data points with timestamps, masks and conditional means of the noise $\{(\gO{}_i^{(j)}, \tksup{i}{(j)}, \gM{}_i^{(j)}, \beta_i^{(j)}) \}_{i=0\dots {\gn{(j)}}}$ for $1 \leq j \leq N$
   \STATE {\bfseries Output:} trained network parameters $\theta = (\theta_1, \theta_2, \tildetheta{}_3)$

   \STATE randomly initialize $\theta$ 
   \FOR{$e=0$ {\bfseries to} {$E$}} 
        \STATE split the dataset into random batches $B$
        \FOR{$b$ {\bfseries in} {$B$}}
            \FOR{each sample $j$ {\bfseries in} $b$}
                \STATE compute prediction $\gY{}^{(j)}$ with Algorithm~\ref{algo:1}
            \ENDFOR
            \STATE compute loss $L$ over all samples in the batch $b$ with Algorithm~\ref{algo:2} or~\ref{algo:3}
            \STATE update NN parameters $\theta \leftarrow \theta - \alpha \nabla_{\theta} L$
        \ENDFOR
   \ENDFOR
\end{algorithmic}
\end{algorithm}
\vspace{-0.5cm}
\end{figure*}

In addition to our code repository (\code), we provide pseudocode for the key parts of our method.
We remark that multiple python packages are available for computing the signature of a process numerically, as for example \texttt{ESig}, \texttt{iisignature} and \texttt{signatory}. Therefore, we do not provide pseudocode for its computation. For our implementation, we use the great \texttt{iisignature} package implemented by \cite{reizenstein2018iisignature}. We note that our definition of $\tildeXle{t}$, with its time-consistency, allows for efficient online updates of the signature (i.e., we do not have to recompute the signature for the whole path at every new observation time $\tk{k}$).  

In Algorithm~\ref{algo:1} we present the forward pass of the PD-NJ-ODE, which can be used for training as well as for evaluating the model. Algorithms~\ref{algo:2} and~\ref{algo:3} present the standard loss function and the loss function for noisy observations, respectively. The training for the model follows the standard neural network training approach via stochastic gradient descent (SGD) and is presented in Algorithm~\ref{algo:4}. We assume that a library is used for the neural network implementation which allows for automatic differentiation, as for example \texttt{PyTorch} \citep{NEURIPS2019_9015}, which we used in our implementation. For simplicity, we present the simplest type of SGD, with a fixed step size and no momentum or weight decay. In our implementation we actually use the more sophisticated method Adam \citep{adam}.

\printglossary[title=Notation,toctitle=Notation]

\fi


\end{document}